\documentclass[10pt,twocolumn,letterpaper]{article}

\usepackage{cvpr}              
\usepackage[table,xcdraw,dvipsnames]{xcolor}

\usepackage{multirow}
\usepackage[export]{adjustbox}
\usepackage{bbm}
\usepackage{xcolor,amsmath}
\usepackage[linesnumbered,ruled,vlined]{algorithm2e}
\DontPrintSemicolon

\SetKwComment{Comment}{\color{green!50!black}// }{}

\SetKw{breakloop}{break}
\SetKwProg{Function}{function}{}{}

\newcommand\lft{\mathopen{}\left}
\newcommand\rgt{\aftergroup\mathclose\aftergroup{\aftergroup}\right}

\definecolor{darkgreen}{RGB}{0,160,0}

\newcommand{\scenename}[1]{\textit{#1}}
\newcommand{\figninewidth}{3.1cm}
\newcommand{\figsixwidth}{3.1cm}

\newif\ifblackandwhitecycle
\gdef\patternnumber{0}

\makeatletter
\@namedef{ver@everyshi.sty}{}
\makeatother
\usepackage{tikz}
\usepackage{pgfplots}
\usetikzlibrary{spy,calc}

\pgfplotsset{compat=1.16}

\newcommand{\myparagraph}[1]{ \vspace{3pt}  \noindent {\bf #1}\,\,\,}

\newcommand{\acronym}{EVER}
\newcommand{\longname}{Exact Volumetric Ellipsoid Rendering}

\newif\ifblackandwhitecycle
\gdef\patternnumber{0}

\pgfkeys{/tikz/.cd,
    zoombox paths/.style={
        draw=orange,
        very thick
    },
    black and white/.is choice,
    black and white/.default=static,
    black and white/static/.style={ 
        draw=white,   
        zoombox paths/.append style={
            draw=white,
            postaction={
                draw=black,
                loosely dashed
            }
        }
    },
    black and white/static/.code={
        \gdef\patternnumber{1}
    },
    black and white/cycle/.code={
        \blackandwhitecycletrue
        \gdef\patternnumber{1}
    },
    black and white pattern/.is choice,
    black and white pattern/0/.style={},
    black and white pattern/1/.style={    
            draw=white,
            postaction={
                draw=black,
                dash pattern=on 2pt off 2pt
            }
    },
    black and white pattern/2/.style={    
            draw=white,
            postaction={
                draw=black,
                dash pattern=on 4pt off 4pt
            }
    },
    black and white pattern/3/.style={    
            draw=white,
            postaction={
                draw=black,
                dash pattern=on 4pt off 4pt on 1pt off 4pt
            }
    },
    black and white pattern/4/.style={    
            draw=white,
            postaction={
                draw=black,
                dash pattern=on 4pt off 2pt on 2 pt off 2pt on 2 pt off 2pt
            }
    },
    zoomboxarray inner gap/.initial=3pt,
    zoomboxarray columns/.initial=1,
    zoomboxarray rows/.initial=2,
    subfigurename/.initial={},
    figurename/.initial={zoombox},
    zoomboxarray/.style={
        execute at begin picture={
            \begin{scope}[
                spy using outlines={%
                    zoombox paths,
                    width=\imagewidth / \pgfkeysvalueof{/tikz/zoomboxarray columns} - (\pgfkeysvalueof{/tikz/zoomboxarray columns} - 1) / \pgfkeysvalueof{/tikz/zoomboxarray columns} * \pgfkeysvalueof{/tikz/zoomboxarray inner gap} -\pgflinewidth,
                    height=\imageheight / \pgfkeysvalueof{/tikz/zoomboxarray rows} - (\pgfkeysvalueof{/tikz/zoomboxarray rows} - 1) / \pgfkeysvalueof{/tikz/zoomboxarray rows} * \pgfkeysvalueof{/tikz/zoomboxarray inner gap}-\pgflinewidth,
                    magnification=3,
                    every spy on node/.style={
                        zoombox paths
                    },
                    every spy in node/.style={
                        zoombox paths
                    }
                }
            ]
        },
        execute at end picture={
            \end{scope}
            \node at (image.north) [anchor=north,inner sep=0pt] {\phantomimage};
            \node at (zoomboxes container.north) [anchor=north,inner sep=0pt] {\phantomimage};
     \gdef\patternnumber{0}
        },
        spymargin/.initial=0.5em,
        zoomboxes xshift/.initial=1,
        zoomboxes right/.code=\pgfkeys{/tikz/zoomboxes xshift=1},
        zoomboxes left/.code=\pgfkeys{/tikz/zoomboxes xshift=-1},
        zoomboxes yshift/.initial=0,
        zoomboxes above/.code={
            \pgfkeys{/tikz/zoomboxes yshift=1},
            \pgfkeys{/tikz/zoomboxes xshift=0}
        },
        zoomboxes below/.code={
            \pgfkeys{/tikz/zoomboxes yshift=-1},
            \pgfkeys{/tikz/zoomboxes xshift=0}
        },
        caption margin/.initial=0ex,
    },
    adjust caption spacing/.code={},
    image container/.style={
        inner sep=0pt,
        at=(image.north),
        anchor=north,
        adjust caption spacing
    },
    zoomboxes container/.style={
        inner sep=0pt,
        at=(image.north),
        anchor=north,
        name=zoomboxes container,
        xshift=\pgfkeysvalueof{/tikz/zoomboxes xshift}*(\imagewidth+\pgfkeysvalueof{/tikz/spymargin}),
        yshift=\pgfkeysvalueof{/tikz/zoomboxes yshift}*(\imageheight+\pgfkeysvalueof{/tikz/spymargin}+\pgfkeysvalueof{/tikz/caption margin}),
        adjust caption spacing
    },
    calculate dimensions/.code={
        \pgfpointdiff{\pgfpointanchor{image}{south west} }{\pgfpointanchor{image}{north east} }
        \pgfgetlastxy{\imagewidth}{\imageheight}
        \global\let\imagewidth=\imagewidth
        \global\let\imageheight=\imageheight
        \gdef\columncount{1}
        \gdef\rowcount{1}
        
    },
    image node/.style={
        inner sep=0pt,
        name=image,
        anchor=south west,
        append after command={
            [calculate dimensions]
            node [image container,subfigurename=\pgfkeysvalueof{/tikz/figurename}-image] {\phantomimage}
            node [zoomboxes container,subfigurename=\pgfkeysvalueof{/tikz/figurename}-zoom] {\phantomimage}
        }
    },
    color code/.style={
        zoombox paths/.append style={draw=#1}
    },
    connect zoomboxes/.style={
    spy connection path={\draw[draw=none,zoombox paths] (tikzspyonnode) -- (tikzspyinnode);}
    },
    help grid code/.code={
        \begin{scope}[
                x={(image.south east)},
                y={(image.north west)},
                font=\footnotesize,
                help lines,
                overlay
            ]
            \foreach \x in {0,1,...,9} { 
                \draw(\x/10,0) -- (\x/10,1);
                \node [anchor=north] at (\x/10,0) {0.\x};
            }
            \foreach \y in {0,1,...,9} {
                \draw(0,\y/10) -- (1,\y/10);                        \node [anchor=east] at (0,\y/10) {0.\y};
            }
        \end{scope}    
    },
    help grid/.style={
        append after command={
            [help grid code]
        }
    },
}

\newcommand\phantomimage{%
    \phantom{%
        \rule{\imagewidth}{\imageheight}%
    }%
}
\newcommand\zoombox[2][]{
    \begin{scope}[zoombox paths]
        \pgfmathsetmacro\xpos{
            (\columncount-1)*(\imagewidth / \pgfkeysvalueof{/tikz/zoomboxarray columns} + \pgfkeysvalueof{/tikz/zoomboxarray inner gap} / \pgfkeysvalueof{/tikz/zoomboxarray columns} )
        }
        \pgfmathsetmacro\ypos{
            (\rowcount-1)*( \imageheight / \pgfkeysvalueof{/tikz/zoomboxarray rows} + \pgfkeysvalueof{/tikz/zoomboxarray inner gap} / \pgfkeysvalueof{/tikz/zoomboxarray rows} )
        }
        \edef\dospy{\noexpand\spy [
            #1,
            zoombox paths/.append style={
                black and white pattern=\patternnumber
            },
            every spy on node/.append style={#1},
            x=\imagewidth,
            y=\imageheight
        ] on (#2) in node [anchor=north west] at ($(zoomboxes container.north west)+(\xpos pt,-\ypos pt)$);}
        \dospy
        \pgfmathtruncatemacro\pgfmathresult{ifthenelse(\columncount==\pgfkeysvalueof{/tikz/zoomboxarray columns},\rowcount+1,\rowcount)}
        \global\let\rowcount=\pgfmathresult
        \pgfmathtruncatemacro\pgfmathresult{ifthenelse(\columncount==\pgfkeysvalueof{/tikz/zoomboxarray columns},1,\columncount+1)}
        \global\let\columncount=\pgfmathresult
        \ifblackandwhitecycle
            \pgfmathtruncatemacro{\newpatternnumber}{\patternnumber+1}
            \global\edef\patternnumber{\newpatternnumber}
        \fi
    \end{scope}
}

\definecolor{cvprblue}{rgb}{0.21,0.49,0.74}
\usepackage[pagebackref,breaklinks,colorlinks,citecolor=cvprblue]{hyperref}

\def\paperID{12662} 
\def\confName{ICCV}
\def\confYear{2025}

\title{\acronym{}: \longname{} for Real-time View Synthesis}

\newcommand{\authorbox}[2]{\makebox[0pt][l]{#1$^#2$}\phantom{George Kopanas}}
\newcommand{\aemail}[1]{\makebox[0pt][l]{{\tt\small #1}}\phantom{{\tt\small dfutschik@google.com}}}

\newcommand{\institutes}[0]{$^1$University of California, San Diego \quad $^2$Google \quad $^*$work done while at Google}

\author{
\authorbox{Alexander Mai$^1$}{*}\\
\aemail{amai@ucsd.edu}
\and
\authorbox{Peter Hedman}{2}\\
\aemail{hedman@google.com}
\and
\authorbox{George Kopanas}{2}\\
\aemail{gkopanas@google.com}
\and
\authorbox{Dor Verbin}{2}\\
\aemail{dorverbin@google.com}
\and
\authorbox{David Futschik}{2}\\
\aemail{dfutschik@google.com}
\and
\authorbox{Qiangeng Xu}{2}\\
\aemail{qiangenx@google.com}
\and
\authorbox{Falko Kuester}{3}\\
\aemail{fkuester@ucsd.edu}
\and
\authorbox{Jonathan T. Barron}{2}\\
\aemail{barron@google.com}
\and
\authorbox{Yinda Zhang}{2}\\
\aemail{yindaz@google.com}
}

\begin{document}

\twocolumn[{%
\renewcommand\twocolumn[1][]{#1}%
\maketitle
\begin{center}
    \centering
    \includegraphics[width=\textwidth]{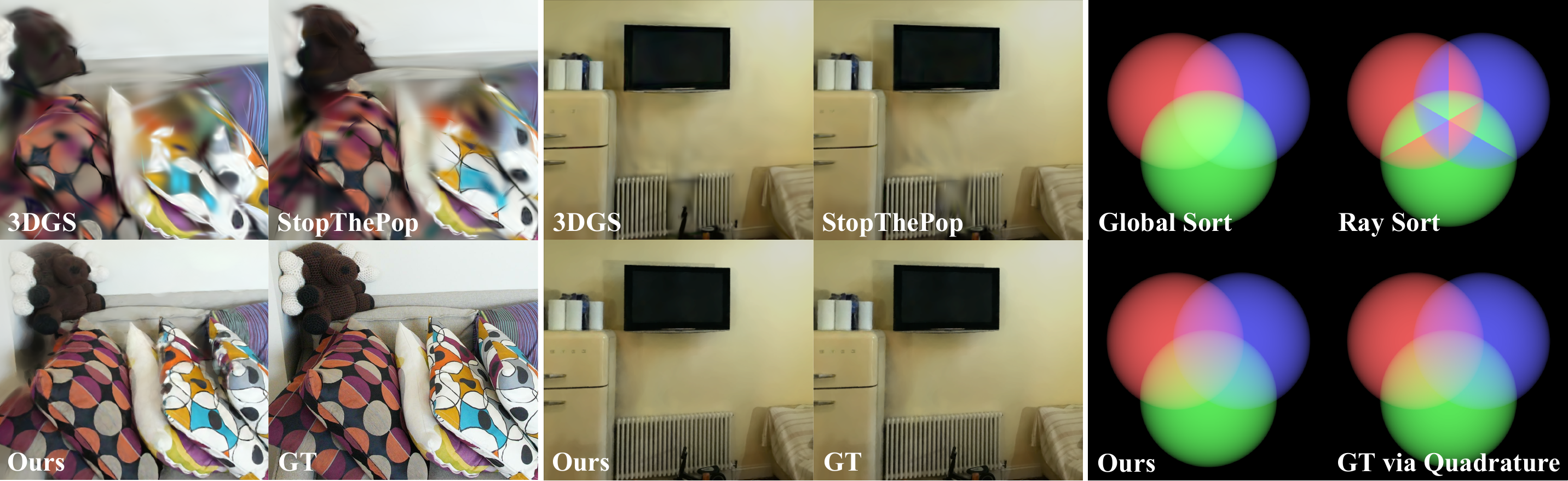}

\captionof{figure}{An overview of the quality benefits of our EVER technique. 
Left: On the Zip-NeRF dataset, our model produces the sharpest results ever. 
Middle: Our method is able to blend primitive colors together to reproduce light and shadow better than other 3DGS-based methods.
Right: The way our approach blends colors is physically accurate over any number of primitives, matching the ground truth computed via quadrature.
}
\label{fig:frontpage}
\end{center}
}]


\begin{abstract}
We present \longname{} (\acronym{}), a method for real-time 3D reconstruction.
\acronym{} accurately blends an unlimited number of overlapping primitives together in 3D space, eliminating the popping artifacts that 3D Gaussian Splatting (3DGS) and other related methods exhibit.
EVER represents a radiance field as a set of constant-density volumetric ellipsoids, which are raytraced by intersecting each primitive twice (once upon ray entrance and another on ray exit) and accumulating the derivatives of the densities and colors along the ray.
Because EVER is built around ray tracing, it also enables effects such as defocus blur and fish-eye camera distortion, while still achieving frame rates of $\sim\!30$ FPS at 720p on an NVIDIA RTX4090. 
We show that our method is more accurate on the challenging large-scale scenes from the Zip-NeRF dataset, where it achieves state of the art SSIM, even higher than Zip-NeRF.

\end{abstract}

    
\section{Introduction}
\label{sec:intro}
The field of 3D reconstruction for novel view synthesis has explored a variety of scene representations: point based~\cite{kerbl20233d}, surface based~\cite{li2023neuralangelo, yariv2023bakedsdf, wang2021neus}, and volume based~\cite{mildenhall2020nerf, muller2022instant}. Since their introduction in NeRF~\cite{mildenhall2020nerf}, differentiable rendering of volumetric scene representations have become popular due to their ability to yield photorealistic 3D reconstructions.
More recently, 3D Gaussian Splatting (3DGS) combined the speed of point based models with the differentiability of volume based representations by representing the scene as a collection of millions of Gaussian primitives that can be rendered via rasterization in real-time.

Unlike NeRF, 3DGS lacks a true volumetric density field, and uses Gaussians to describe the \emph{opacity} of the scene rather than of density.
As such, 3DGS's scene representation isn't volumetric rendering, as an anisotropic Gaussian primitive in 3DGS will appear to have the same opacity regardless of the viewing direction of the camera.
This lack of a consistent underlying density field prohibits the use of various algorithms and regularizers (such as the distortion loss from Mip-NeRF360~\cite{barron2021mip}), but the biggest downside of this model is popping.
Popping in 3DGS is due to two assumptions made by the model: that primitives do not overlap, and that (given a camera position) primitives can be sorted accurately using only their centers. These assumptions are almost always violated in practice, which causes the rendered image to change significantly as the camera moves due to the sort-order of primitives changing.
This popping may not be noticeable when primitives are small, but describing a large scene using many small primitives requires a prohibitively large amount of memory.

In this work we build on the primitive based representation of 3DGS, but introduce a method that allows for the physically accurate blending of an infinite number of overlapping number of primitives, specifically constant density ellipsoids.
We implement this blending method in a ray-tracing framework, which allows us to model various optical effects like radial distortion lenses including fisheye and defocus blur in a straightforward way, all at real-time framerates.
Our method guarantees 3D consistent real-time rendering while also improving the image quality of our 3DGS baselines. This is particularly pronounced in the challenging large-scale Zip-NeRF scenes~\cite{barron2023zip} where our method matches the quality of state-of-the-art offline rendering methods.

\section{Related Work}

\myparagraph{Neural Volume Rendering} Neural Radiance Fields (NeRF)~\cite{mildenhall2020nerf} introduced the paradigm of representing a scene as an emissive volume, using a neural network with position encoding that is rendered differentiably using quadrature~\cite{max1995optical, drebin1988volume} and optimized using gradient descent. 
While the original formulation used a neural network, follow up work has used voxel grids~\cite{fridovich2022plenoxels, sun2021direct}, hash grids~\cite{muller2022instant}, tri-planes~\cite{chan2022efficient}, primitives~\cite{lombardi2021mixture}, and points~\cite{xu2022point}. These papers all use numerical quadrature to approximately integrate the volume rendering equation.

While NeRF reconstructions are slow to render they can be converted into faster representations, such as triangle meshes~\cite{chen2022mobilenerf,yariv2023bakedsdf,rakotosaona2023nerfmeshing,rojas2023re,Reiser2024SIGGRAPH}, sparse volumes~\cite{garbin2021fastnerf,yu2021plenoctrees,reiser2023merf,duckworth2024smerf}, mesh-volume hybrids~\cite{Wan_2023_CVPR,adaptiveshells2023,turki2023hybridnerf} and 3D Gaussians~\cite{niemeyer2024radsplat}. While these facilitate real-time rendering, creating them is a slow two-step process that first trains a NeRF and then converts it to a faster representation. In the experiments we compare with SMERF~\cite{duckworth2024smerf}, the current state-of-the-art for real-time NeRF rendering.


\myparagraph{Differentiable Point-based Rendering} Like NeRF, 3D Gaussian Splatting (3DGS)~\cite{kerbl20233d} models the scene as a radiance field, but 3DGS represents radiance as a set of Gaussians that are rendered via splatting~\cite{zwicker2001ewa} while NeRF represents radiance using a field that is rendered via ray-marching. 
While 3DGS approximates volume rendering with view-independent opacity and view-dependent radiance, in practice it often yields highly accurate renderings, and these approximations allow it to be trained and rendered quickly, hence its popularity~\cite{chen2024survey}. Since its inception, improvements have been made to 3DGS in terms of aliasing~\cite{geiger2023mip}, camera models~\cite{moenne20243d}, heuristic densification~\cite{bulo2024revising, kheradmand20243d, ververas2024sags, cao2024lightweight, ye2024absgs, yu2024gaussian}, and view-consistency (i.e. ``popping''). 

Popping results from how 3DGS sorts Gaussians once per-frame using their mean, and it has been partly addressed by StopThePop~\cite{radl2024stopthepop} during rasterization and 3DGRT~\cite{moenne20243d} during ray-tracing.  However, StopThePop only approximately sorts per-ray, and both StopThePop and 3DGRT ignore overlap between the primitives. This introduces a blend order artifact shown in Fig.~\ref{fig:popping_train}.  
One way of addressing popping is to compute the volumetric rendering equation with fewer (or zero) approximations. Concurrently with our work, a number of preprints attempt this~\cite{blanc2024, condor2024volumetric, zhou2024unified} but they all achieve significantly slower performance and are subject to significant limitations regarding how much primitives may overlap, or use a different approximation~\cite{wang2022voge}. 
By representing the scene with constant density primitives, our model is able to quickly and exactly compute the volumetric rendering equation, even with unlimited overlap.

\newcommand{\figtwowidth}{3.9cm}
\begin{figure*}
    \centering
    \begin{tabular}{@{}c c@{\hspace{3pt}}ccc@{}}
    \includegraphics[width=\figtwowidth,valign=c]{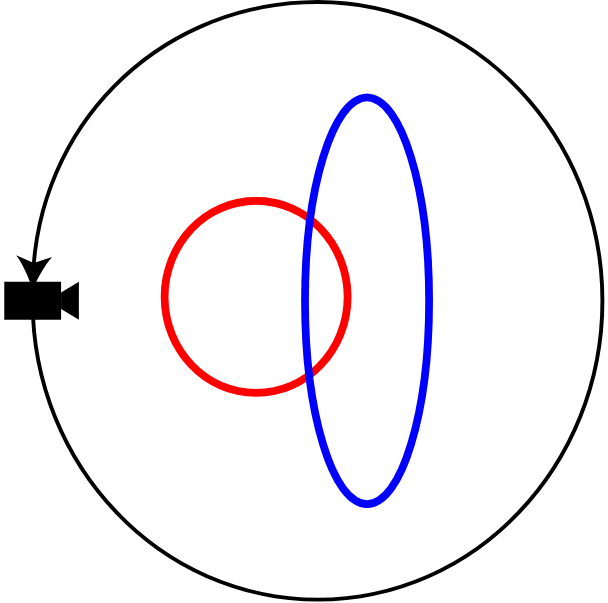} &
    \raisebox{-10pt}{$\bar{\downarrow}$} &
    \includegraphics[width=\figtwowidth,valign=c]{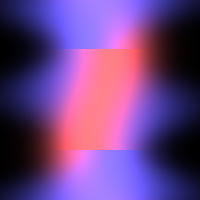} &
    \includegraphics[width=\figtwowidth,valign=c]{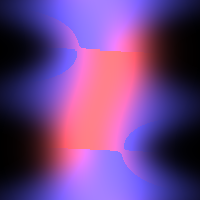} &
    \includegraphics[width=\figtwowidth,valign=c]{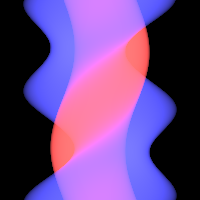} \\
    \small (a) Top Down View & &
    \small (b) 3DGS~\cite{kerbl20233d} &
    \small (c) StopThePop~\cite{radl2024stopthepop} &
    \small (d) Our Model
    \end{tabular}
    \vspace{-3pt}
    \caption{
    (a) Here we show a simple ``flatland'' scene containing two primitives (one red, one blue) with a camera orbiting them, viewed from above.
    We render this orbit using three different techniques, where each camera position yields a one-dimensional ``image'' (a scanline) which are stacked vertically to produce these epipolar plane image (EPI) visualizations.
    (b, c) The approximations made by approximate splatting-based techniques result in improper blending due to discontinuities, which are visible as horizontal lines across the EPI.
    In contrast, (d) our method's exact rendering yields a smooth EPI, with bands of purple from color blending.
    }
    \label{fig:popping_diagram}
\end{figure*}

\section{Motivation}\label{sec:motivation}

\myparagraph{Neural Radiance Fields} 
A radiance field is comprised of two spatially-varying fields as a function of spatial coordinate $\mathbf{x}$: density $\sigma(\mathbf{x})$ and color $c(\mathbf{x}, \mathbf{d})$, where the color field may depend on viewing direction $\mathbf{d}$ in addition to $\mathbf{x}$. A ray with origin $\mathbf{o}$ and direction $\mathbf{d}$ is rendered by integrating these fields using standard volume rendering integral (also known as the radiative transfer equation~\cite{chandrasekhar1960radiative}:
\begin{equation} \label{eqn:volumerendering}
    C = \int_0^\infty c_\mathbf{r}(t)\sigma_\mathbf{r}(t)\exp\lft(-\int_0^t \sigma_\mathbf{r}(s)\,ds\rgt)\,dt \, ,
\end{equation}
where $t$ is distance along a ray, and $\sigma_\mathbf{r}(t) = \sigma(\mathbf{o}+t\mathbf{d})$ and $c_\mathbf{r}(t) = c(\mathbf{o}+t\mathbf{d}, \mathbf{d})$ are the density and color along the ray, respectively.

The parameterization of the density and color fields can vary from small MLPs~\cite{mildenhall2020nerf} to large hierarchies of hashes and grids~\cite{muller2022instant}. Though this approach can produce highly realistic renderings, accurately integrating Equation~\ref{eqn:volumerendering} requires a large number of calls to the underlying field, which means that training and rendering may be slow.

\myparagraph{3D Gaussian Splatting} 
Like NeRF, 3DGS uses alpha compositing to render an image from volumetric primitives, but unlike NeRF it does not directly model a density field. Instead, a collection of Gaussians are projected onto frontoparallel billboards, multiplied by their opacities, sorted, and alpha composited together.
Because this rendering process is data-parallel, 3DGS can be implemented to yield extremely high framerates. However, 3DGS is not 3D consistent;
when the order in which the Gaussians are composed changes, the color abruptly changes as well. This lack of blending is visible as popping artifacts, as can be seen in Figure~\ref{fig:popping_diagram}.

\section{Method}
Like 3DGS, our method takes as input a set of posed images and a sparse point cloud.
We optimize a mixture of ellipsoids (each with a constant density and color) to reproduce the appearance of the input images.
Our method enables \emph{exact} volume rendering. In Section~\ref{sec:rendering} we describe our scene representation and the algorithm for volume rendering it, and in Section~\ref{sec:parameterizing_density} we describe how we optimize it to the input images.

\subsection{Exact Primitive-based Rendering} \label{sec:rendering}

\begin{figure}[b]
    \centering
    \includegraphics[width=0.45\textwidth]{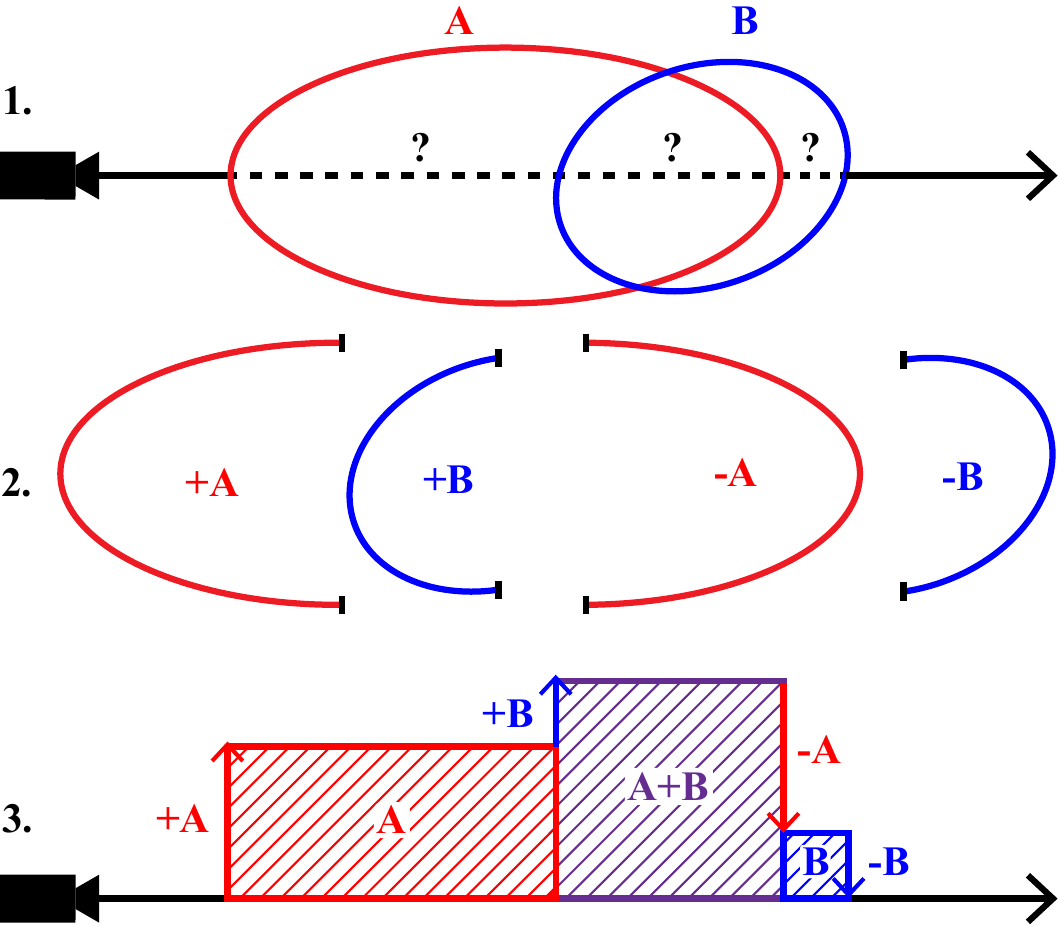}
    \caption{A visualization of our blending algorithm. 1. We start with our primitives, in this case two, and we are trying to recover the value of the intervals along the ray. 2. We take the change in density and color and sort then from closest to furthest. 3. We sum these changes as we progress through the ray, retrieving the interval values.
    }
    \label{fig:ellipsoid_field}
\end{figure}

We use a simple primitive-based rendering model, where each primitive has a constant density and a single color (in 3D space).
We choose our primitive's shape to be an ellipsoid, which, similar to the Gaussians in 3DGS, is fully characterized by a position vector, rotation quaternion, and scale vector. To model view-dependent color, the ellipsoid's color changes with the view direction, which we store as a spherical function represented by the coefficients for the first two degrees of spherical harmonics.

\newcommand{\vcolor}{\mathbf{c}}
To render a given ray, we iterate over all front and back surfaces of all primitives intersected by the ray. When we step through the front of a primitive, we add its density $\Delta\sigma_k$ and premultiplied color $\Delta\sigma \Delta\vcolor_k$ to two running totals, $\sigma_i, \hat{\vcolor}_i$. When we step through the back, we subtract the same density and pre-multiplied color. Using these running totals, we can calculate the color $\vcolor_i$ and density $\sigma_i$ for the $i$th interval:
\begin{align}
    \sigma_i = \sum_{k=1}^i \Delta \sigma_k\,, \quad
    \vcolor_i = \frac{1}{\sigma_i} \hat{\vcolor}_i = \frac{1}{\sigma_i} \sum_{k=1}^i \Delta \sigma_k \Delta \vcolor_k\,.
\end{align}
The key here is that this decoupling of the primitive into the front and back allow us to rearrange the surfaces along the ray, as seen in Figure~\ref{fig:ellipsoid_field}. In this example, we can see two overlapping primitives. In 3DGS, we would simply render A and B separately, ignoring their overlap, and then combine them. In contrast, our method allows rendering this representation exactly by computing the density and color at the three intervals A, A+B, and B, by tracking their changes as we progress through the ray.
Since color and density are constant within each interval of length $\Delta t_i$ between consecutive intersection points, there is a simple closed-form solution to the volume rendering equation:
\begin{equation}
    \vcolor_i (1-\exp(-\sigma_i \Delta t_i)) \,,\label{eqn:closed form}
\end{equation}
By tracking the running total and the distance between the intersection steps, $\Delta t_i$, we are able to integrate the volume rendering equation exactly by alpha compositing these simple segments, as follows:
\begin{equation}
    C = \sum_{i=1}^N \vcolor_i (1-\exp(-\sigma_i \Delta t_i)) \prod_{j=1}^{i-1} \exp(-\sigma_j \Delta t_j)\,.\label{eqn:alphacomp}
\end{equation}
Unsurprisingly, the expression in Equation~\ref{eqn:alphacomp} is identical to the quadrature rule derived by Max~\cite{max1995optical}, which uses a piecewise-constant approximation to density and color --- though in our case this piecewise-constant property is an intentional design decision, not an approximation. Our work uses this closed-form solution to enable real-time rendering of the entire collection of ellipsoids without introducing any approximation errors which may cause artifacts.

\begin{figure}
    \centering
    \includegraphics[width=0.45\textwidth]{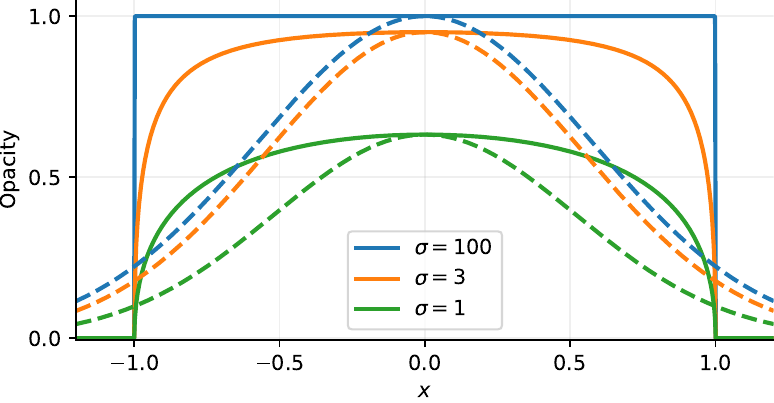}
    \caption{Here we show 1D screen-space slices of rendered opacity profiles for different rendering methods, to demonstrate one advantage of our density-based parameterization. We show 3 different peak opacities and their profiles, where solid lines represent our density based ellipsoid primitives and dashed lines represent 3DGS. For our model we show densities $\sigma \in \{1, 3, 100\}$, while for 3DGS we compute the corresponding peak opacity such that each slice has the same maximum opacity as our model. While Gaussians always have smooth opacity profiles that limit their ability to reproduce edges in image space, our opacity profiles can be smooth ($\sigma = 1$) or sharp ($\sigma = 100$).} 
    \label{fig:alpha_profiles}
\end{figure}

Note that although our scene representation looks superficially similar to that of 3DGS, it is different in two key ways: First, while 3DGS Gaussians are treated as 2D ``billboards'', our ellipsoids blend together so as to constitute a proper and consistent 3D radiance field. Second, while the 2D opacity profile of the primitives used in 3DGS always has a smooth Gaussian falloff, the 2D opacity profile of one of our ellipsoidal primitives can range from extremely smooth to a perfect step function in the limit of infinite density. See Figure~\ref{fig:alpha_profiles} for examples of our primitive's opacity profiles for different density values. 
\subsection{Optimization} \label{sec:parameterizing_density}

Our optimization framework is built on top of 3DGS, where instead of rasterizing Gaussians we ray trace and volume render our ellipsoid-based representation using the method described in Section~\ref{sec:rendering}. In particular, we reuse the Adaptive Density Control (ADC) introduced in 3DGS, with some modifications for our density-based primitives.

The first, essential change is to convert from the 3DGS opacity framework to a density framework.
The naive approach would be to assign each primitive a density instead of opacity, but this presents a challenge. As the  density of a primitive grows and its opacity approaches $1$, the gradient used for updating the parameters of the primitive approaches $0$ and would require special treatment~\cite{li2018differentiable}. Intuitively, this is because the outside of the primitive becomes opaque.

We avoid this problem by optimizing over a parameter $\alpha_k$ which is mapped to density value used during rendering:
\begin{equation}
    \sigma(\alpha_k) = -\frac{\log(1-0.99 \cdot \alpha_k)}{\min(s_{kx}, s_{ky}, s_{kz})}\, .
\end{equation}
With this, a primitive with density $\sigma(\alpha_k)$ when viewed along its shortest axis will have a peak opacity equal to $\alpha_k$, ensuring at least one view will have non-zero gradient.

Finally, we add an additional splitting condition: We split primitives if the opacity is near 1 across the entire primitive when viewed from the major axis to see if the gradient has vanished using the following test:
\begin{equation}
    0.99 < 1 - \exp\lft(-\sigma(\alpha_k) \cdot \max\lft(s_{kx}, s_{ky}, s_{kz}\rgt)\rgt)\,.
\end{equation}
\section{Implementation}
Our model was implemented using PyTorch, CUDA, OptiX~\cite{parker2010optix}, and Slang~\cite{he2018slang}. 
We use OptiX to ray trace through the primitives and to provide a per-ray sorting of distances. During training we rebuild our BVH at every step, which takes tens of milliseconds.
All shaders are written in Slang~\cite{he2018slang}, a language that supports automatic differentiation with the Slang.D extension~\cite{bangaru2023slangd}. We use adjoint rendering to propagate the gradient, which means we skip storage of all function values and simply recompute them during the backwards pass.
To optimize the representation, we use our differentiable renderer in the 3DGS~\cite{kerbl20233d} codebase, and make a few adjustments to handle density based primitives (Section~\ref{sec:adc_changes} and Appendix~\ref{sec:hparams}).

\begin{figure}
    \centering
    \begin{tabular}{@{}c@{}}
    \includegraphics[width=0.8\linewidth]{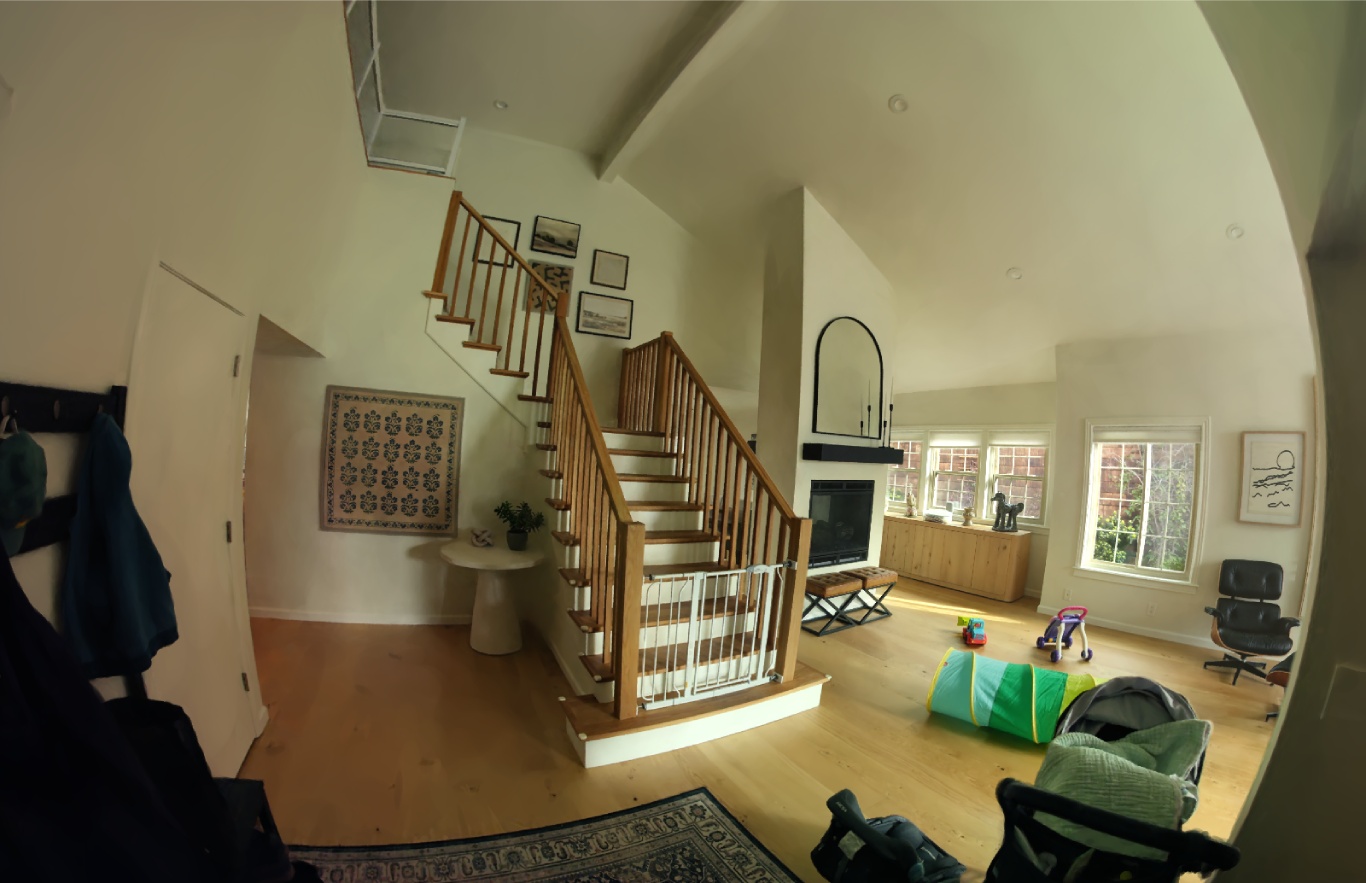} \\
    \small{(a) Fisheye training and rendering} \\[0.15cm]
    \includegraphics[width=0.8\linewidth]{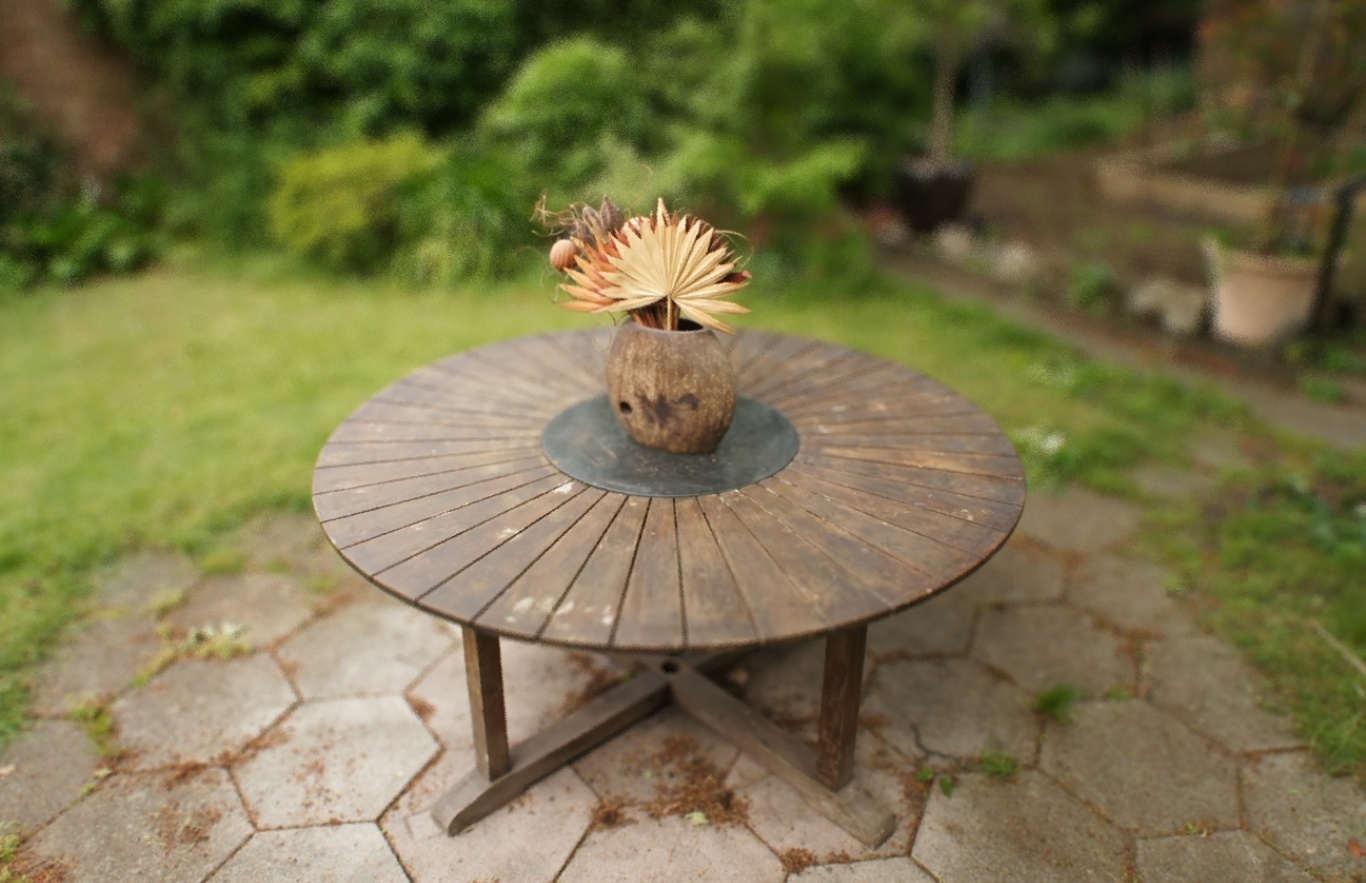} \\
    \small{(b) Shallow depth of field rendering}
    \end{tabular}
    \vspace{-0.2cm} 
    \caption{A depiction of two benefits of ray tracing.}
    \label{fig:blur_fisheye}
\end{figure}

\subsection{Ray Tracing (BVHs) for Sorting}\label{sec:bvh}

We use ray tracing to sort primitives since it offers more flexibility than a rasterization approach like StP~\cite{radl2024stopthepop}. Ray tracing allows for effects like fisheye projection and defocus blur (see Figure~\ref{fig:blur_fisheye}) as well as random pixel offsets during training, which we find improves performance.

We start by providing the NVIDIA OptiX~\cite{parker2010optix} framework with an Axis-Aligned Bounding Box (AABB) for the ellipsoid, which is used to construct a binary tree. We then provide an intersection test to refine the search if a ray hits the bounding box, for which we adapt the stable isotropic ray/sphere intersection described by Haines et al~\cite{Haines2019}.
To further accelerate rendering, we also integrate the recent approach of 3DGRT~\cite{moenne20243d}, which accumulates multiple hits within a queue during each call to OptiX trace. 

BVH efficiency depends strongly on how often the bounding boxes of primitives overlap. Rotated anisotropic primitives tend to have large axis-aligned bounding boxes that slow down rendering. To ameliorate this, we impose a loss during training to discourage overly anisotropic primitives:
\begin{equation}
    \!\operatorname{stopgrad}\lft(1-\alpha\rgt)\lft(\max\lft(s_x, s_y, s_z\rgt) - \min\lft(s_x, s_y, s_z\rgt)\rgt),\!
\end{equation}
where $\alpha\in[0,1]$ is the opacity and  $\mathbf{s}\in \mathbb{R}^3$ is the ellipsoid axis scaling vector for each primitive.
This regularizer is applied only to primitives that are visible in the current batch. 


\subsection{Changes to Adaptive Density Control} \label{sec:adc_changes}

The Adaptive Density Control (ADC) used by 3DGS is critical to its success.
During training 3DGS uses a variety of strategies for periodically cloning, splitting, and pruning 3D Gaussians.
This is essential to avoid local minima by creating primitives where they are necessary and removing primitives where they are invisible.
Though these heuristics were developed specifically for 3DGS, they work well for our method with some minor necessary adjustments.

To decide where to create primitives, 3DGS accumulates the $xy$ gradients of each primitive, then splits or clones them if they are above a certain threshold every $n$ steps.
The spatial gradients of our model tend to be smaller, so the thresholds used by these splitting and cloning heuristics must be modified (splitting is $2.5\times10^{-7}$, cloning is $0.1$).
We use a similar splitting method as 3DGS: the primitive size is reduced and the new position is perturbed by some random amount sampled from a normal distribution, but we additionally also divide the density in half, similar to~\cite{bulo2024revising}.


\section{Results}

\begin{table*}[t!]
    \centering
    \resizebox{\linewidth}{!}{
        \begin{tabular}{l|rrr|rrr|rrr}
                                     & \multicolumn{3}{c|}{Mip-NeRF360}                                                                                                      & \multicolumn{3}{c|}{Zip-NeRF}                                                                                             & \multicolumn{3}{c}{Tanks\&Temples \& DeepBlending}                                                                       \\
                                     & \multicolumn{1}{l}{PSNR $\uparrow$}                 & \multicolumn{1}{l}{SSIM $\uparrow$}   & \multicolumn{1}{l|}{LPIPS $\downarrow$} & \multicolumn{1}{l}{PSNR $\uparrow$}     & \multicolumn{1}{l}{SSIM $\uparrow$}   & \multicolumn{1}{l|}{LPIPS $\downarrow$} & \multicolumn{1}{l}{PSNR $\uparrow$}     & \multicolumn{1}{l}{SSIM $\uparrow$}   & \multicolumn{1}{l}{LPIPS $\downarrow$} \\ \hline
3DGS~\cite{kerbl20233d}              & \cellcolor[HTML]{FFFFB4}27.48                       & \cellcolor[HTML]{FFFFB4}.816          & .216                                    & 25.84                                   & .817                                  & .358                                    & \cellcolor[HTML]{FFB3B3}26.65           & \cellcolor[HTML]{FFFFB4}.848          & .263                                   \\
StopThePop~\cite{radl2024stopthepop} & 27.33                                               & \cellcolor[HTML]{FFFFB4}.816          & .212                                    & \cellcolor[HTML]{FFFFB4}25.92           & \cellcolor[HTML]{FFFFB4}.819          & \cellcolor[HTML]{FFFFB4}.352            & \cellcolor[HTML]{FFD9B3}26.60           & .847                                  & \cellcolor[HTML]{FFFFB4}.252           \\
3DGRT~\cite{moenne20243d}            & 27.20                                               & \cellcolor[HTML]{FFD9B3}.818          & \cellcolor[HTML]{FFFFB4}.248            & -                                       & -                                     & -                                       & 26.22                                   & \cellcolor[HTML]{FFD9B3}.865          & \cellcolor[HTML]{FFD9B3}.254           \\
SMERF~\cite{duckworth2024smerf}      & \cellcolor[HTML]{FFB3B3}27.99                       & \cellcolor[HTML]{FFD9B3}.818          & \cellcolor[HTML]{FFD9B3}.238            & \cellcolor[HTML]{FFB3B3}27.28           & \cellcolor[HTML]{FFD9B3}.829          & \cellcolor[HTML]{FFD9B3}.339            & -                                       & -                                     & -                                      \\
Our model                            & \cellcolor[HTML]{FFD9B3}27.51                       & \cellcolor[HTML]{FFB3B3}.825          & \cellcolor[HTML]{FFB3B3}.194            & \cellcolor[HTML]{FFD9B3}26.58           & \cellcolor[HTML]{FFB3B3}.845          & \cellcolor[HTML]{FFB3B3}.308            & \cellcolor[HTML]{FFFFB4}26.59           & \cellcolor[HTML]{FFB3B3}.889          & \cellcolor[HTML]{FFB3B3}.234           \\ \hline
ZipNeRF~\cite{barron2023zip}         & 28.54                                               & .828                                  & .198                                    & 27.37                                   & .836                                  & .305                                    & -                                       & -                                     & -                                      \\
                                     & \multicolumn{1}{l}{GPU-hr $\downarrow$}             & \multicolumn{1}{l}{Mem. $\downarrow$} & \multicolumn{1}{l|}{FPS $\uparrow$}     & \multicolumn{1}{l}{GPU-hr $\downarrow$} & \multicolumn{1}{l}{Mem. $\downarrow$} & \multicolumn{1}{l|}{FPS $\uparrow$}     & \multicolumn{1}{l}{GPU-hr $\downarrow$} & \multicolumn{1}{l}{Mem. $\downarrow$} & \multicolumn{1}{l}{FPS $\uparrow$}     \\ \hline
3DGS~\cite{kerbl20233d}              & {\color[HTML]{212121} 0.54}                         & 763MB                                 & 224                                     & 1.07                                    & 222MB                                 & 559                                     & 0.34                                    & 563MB                                 & 560                                    \\
StopThePop~\cite{radl2024stopthepop} & \cellcolor[HTML]{FFFFFF}{\color[HTML]{212121} 0.60} & 780MB                                 & 180                                     & 0.24                                    & 223MB                                 & 403                                     & 0.39                                    & 549MB                                 & 179                                    \\
3DGRT~\cite{moenne20243d}            & 0.83$\textsuperscript{\textdagger}$                 & 383MB                                 & 156$\textsuperscript{\textdagger}$      & -                                       & -                                     & -                                       & 0.83$\textsuperscript{\textdagger}$     & 388MB                                 & 190$\textsuperscript{\textdagger}$     \\
SMERF~\cite{duckworth2024smerf}      & 272                                                 & 139MB                                 & 454$\textsuperscript{\textdagger}$      & 528                                     & 4108MB                                & 408$\textsuperscript{\textdagger}$      & -                                       & -                                     & -                                      \\
Our model                            & \cellcolor[HTML]{FFFFFF}{\color[HTML]{212121} 1.04} & 1134MB                                & 36                                      & 1.26                                    & 1694MB                                & 24                                      & 0.86                                    & 1173MB                                & 44                                     \\ \hline
ZipNeRF~\cite{barron2023zip}         & \cellcolor[HTML]{FFFFFF}{\color[HTML]{212121} 32}   & 903MB                                 & ~0.5                                    & 48                                      & 903M                                  & ~0.5                                    & -                                       & -                                     & -                                     
\end{tabular}
    }
    \caption{Results on the 9 scenes from Mip-NeRF360~\cite{barron2021mip}, 4 large scenes from Zip-NeRF~\cite{barron2023zip} datasets, and 4 scenes from DeepBlending~\cite{hedman2018deep} and Tanks\&Temples~\cite{Knapitsch2017} combined.
    The increased sharpness and perceptual quality are reflected in the qualitative comparison in Fig.~\ref{fig:image_comparison}.
    ``GPU-hr'' indicates the number of GPU hours it takes to train a model. ``Mem.'' indicates the amount of memory the finished model consumes on disk. Variations in memory usage between 3DGS, StopThePop, 3DGRT, and our model are due to densification differences, as memory per primitive is the same.
    FPS results are computed at test resolution, and results with a $\textsuperscript{\textdagger}$ were not recomputed and use different high end GPUs.
    }
    \label{tab:main_results}
\end{table*}


We evaluate \acronym{} on the 9 scenes introduced in Mip-NeRF 360~\cite{barron2022mip}, the 4 scenes from DeepBlending~\cite{hedman2018deep} and Tanks\&Temples~\cite{Knapitsch2017}, and the 4 scenes introduced in Zip-NeRF~\cite{barron2023zip}. 
We use the same parameters across all scenes, except one: the size threshold for splitting and cloning is adjusted for large scenes to avoid total densification failure for both 3DGS and our method. For the \scenename{alameda} scene, we use the exposure trick introduced in~\cite{duckworth2024smerf} to allow evaluation on data sets with multiple exposures. We also re-tuned 3DGS to work slightly better than the tune done by Duckworth et al.~\cite{duckworth2024smerf} on the Zip-NeRF dataset.

We evaluate \acronym{} on the 9 scenes introduced in Mip-NeRF 360~\cite{barron2022mip}, the 4 scenes from DeepBlending~\cite{hedman2018deep} and Tanks\&Temples~\cite{Knapitsch2017}, and the 4 scenes introduced in Zip-NeRF~\cite{barron2023zip}. 
We use the same parameters across all scenes, except one: the size threshold for splitting and cloning is adjusted for large scenes to avoid total densification failure for both 3DGS and our method. For the \scenename{alameda} scene, we use the exposure trick introduced in~\cite{duckworth2024smerf} to allow evaluation on data sets with multiple exposures. We also re-tuned 3DGS to work slightly better than the tune done by Duckworth et al.~\cite{duckworth2024smerf} on the Zip-NeRF dataset.

There have been many follow-ups to 3DGS, each introducing different kinds of improvement, so we limit the scope of our comparison to best isolate our contribution. There are many recent approaches for densification~\cite{bulo2024revising, ververas2024sags, cao2024lightweight, ye2024absgs, yu2024gaussian}, but we chose to change densification only in ways required to handle density-based parameterization. It is likely that recent progress on improved 3DGS heuristics may similarly improve the performance of our model. There are also many different ways to change rendering~\cite{geiger2023mip, radl2024stopthepop, blanc2024, zhou2024unified, huang20242d, kheradmand20243d}, so we chose to only compare rendering methods that specifically address popping.


\newcommand{\graphimsix}[1]{
\begin{tikzpicture}[zoomboxarray, zoomboxarray rows=1, zoomboxes below, zoomboxarray inner gap=0.0cm]
    \node [image node] { \includegraphics[width=\figsixwidth,valign=b]{#1} };
    \zoombox[magnification=4]{0.40,0.60}
\end{tikzpicture}
}

\newcommand{\graphimseven}[1]{
\begin{tikzpicture}[zoomboxarray, zoomboxarray rows=2, zoomboxes below, zoomboxarray inner gap=0.0cm]
    \node [image node] { \includegraphics[width=\figninewidth,valign=b]{#1} };
    \zoombox[magnification=6]{0.32,0.85}
    \zoombox[magnification=10]{0.87,0.36}
\end{tikzpicture}
}

\newcommand{\graphimnine}[1]{
\begin{tikzpicture}[zoomboxarray, zoomboxarray rows=1, zoomboxes below, zoomboxarray inner gap=0.0cm]
    \node [image node] { \includegraphics[width=3.5cm,valign=b]{#1} };
    \zoombox[magnification=10]{0.5,0.5}
\end{tikzpicture}
}

\newcommand{\graphimtwo}[1]{
\begin{tikzpicture}[zoomboxarray, zoomboxarray rows=1, zoomboxes below, zoomboxarray inner gap=0.0cm]
    \node [image node] { \includegraphics[width=3.5cm,valign=b]{#1} };
    \zoombox[magnification=4]{0.45,0.30}
\end{tikzpicture}
}

\newcommand{\graphimeleven}[1]{
\begin{tikzpicture}[zoomboxarray, zoomboxarray rows=1, zoomboxes below, zoomboxarray inner gap=0.0cm]
    \node [image node] { \includegraphics[width=\figsixwidth,valign=b]{#1} };
    \zoombox[magnification=3]{0.20,0.60}
\end{tikzpicture}
}
\newcommand{\graphimtwelve}[1]{
\begin{tikzpicture}[zoomboxarray, zoomboxarray rows=1, zoomboxes below, zoomboxarray inner gap=0.0cm]
    \node [image node] { \includegraphics[width=\figsixwidth,valign=b]{#1} };
    \zoombox[magnification=4]{0.27,0.23}
\end{tikzpicture}
}
\newcommand{\graphimthirteen}[1]{
\begin{tikzpicture}[zoomboxarray, zoomboxarray rows=1, zoomboxes below, zoomboxarray inner gap=0.0cm]
    \node [image node] { \includegraphics[width=\figsixwidth,valign=b]{#1} };
    \zoombox[magnification=3]{0.45,0.45}
\end{tikzpicture}
}
\newcommand{\graphimfourteen}[1]{
\begin{tikzpicture}[zoomboxarray, zoomboxarray rows=1, zoomboxes below, zoomboxarray inner gap=0.0cm]
    \node [image node] { \includegraphics[width=\figninewidth,valign=b]{#1} };
    \zoombox[magnification=6]{0.51,0.55}
\end{tikzpicture}
}
\newcommand{\graphimfifteen}[1]{
\begin{tikzpicture}[zoomboxarray, zoomboxarray rows=1, zoomboxes below, zoomboxarray inner gap=0.0cm]
    \node [image node] { \includegraphics[width=\figsixwidth,valign=b]{#1} };
    \zoombox[magnification=4]{0.45,0.30}
\end{tikzpicture}
}

\newcommand{\graphimsixteen}[1]{
\begin{tikzpicture}[zoomboxarray, zoomboxarray rows=1, zoomboxes below, zoomboxarray inner gap=0.0cm]
    \node [image node] { \includegraphics[width=\figsixwidth cm,valign=b]{#1} };
    \zoombox[magnification=4]{0.45,0.30}
\end{tikzpicture}
}

\newcommand{\graphimseventeen}[1]{
\begin{tikzpicture}[zoomboxarray, zoomboxarray rows=1, zoomboxes below, zoomboxarray inner gap=0.0cm]
    \node [image node] { \includegraphics[width=4cm,valign=b]{#1} };
    \zoombox[magnification=4]{0.13,0.27}
\end{tikzpicture}
}

\newcommand{\graphimone}[1]{
\begin{tikzpicture}[zoomboxarray, zoomboxarray rows=1, zoomboxes below, zoomboxarray inner gap=0.0cm]
    \node [image node] { \includegraphics[width=\figninewidth,valign=b]{#1} };
    \zoombox[magnification=4]{0.225,0.55}
\end{tikzpicture}
}

\newcommand{\graphimthree}[1]{
\begin{tikzpicture}[zoomboxarray, zoomboxarray rows=1, zoomboxes below, zoomboxarray inner gap=0.0cm]
    \node [image node] { \includegraphics[width=\figninewidth,valign=b]{#1} };
    \zoombox[magnification=4]{0.15,0.25}
\end{tikzpicture}
}
\newcommand{\graphimfour}[1]{
\begin{tikzpicture}[zoomboxarray, zoomboxarray rows=1, zoomboxes below, zoomboxarray inner gap=0.0cm]
    \node [image node] { \includegraphics[width=\figninewidth,valign=b]{#1} };
    \zoombox[magnification=8]{0.85,0.83}
\end{tikzpicture}
}
\newcommand{\graphimfive}[1]{
\begin{tikzpicture}[zoomboxarray, zoomboxarray rows=2, zoomboxes below, zoomboxarray inner gap=0.0cm]
    \node [image node] { \includegraphics[width=\figninewidth,valign=b]{#1} };
    \zoombox[magnification=4]{0.35,0.90}
    \zoombox[magnification=4]{0.75,0.87}
\end{tikzpicture}
}

\newcommand{\graphimeight}[1]{
\begin{tikzpicture}[zoomboxarray, zoomboxarray rows=1, zoomboxes below, zoomboxarray inner gap=0.0cm]
    \node [image node] { \includegraphics[width=\figninewidth,valign=b]{#1} };
    \zoombox[magnification=8]{0.13,0.37}
\end{tikzpicture}
}

\begin{figure*}[ht]
\centering
\bgroup
\begin{tabular}{c@{}c@{}c@{}c@{}c}
\graphimfour{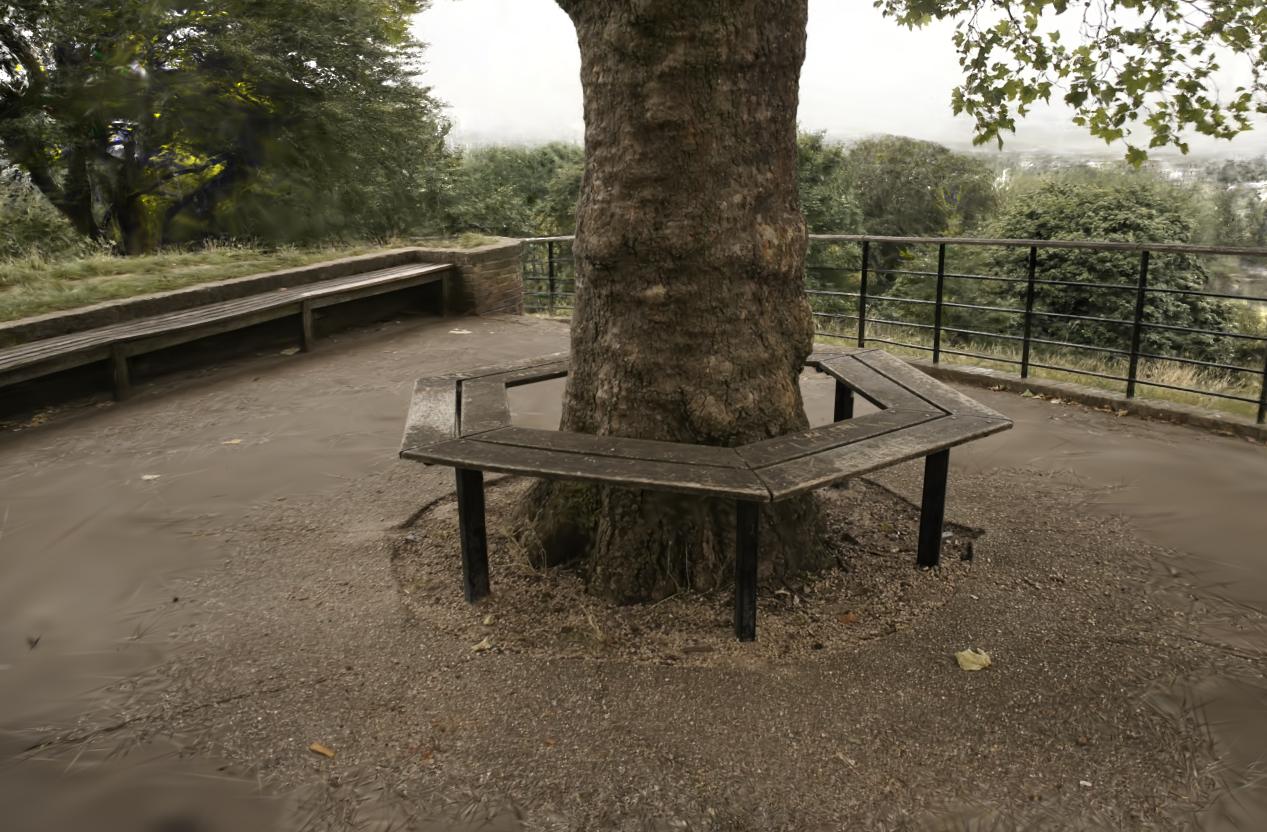} &
\graphimfour{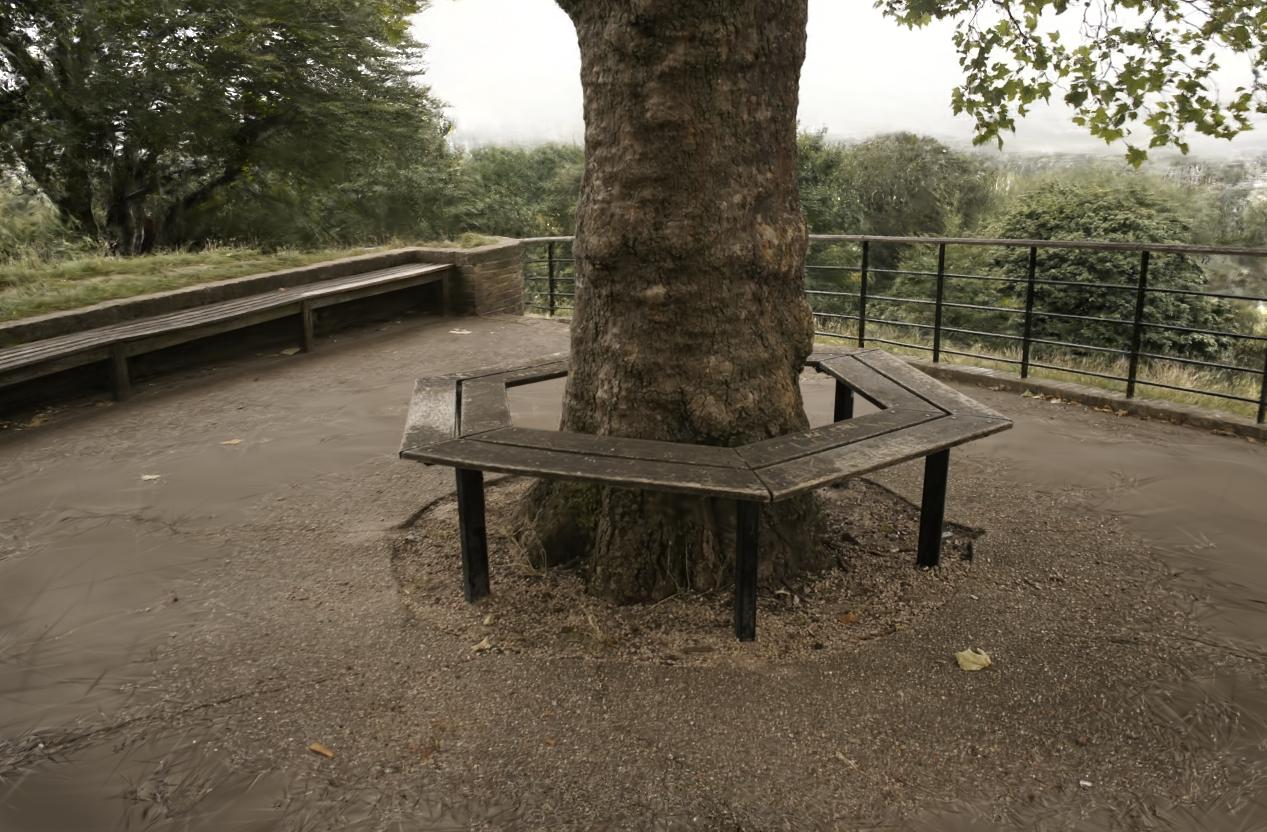} &
\graphimfour{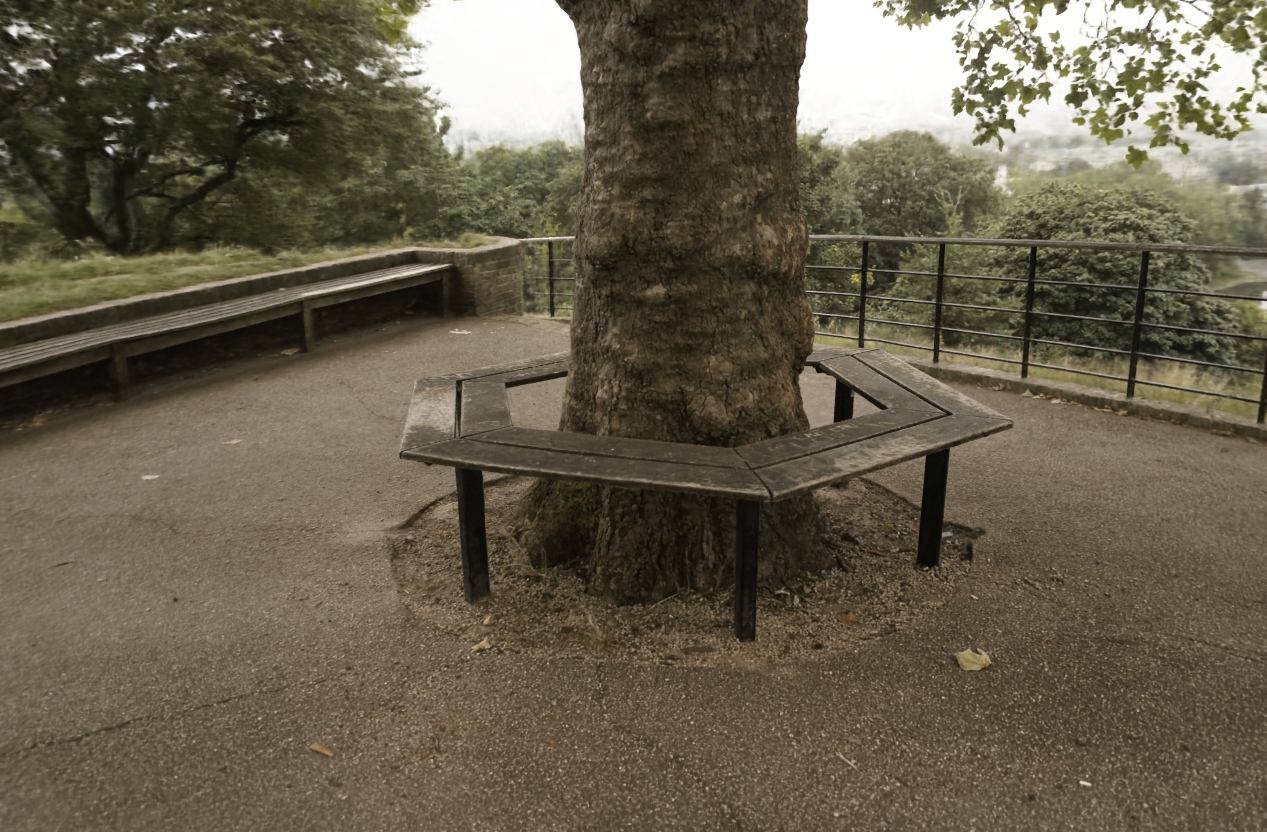} &
\graphimfour{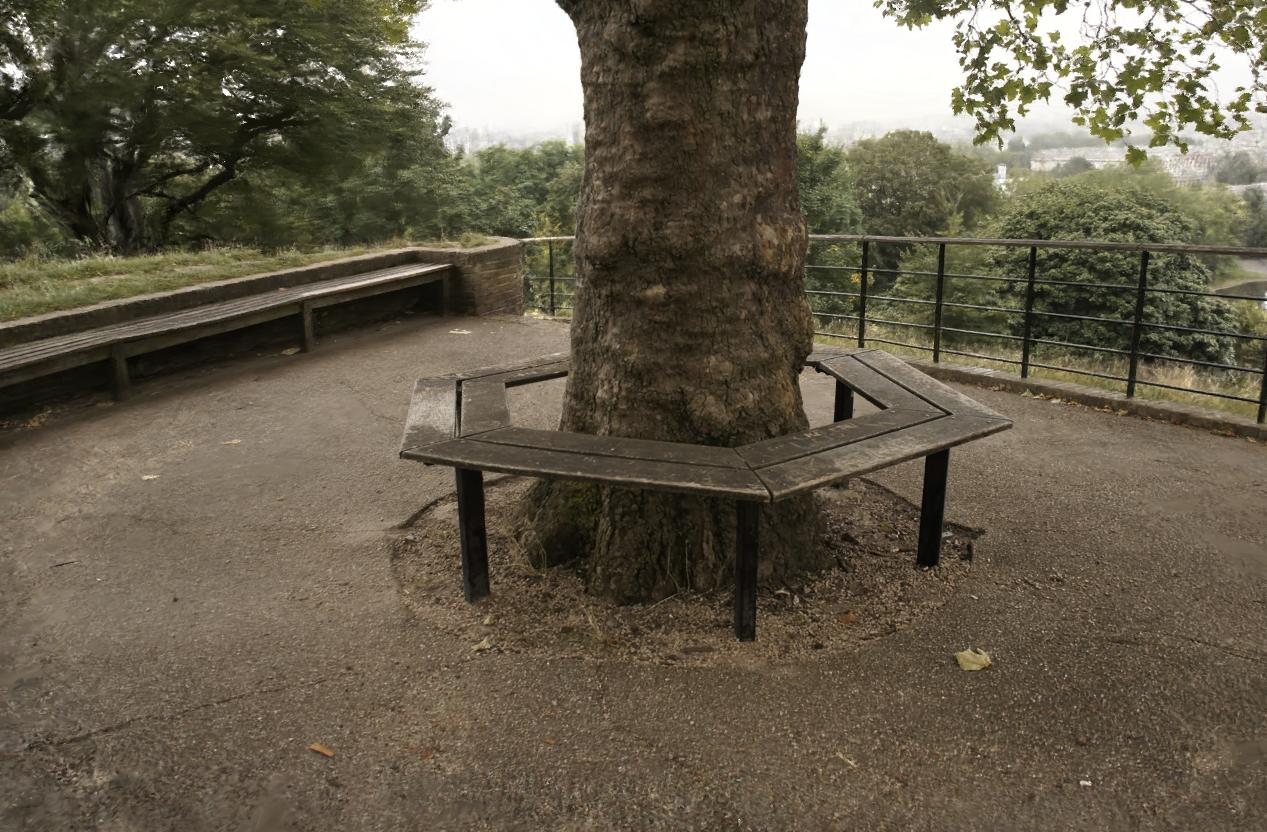} &
\graphimfour{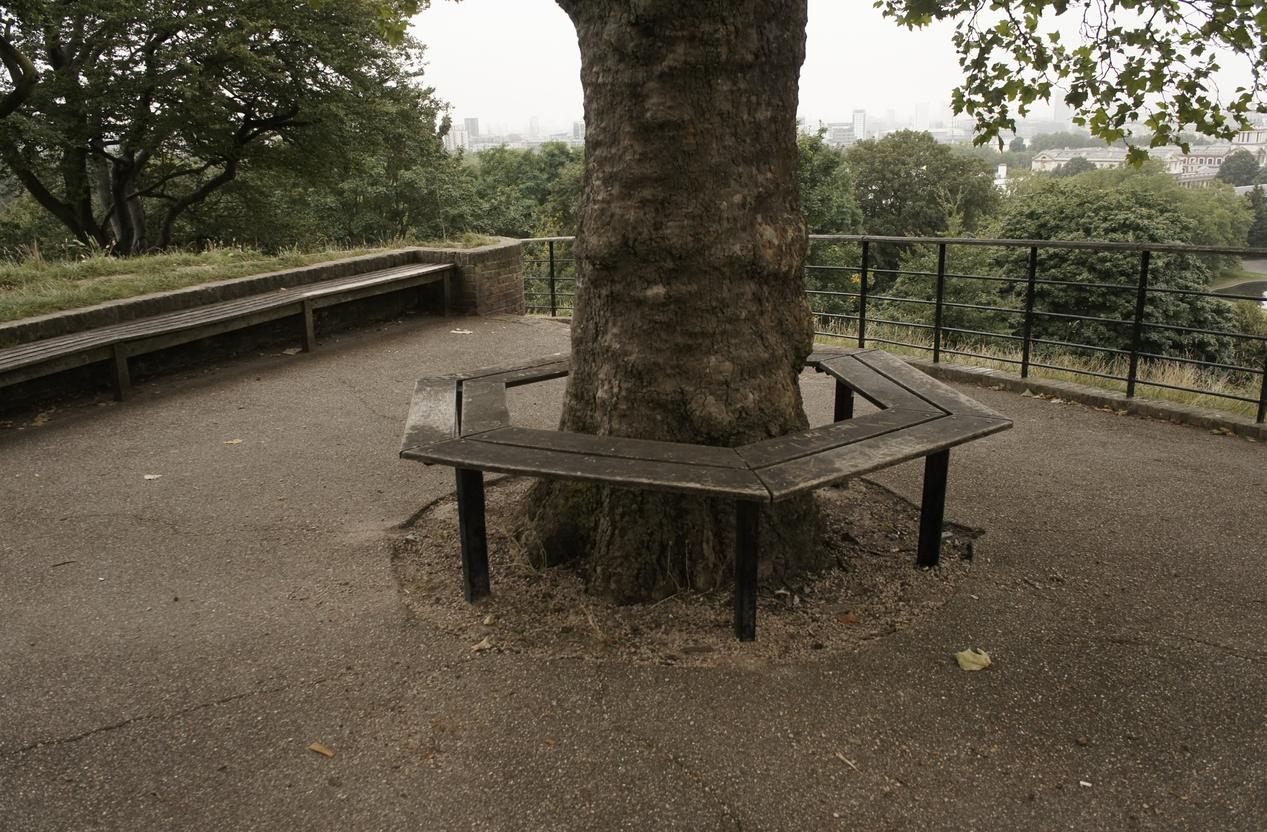} \\
\graphimeight{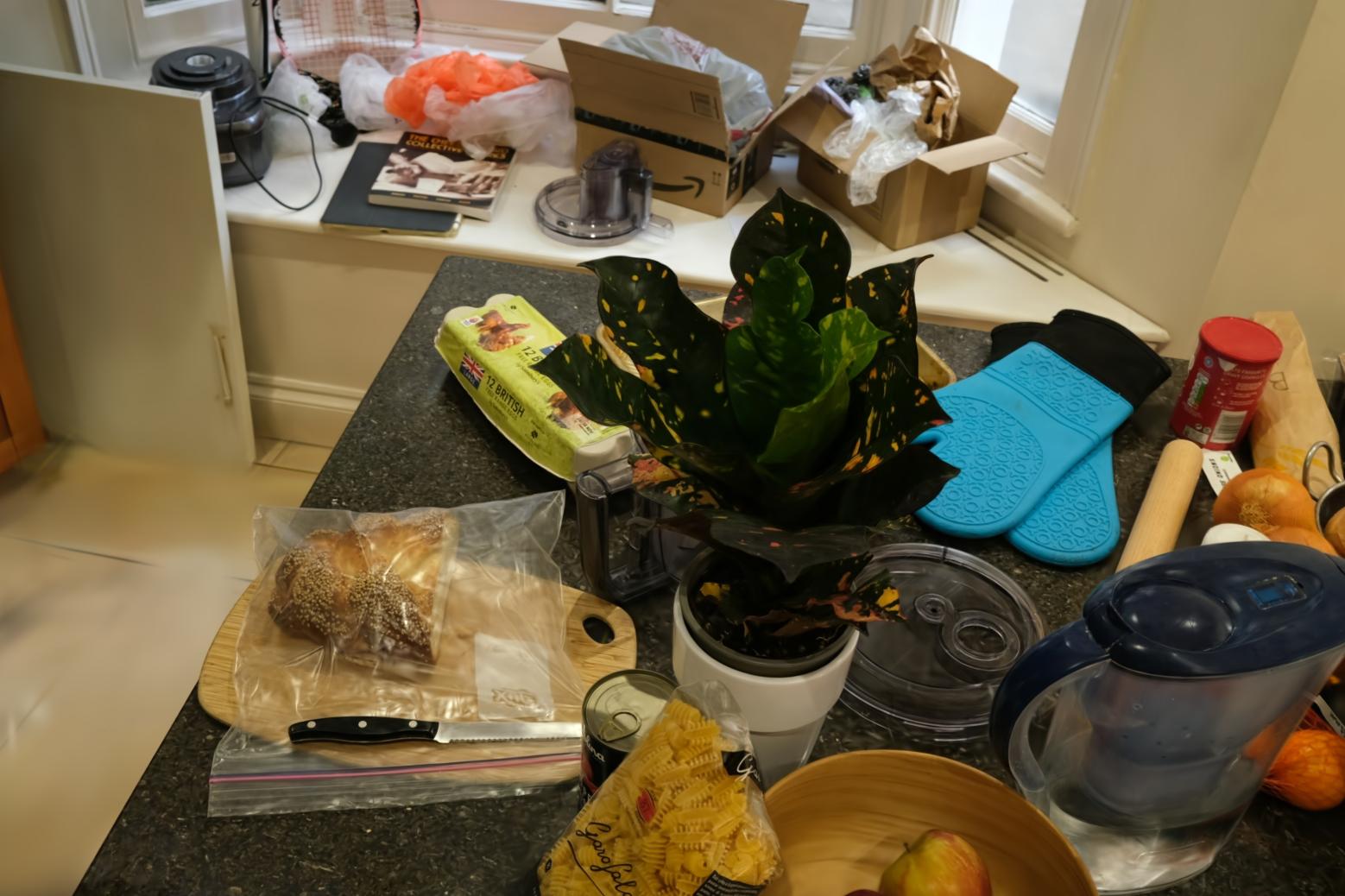} &
\graphimeight{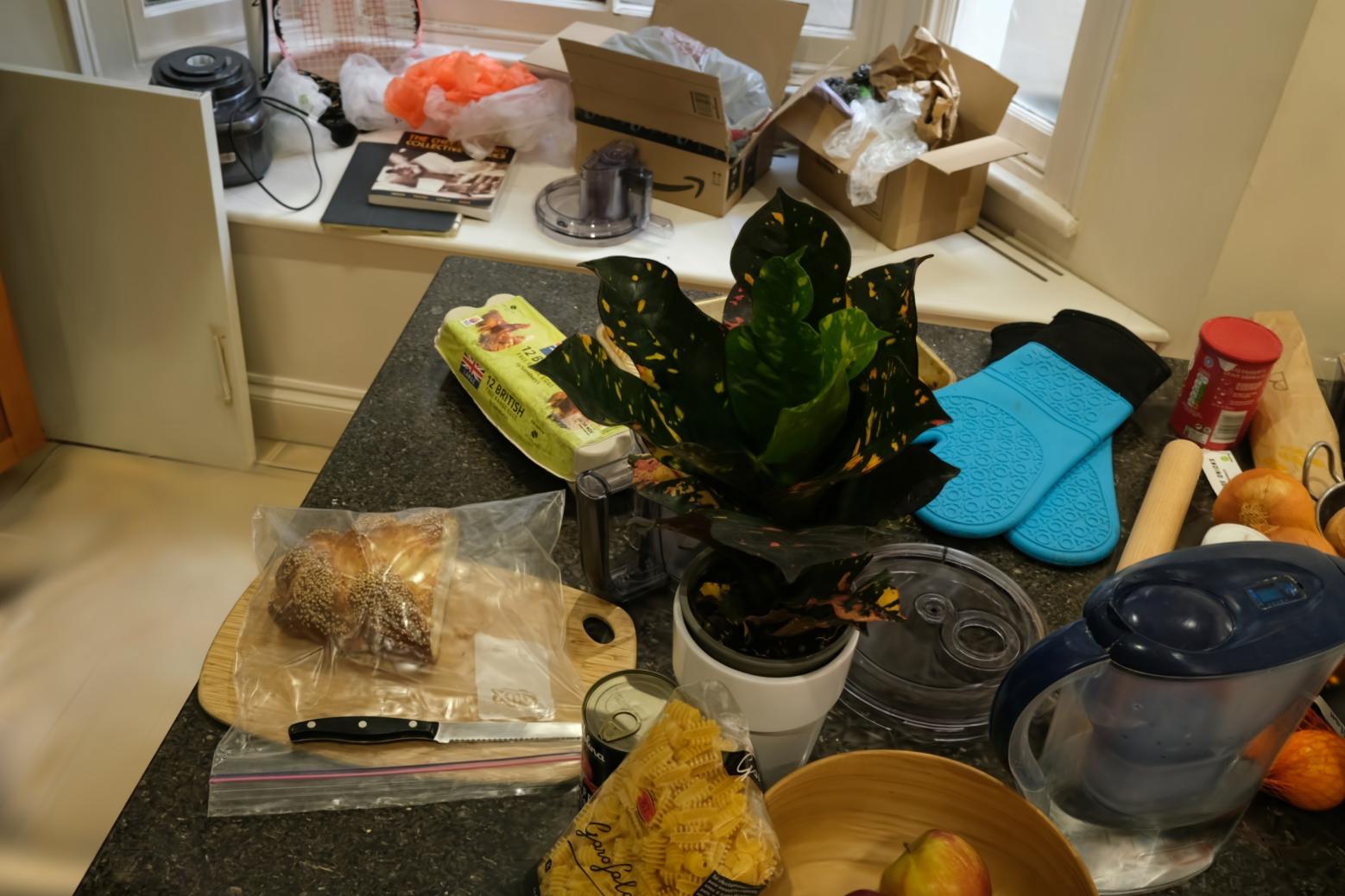} &
\graphimeight{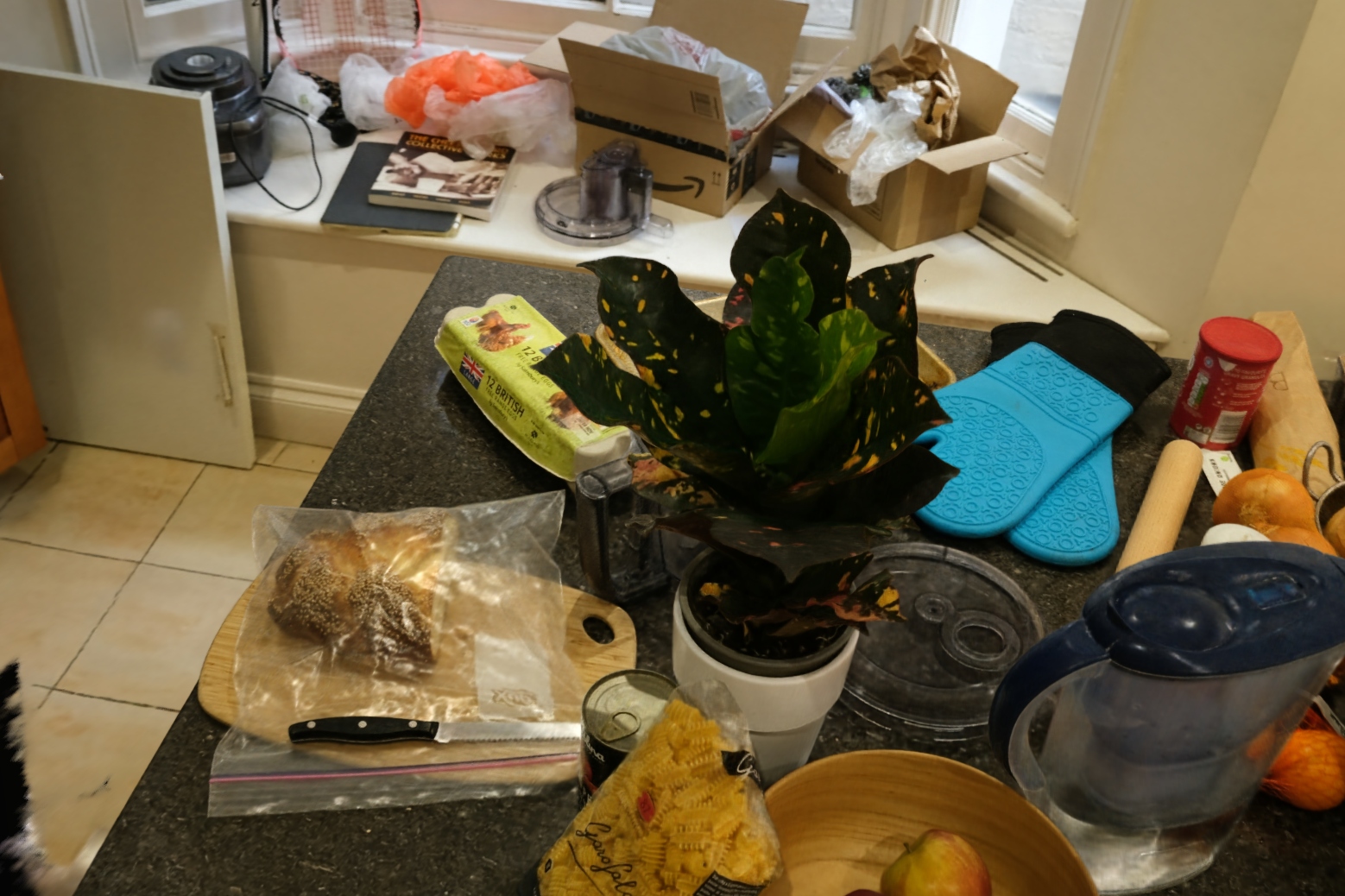} &
\graphimeight{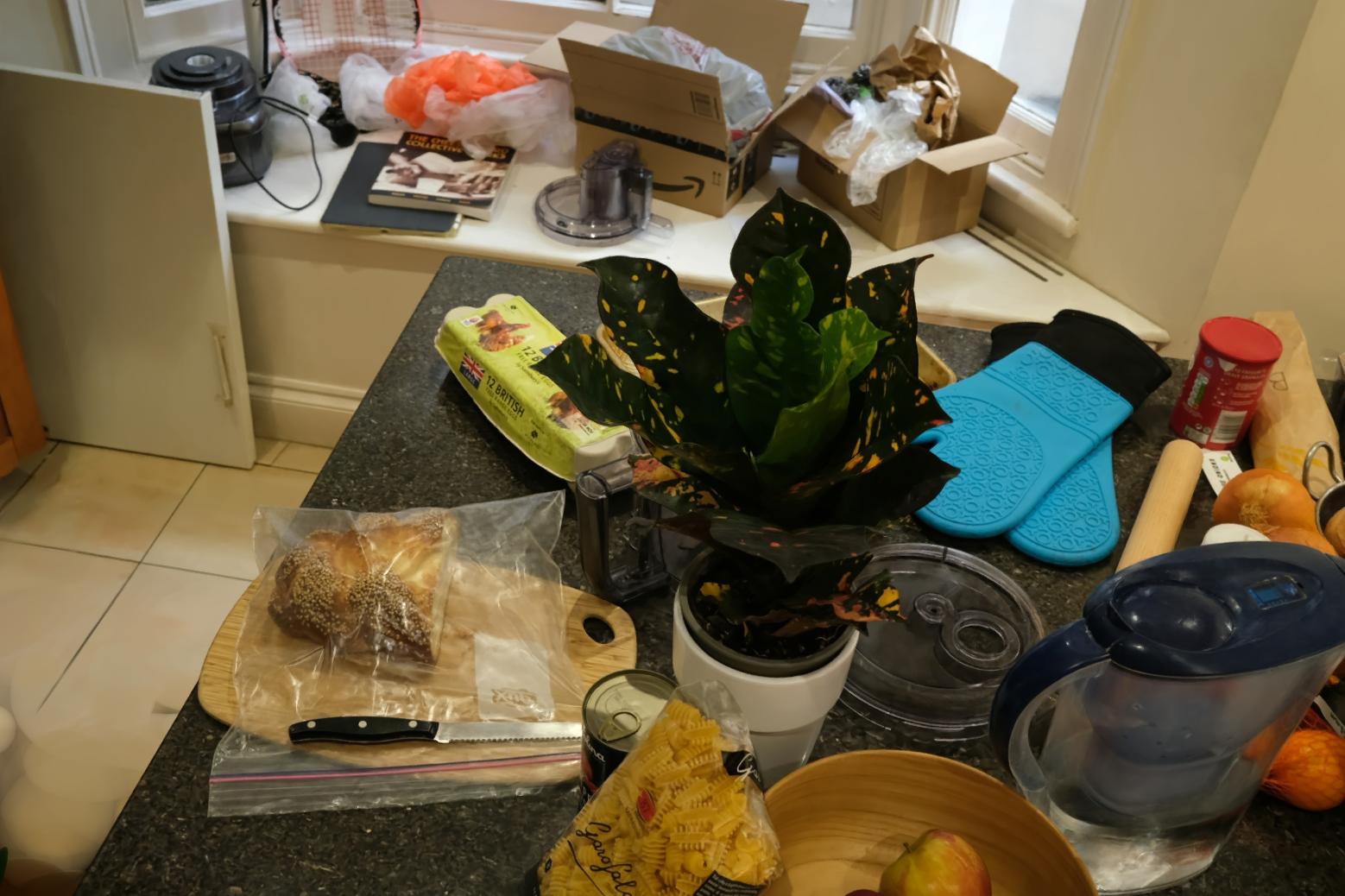} &
\graphimeight{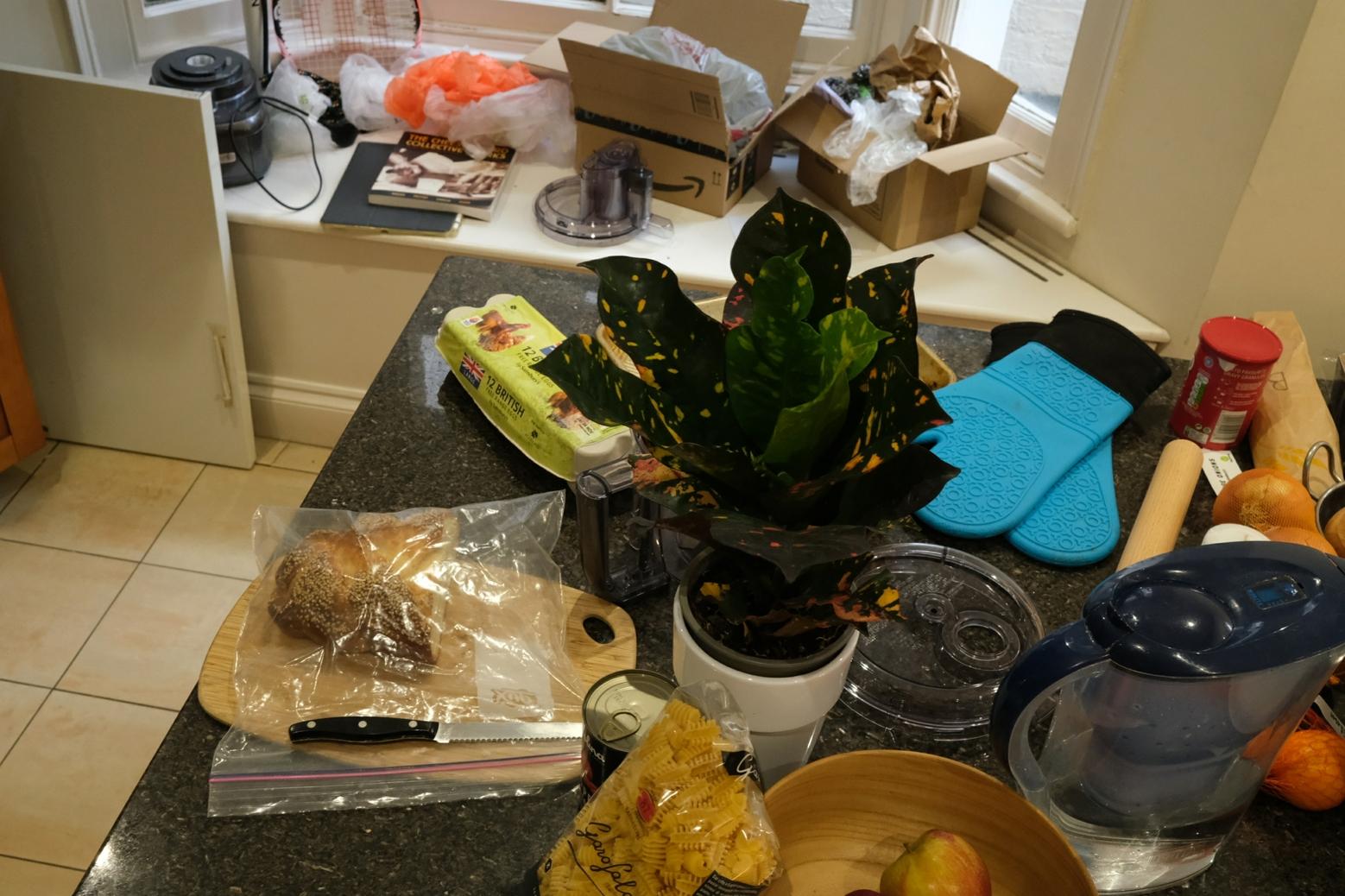} \\
\graphimsix{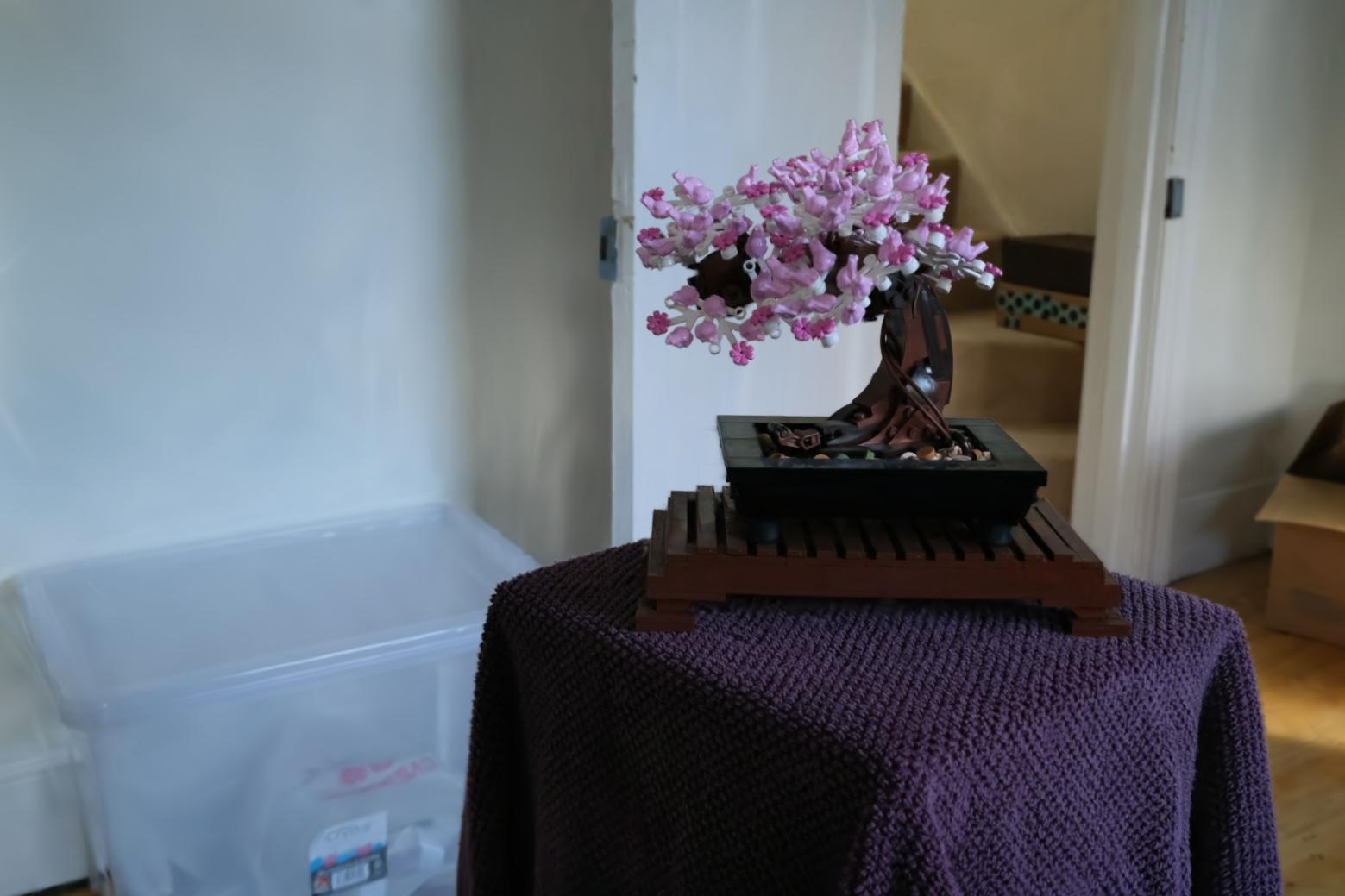} &
\graphimsix{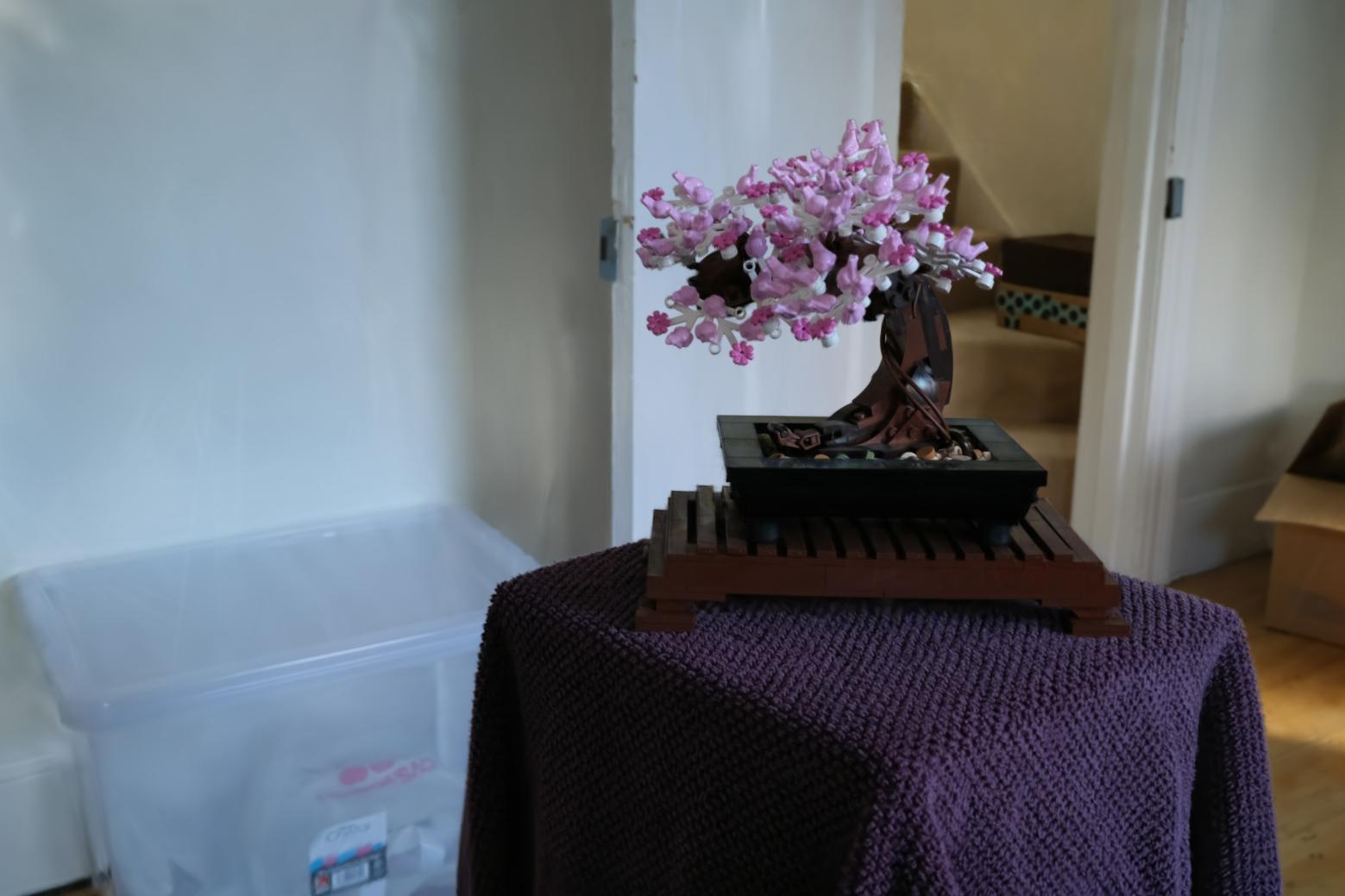} &
\graphimsix{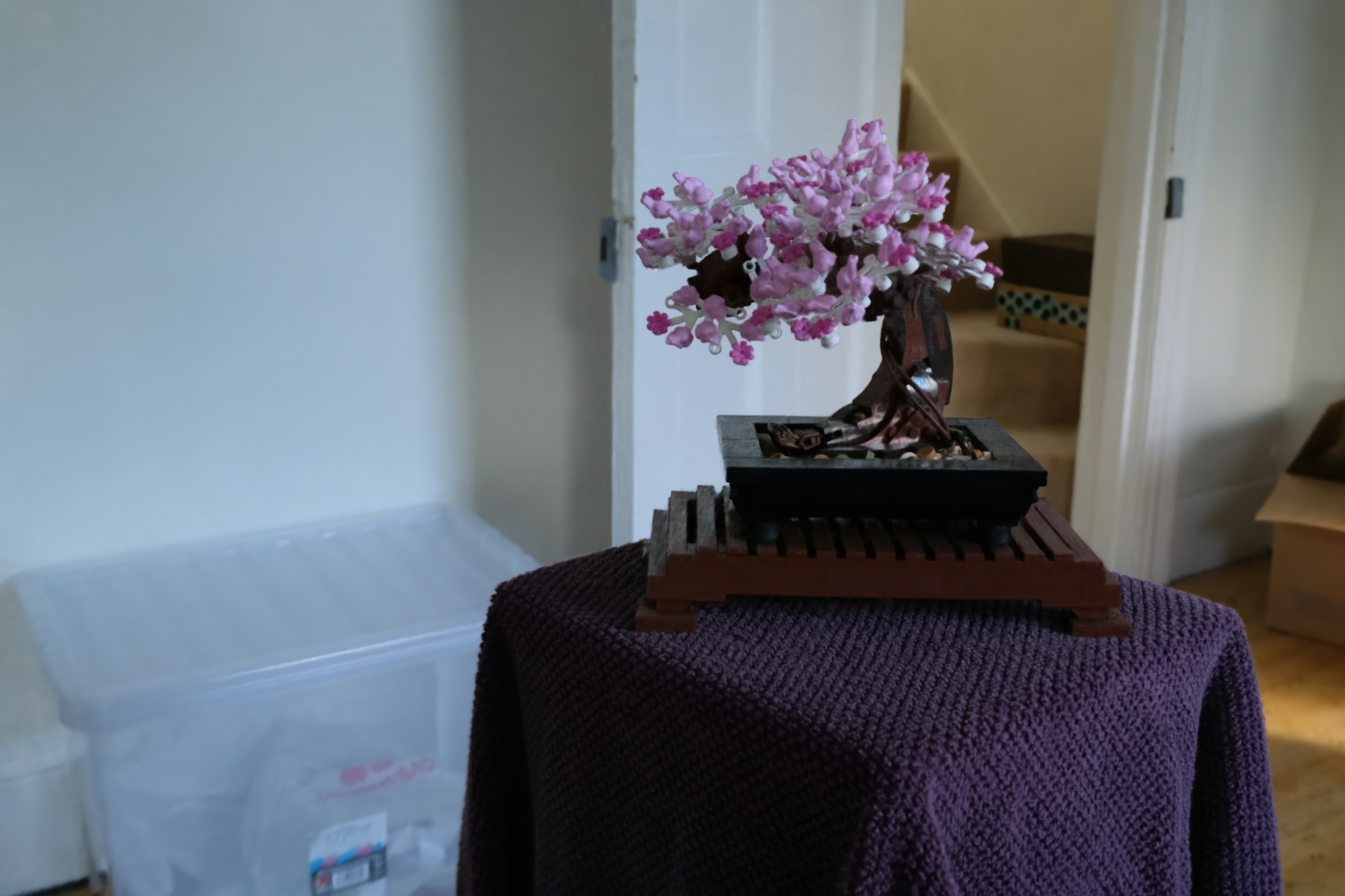} &
\graphimsix{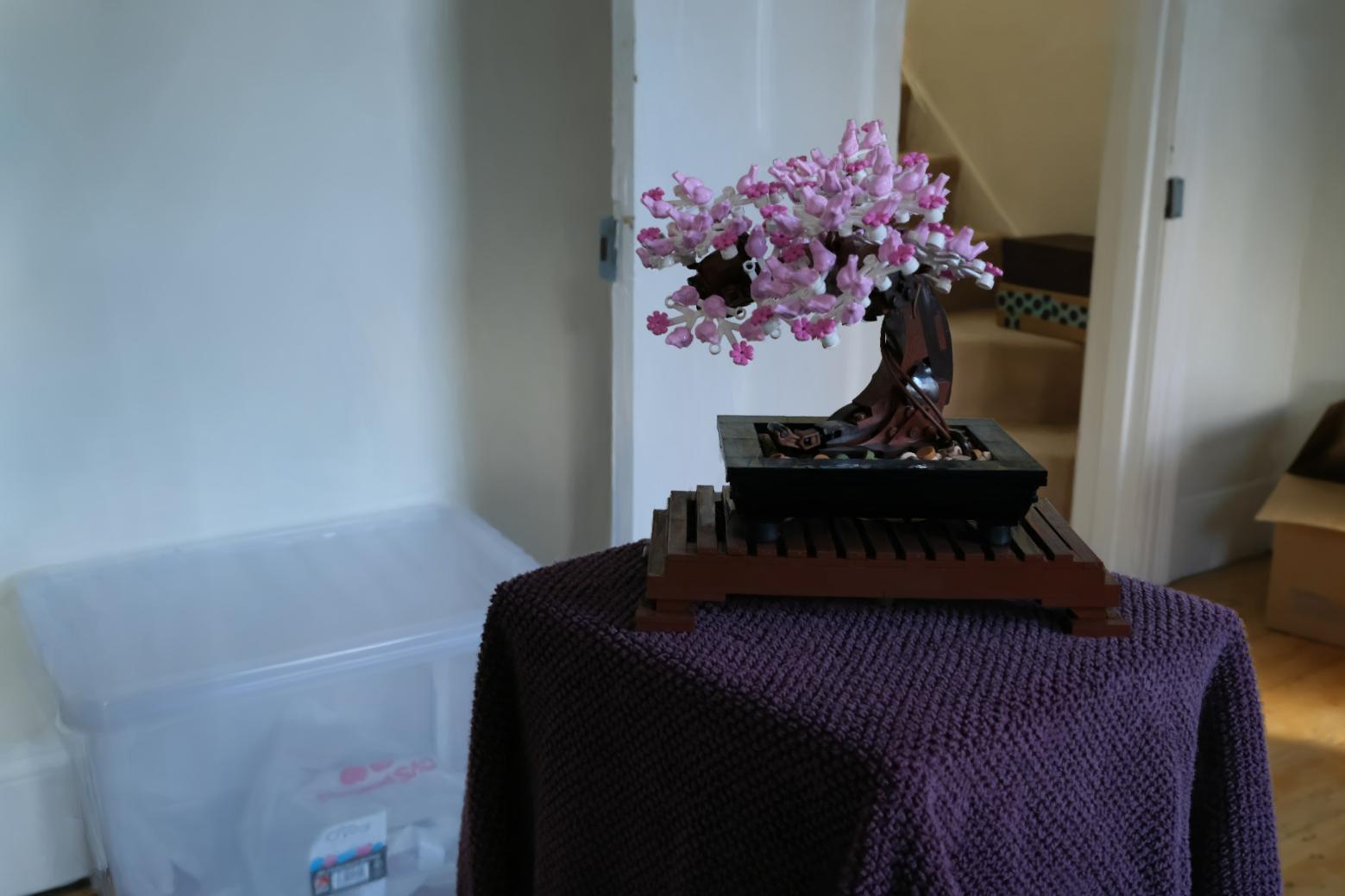} &
\graphimsix{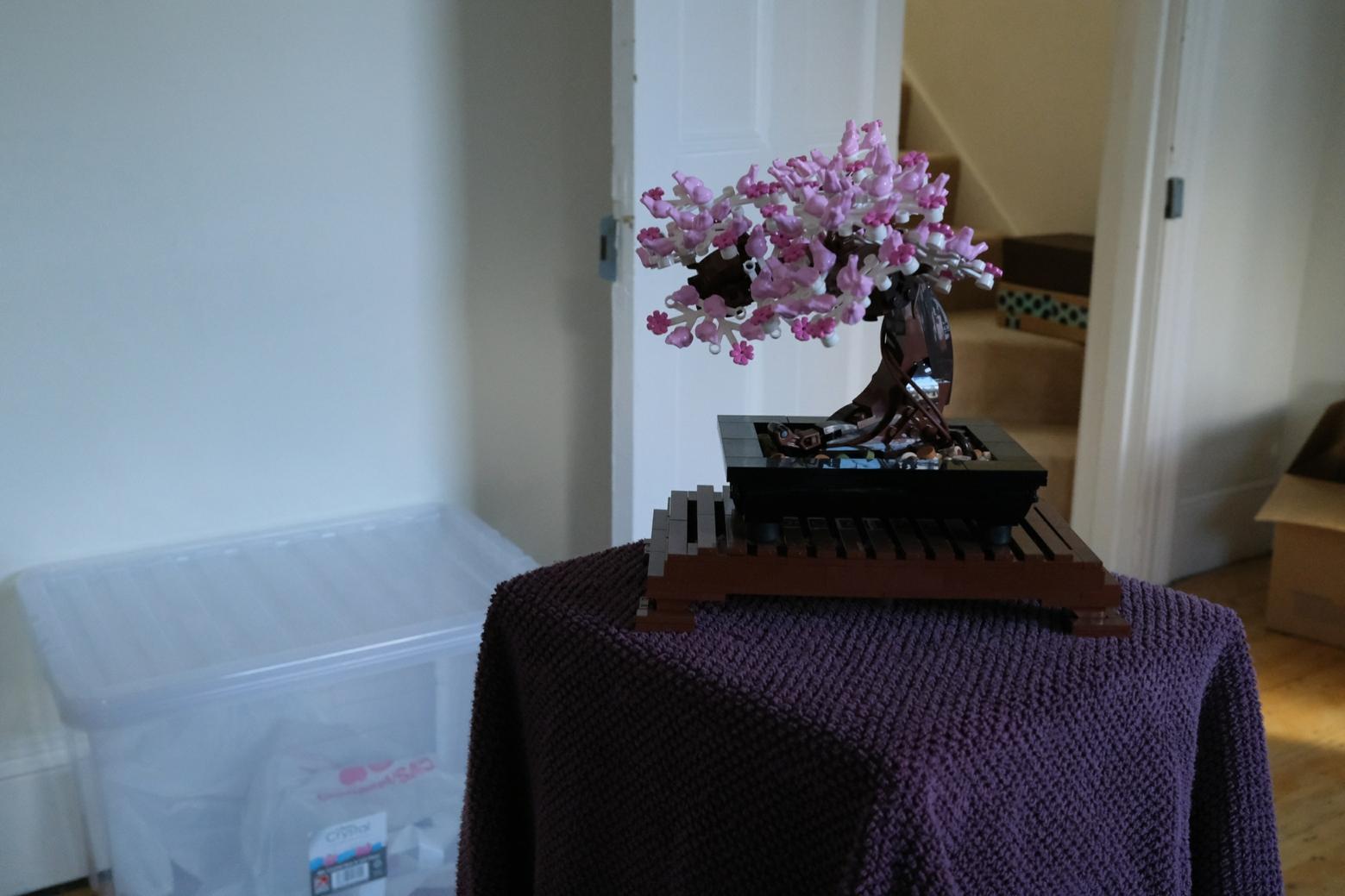} \\
\graphimeleven{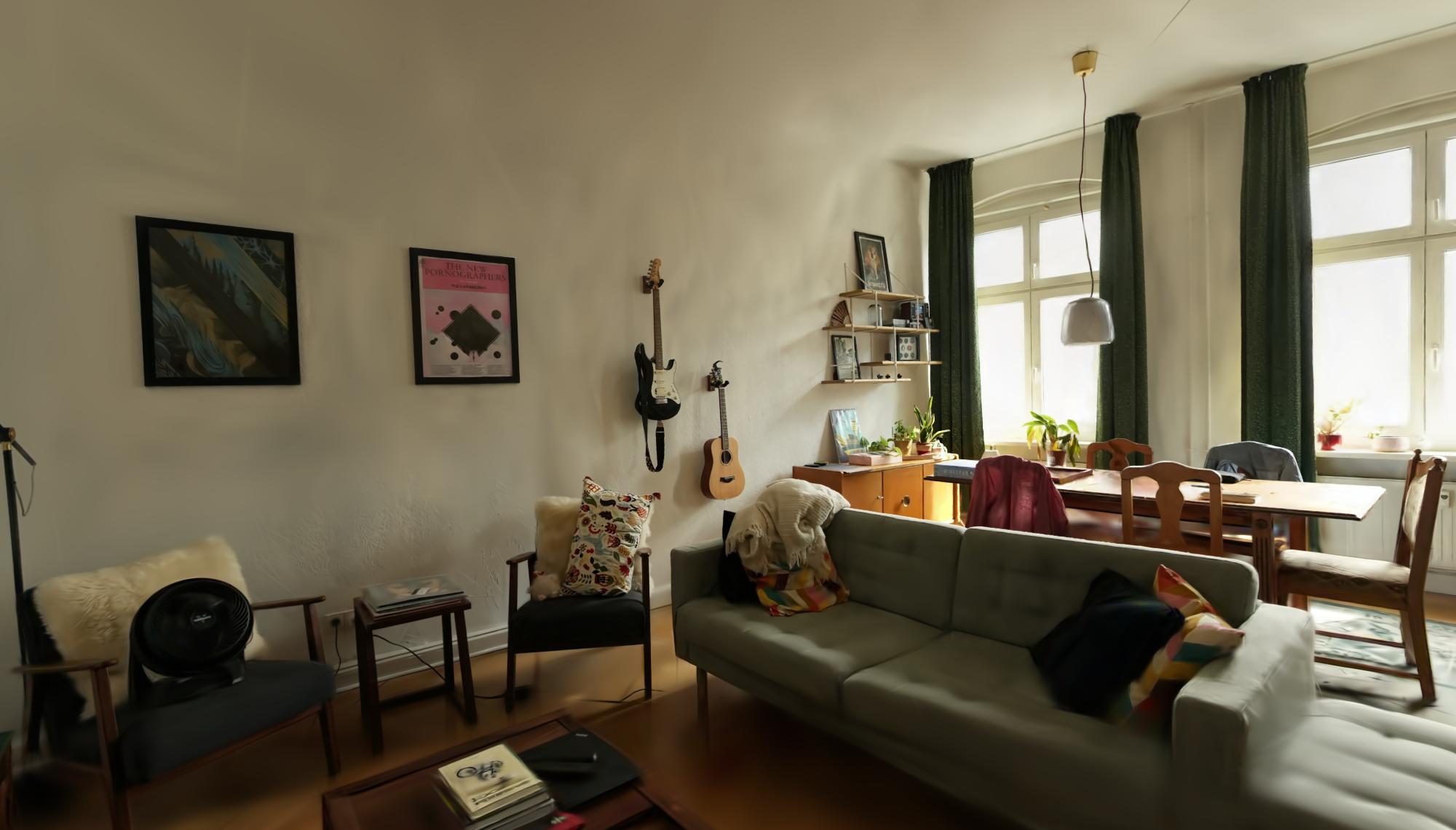} &
\graphimeleven{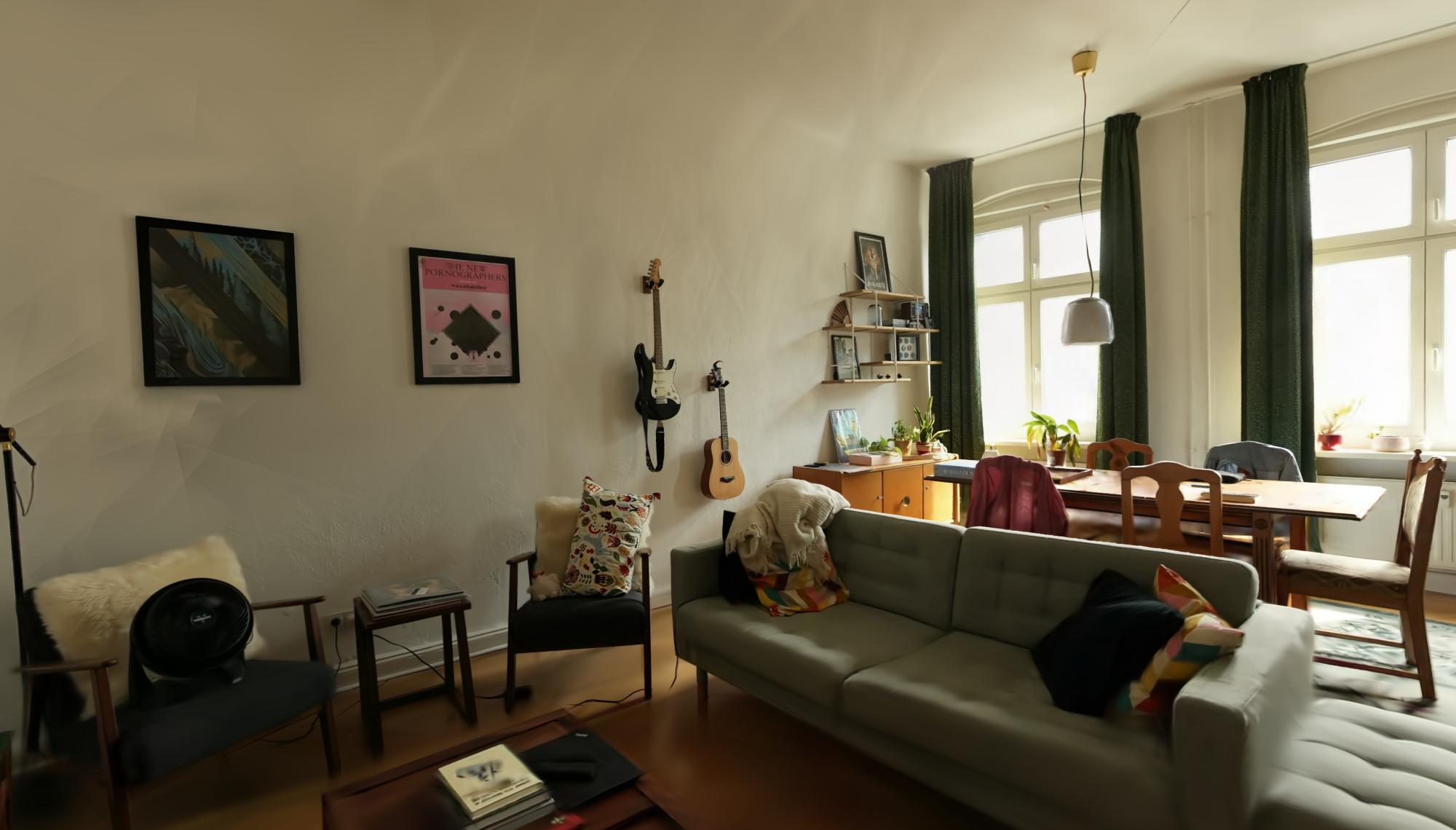} &
\graphimeleven{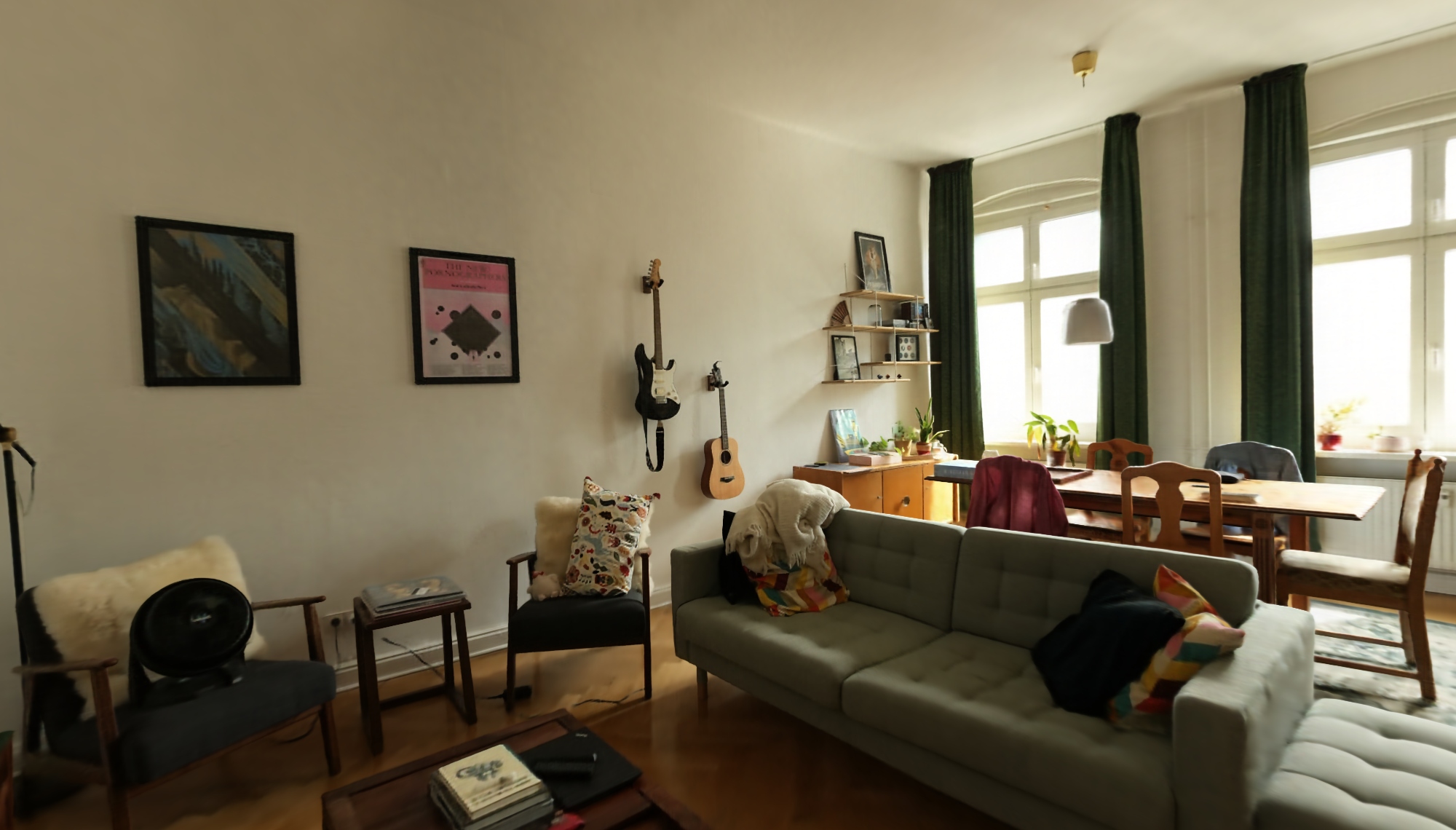} &
\graphimeleven{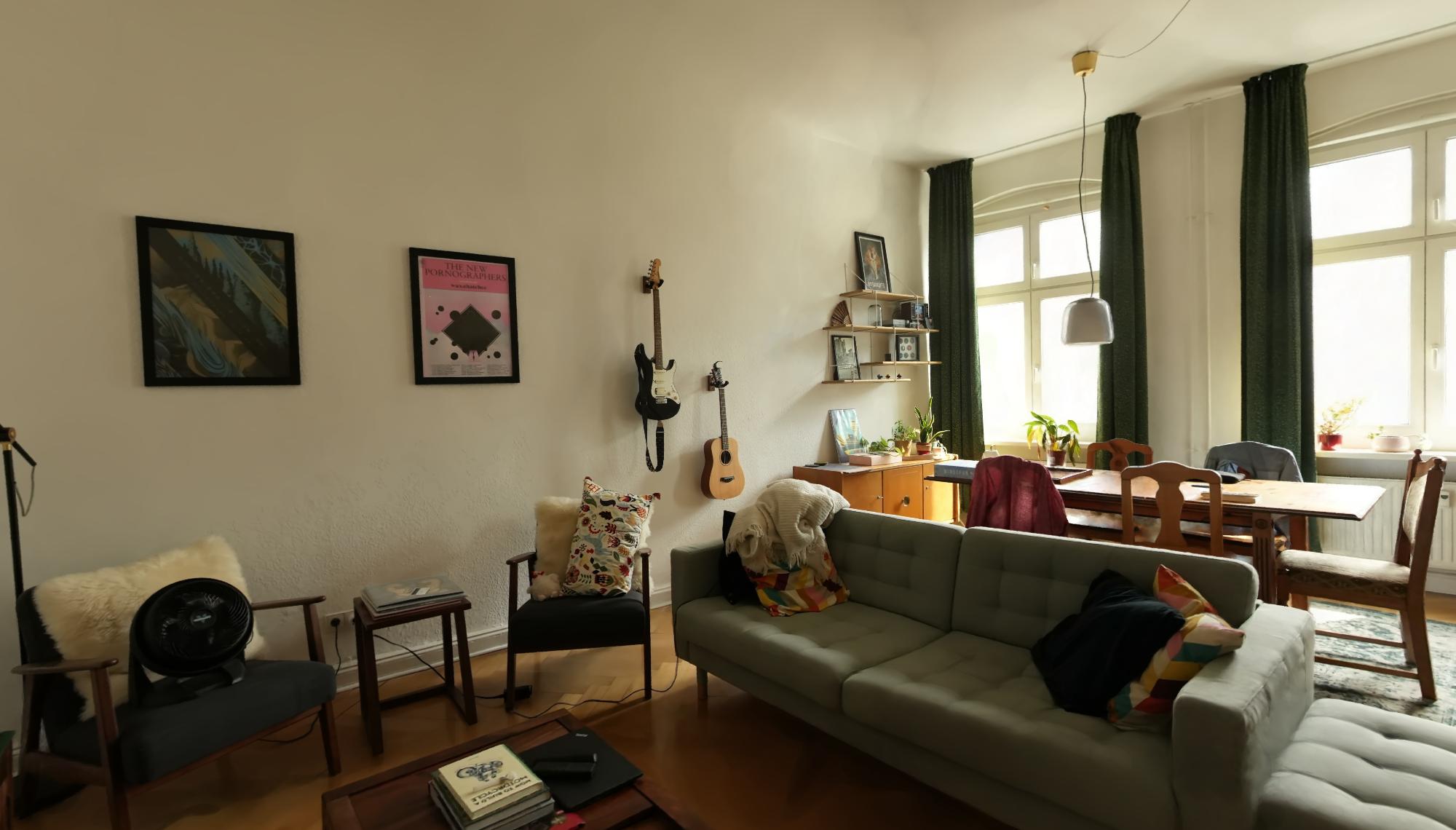} &
\graphimeleven{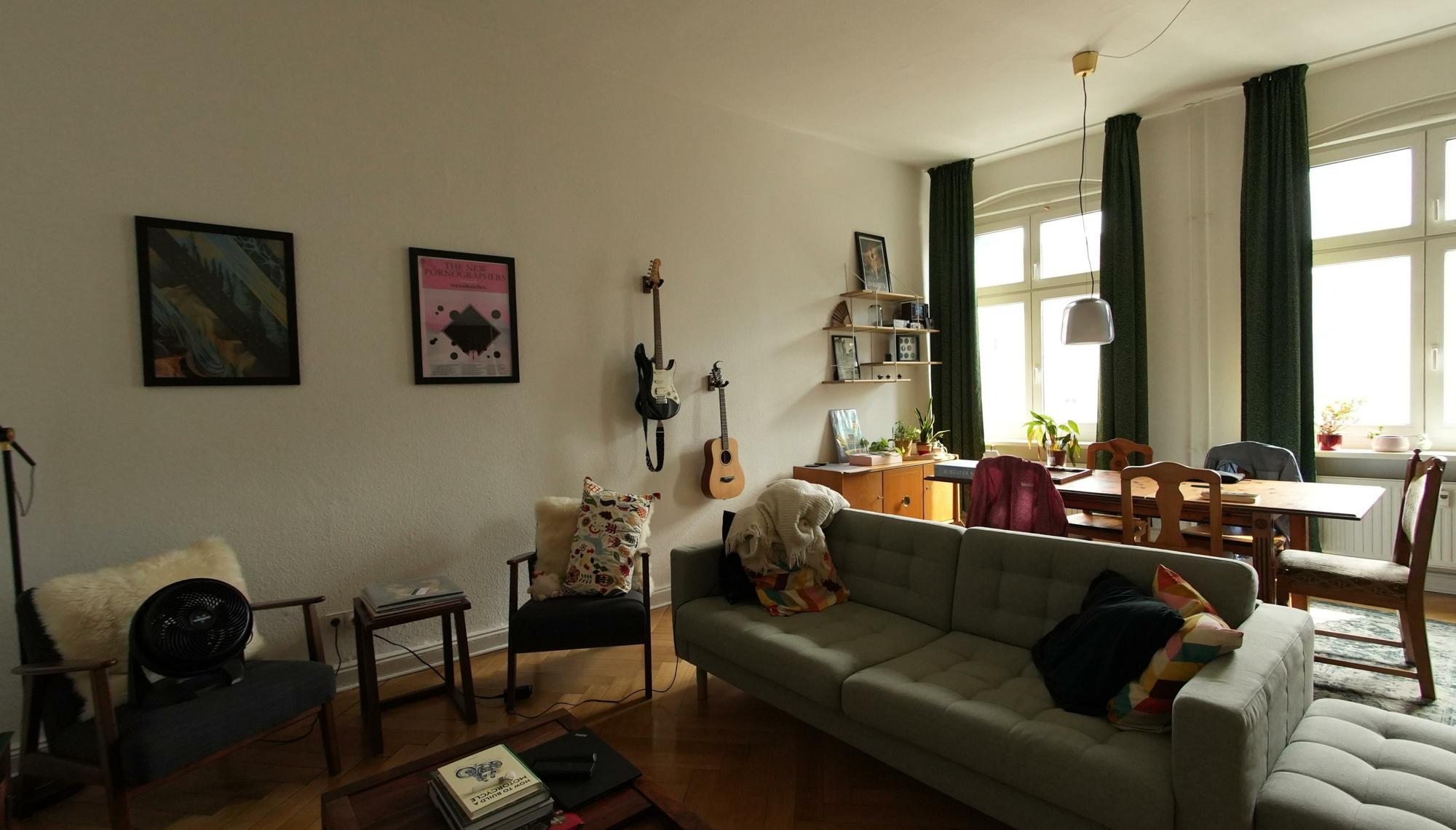} \\
3DGS & StopThePop & SMERF & Ours & GT
\end{tabular}
\egroup
\caption{Visual comparison of our method on the Mip-NeRF~\cite{barron2021mip} and Zip-NeRF~\cite{barron2023zip} datasets. Rows 1 and 2 show regions where our results are sharper. Rows 3 and 4 show how proper blending of primitives helps with handling lighting on textureless surfaces. 3DGRT is omitted due to missing Zip-NeRF results.}
\label{fig:image_comparison}
\end{figure*}

\subsection{Reconstruction Quality}
We measure image quality through the standard image metrics PSNR, SSIM and LPIPS.
We compare our method with 3DGS~\cite{kerbl20233d}, and with Zip-NeRF~\cite{barron2023zip}, the current offline rendering state-of-the-art for these datasets. We also compare with StopThePop since it improves view-consistency for 3DGS, 3DGRT~\cite{moenne20243d} because our work uses their ray tracing algorithm, and with SMERF~\cite{duckworth2024smerf}, because even though it is a distillation based method that takes 33 hours to train on 16xA100 GPUs compared to 1-2 hours for our method, it is a comparable real-time 3D consistent volume rendering method.

Quantitatively, we outperform the other 3DGS-based methods across the board, and set a new state of the art in the Zip-NeRF dataset in terms of sharpness as measured by LPIPS and SSIM --- even when compared to offline and distillation-based methods. 
Qualitatively, we notice two main factors: improved blending, and improved sharpness.
The combination of more flexible primitive appearance, combined with color blending, allows our method to represent gradients better, which can be seen in Figure~\ref{fig:alpha_profiles}.
This effect can also be seen when comparing how these methods model light falling on flat walls, as shown by the bottom two rows of Figure~\ref{fig:image_comparison}. 
While our method is able to reproduce the smooth image gradients in the shadowed areas, all of the Gaussian based methods struggle.
The improved sharpness is more apparent on the larger and more difficult Zip-NeRF dataset, both in the metrics and in qualitative examples (see the second two rows of Figure~\ref{fig:image_comparison}). 

\begin{figure*}[ht]
\includegraphics[width=\linewidth]{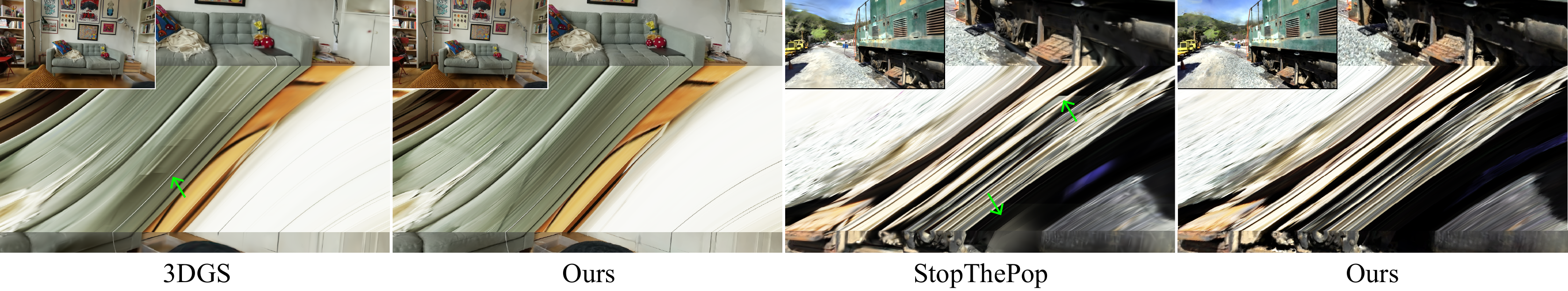}
\caption{
Here we show epipolar plane images \cite{bolles1987epipolar} to visualize popping in 3DGS~\cite{kerbl20233d} and StopThePop~\cite{radl2024stopthepop}. We recorded sequences of images while rotating the camera in the \scenename{berlin}~\cite{barron2023zip} (left) and \scenename{train}~\cite{knapitsch2017tanks} (right) scenes. The region in the middle are the epipolar plane images, with green arrows highlighting discontinuities in the color caused by popping/blend order artifacts. Our method lacks these horizontal discontinuities. Contrast and brightness have been boosted here for visibility's sake. See the supplemental video for more.}
\label{fig:popping_train}
\end{figure*}

\newcommand{\graphablation}[1]{
\begin{tikzpicture}[zoomboxarray]
    \node [image node] { \includegraphics[width=4.15cm]{#1} };
    \zoombox[magnification=6]{0.85,0.93}
    \zoombox[magnification=4]{0.6,0.55}
\end{tikzpicture}
}

\subsection{Popping}

\begin{table}[b!]
\resizebox{\linewidth}{!}{
\large
\begin{tabular}{l|rr|rr|rr|rr}
StP Popping & \multicolumn{2}{l|}{DeepBlending}                                                      & \multicolumn{2}{l|}{M360 Indoor}                                                       & \multicolumn{2}{l|}{M360 Outdoor}                                                      & \multicolumn{2}{l}{Tanks \& Temples}                                                  \\
Metric      & \multicolumn{1}{l}{\reflectbox{F}LIP$_1$} & \multicolumn{1}{l|}{\reflectbox{F}LIP$_7$} & \multicolumn{1}{l}{\reflectbox{F}LIP$_1$} & \multicolumn{1}{l|}{\reflectbox{F}LIP$_7$} & \multicolumn{1}{l}{\reflectbox{F}LIP$_1$} & \multicolumn{1}{l|}{\reflectbox{F}LIP$_7$} & \multicolumn{1}{l}{\reflectbox{F}LIP$_1$} & \multicolumn{1}{l}{\reflectbox{F}LIP$_7$} \\ \hline
3DGS        & \cellcolor[HTML]{FFFFB4}0.0063            & \cellcolor[HTML]{FFFFB4}0.0122             & \cellcolor[HTML]{FFFFB4}0.0072            & \cellcolor[HTML]{FFFFB4}0.0149             & \cellcolor[HTML]{FFFFB4}0.0083            & \cellcolor[HTML]{FFFFB4}0.0154             & \cellcolor[HTML]{FFFFB4}0.0107            & \cellcolor[HTML]{FFFFB4}0.0315            \\
StopThePop  & \cellcolor[HTML]{FFD9B3}0.0052            & \cellcolor[HTML]{FFD9B3}0.0055             & \cellcolor[HTML]{FFD9B3}0.0060            & \cellcolor[HTML]{FFD9B3}0.0073             & \cellcolor[HTML]{FFD9B3}0.0083            & \cellcolor[HTML]{FFD9B3}0.0115             & \cellcolor[HTML]{FFD9B3}0.0076            & \cellcolor[HTML]{FFD9B3}0.0114            \\
Ours        & \cellcolor[HTML]{FFB3B3}0.0031            & \cellcolor[HTML]{FFB3B3}0.0000             & \cellcolor[HTML]{FFB3B3}0.0049            & \cellcolor[HTML]{FFB3B3}0.0000             & \cellcolor[HTML]{FFB3B3}0.0067            & \cellcolor[HTML]{FFB3B3}0.0000             & \cellcolor[HTML]{FFB3B3}0.0040            & \cellcolor[HTML]{FFB3B3}0.0000           
\end{tabular}
}\\
\caption{Experimental evidence that our method does not suffer popping, using the metric from StopThePop~\cite{radl2024stopthepop}. \reflectbox{F}LIP$_7$ is considered a more reliable step size by the authors because the signal is stronger than the noise caused by the optical flow used to compute the metric.}
\label{tab:popping}
\end{table}

As established in Figure~\ref{fig:popping_diagram}, both 3DGS and StopThePop exhibit artifacts due to their view-dependent density. We show what these artifacts look like using epipolar plane images \cite{bolles1987epipolar} in Figure~\ref{fig:popping_train}.
Because our rendering model is exact, and because the volume rendering integral is a continuous function of ray direction, our renderings exhibit no popping. Please see the supplemental video for a clear visualization.

To accompany our theoretical explanation of why our method does not suffer popping, we compare our method to StopThePop and 3DGS using the metric from StopThePop~(StP) to measure the amount of popping, closely following the original author's methodology across all the original scenes measured. EVER achieves a lower (better) score at step=1, and 0.0 error (lowest) with step=7, which StP called a more reliable step size, see Table.~\ref{tab:popping}. 

\subsection{Performance}
At 720p, we achieve average framerates of 36 FPS on the mip-NeRF 360 outdoor scenes, 66 FPS on the mip-NeRF 360 indoor scenes, and 30 FPS on the Zip-NeRF scenes on an NVIDIA RTX4090. The test set resolution is high on a couple of the Zip-NeRF and Mip-NeRF360 scenes, which causes the drop in FPS in Table~\ref{tab:main_results}. Training takes around 1-2 hours, also on an NVIDIA RTX4090. 
The majority of the time is spent tracing through the scene for rendering, which takes 20-71 ms, depending on the BVH.
For back propagation, 16 ms are spent storing intersections, and 13-20 ms are spent loading the primitives and atomically adding up the gradients. The BVH is rebuilt every training iteration, which takes 2-20 ms.

\newcommand{\graphimten}[1]{
\begin{tikzpicture}[zoomboxarray, zoomboxarray rows=1, zoomboxes below, zoomboxarray inner gap=0.0cm]
    \node [image node] { \includegraphics[width=4.5cm,valign=b]{#1} };
    \zoombox[magnification=4]{0.125,0.54}
\end{tikzpicture}
}

\begin{table}[b!]
\resizebox{\linewidth}{!}{
\large
\begin{tabular}{l|ccc|c}
                   & \multicolumn{1}{c}{PSNR$\uparrow$} & \multicolumn{1}{c}{SSIM$\uparrow$} & \multicolumn{1}{c|}{LPIPS$\downarrow$} & \multicolumn{1}{c}{FPS$\uparrow$}               \\ \hline
Splatted           & 26.63                               & .845                                & .329                                    & -                                                \\
No Mixing          & 26.79                               & .849                                & .305                                    & -                                                \\
3DGS               & 27.02                               & .853                                & .301                                    & -                                                \\
3DGS + Our Changes & 26.94                               & .832                                & .323                                    & -                                                \\
No Anisotropic     & \cellcolor[HTML]{FFB3B3}27.25       & \cellcolor[HTML]{FFB3B3}.865        & \cellcolor[HTML]{FFB3B3}.272            & \multicolumn{1}{r}{27.3}                         \\
No Points          & \cellcolor[HTML]{FFD9B3}27.19       & \cellcolor[HTML]{FFD9B3}.863        & \cellcolor[HTML]{FFD9B3}.277            & \multicolumn{1}{r}{\cellcolor[HTML]{FFFFB4}41.2} \\
No Rand Center     & 27.08                               & .859                                & \cellcolor[HTML]{FFFFB4}.278            & \multicolumn{1}{r}{\cellcolor[HTML]{FFD9B3}41.4} \\ \hline
Ours               & \cellcolor[HTML]{FFFFB4}27.17       & \cellcolor[HTML]{FFFFB4}.862        & \cellcolor[HTML]{FFD9B3}.277            & \multicolumn{1}{r}{\cellcolor[HTML]{FFB3B3}41.9}
\end{tabular}
}\\
\caption{
An ablation study reporting average performance on the \scenename{bicycle} and \scenename{counter} scenes from MipNeRF 360~\cite{barron2021mip} and the \scenename{berlin} scene from ZipNeRF~\cite{barron2023zip}. The ``No Anisotropic'' ablation shows the cost of our regularizer that increases frame-rate.}
\label{tab:ablation}
\end{table}

\subsection{Ablations}
To see if it was possible to train our model using ray tracing, then render using splatting, we implemented a splatting version of our 3D ellipsoids, labeled ``Splatted''. We also tested what would happen if we sorted the primitives per ray like StopThePop without our color blending technique, which is labeled ``No Mixing''. Both of these resulted in visual degradation in areas where the method had been mixing primitive colors together, which can be seen on the fridge and candle. 
These ablations also perform worse when trained for their specific rendering styles, as can be seen in Table~\ref{tab:ablation}. ``Splatting'' performs worse than ``No Mixing'', which does not match the comparison between 3DGS and StopThePop, both of which have similar performance. 

To ablate the effect of our densification changes, along with the other changes, we use the 3DGS renderer with the other minor changes we've made to densification, which we label ``3DGS + Our Changes''.
These changes do not help 3DGS because they are aimed at handling density primitives. 
Finally, we run ablations on each of the changes we performed.
``No anisotropic'' refers to ablating anisotropic loss, which results in slightly increased quality, but much lower framerate. ``No points'' refers to no additional points added using the inverse contraction, which reduces the reliance on the SfM reconstruction. The inverse contraction sampling is described in the supplemental material.
``No Rand Centers'' refers to ablating randomizing the pixel centers, which helps with thin structures on the bicycle scene.
We jitter our rays by picking a point within the pixel from a uniform random distribution, as opposed to using the center of the pixel. 

\section{Conclusion}


%

Our double-intersection approach for \longname{} can be used to blend any number of overlapping primitive colors in a physically accurate way. We have shown that our method outperforms Zip-NeRF on the larger Zip-NeRF datasets on sharpness, while outperforming all other real-time methods on SSIM and LPIPS on the Mip-NeRF360 and Zip-NeRF datasets. By isolating the effect of our renderer, we know that this improvement comes from a combination of density based primitives and double-intersection based, physically accurate blending.
This results in an approach that bridges the gap between slow but accurate radiance field methods such as Zip-NeRF, and fast but inaccurate radiance field methods like 3DGS. 

We also show that sorting primitives per a pixel is not sufficient for eliminating the popping artifacts in 3DGS. It is only by combining per pixel sorting, which we achieve with ray tracing, with proper primitive blending, that we are able to eliminate these artifacts. 
Together, \acronym{} is able to produce high-quality, large scale renderings that are guaranteed to be 3D-consistent, and does so at 30 FPS at 720p on a single NVIDIA RTX4090. 



\iftoggle{cvprfinal}{
\myparagraph{Acknowledgements}
We would like to thank Delio Vicini, Stephan Garbin and Ben Lee for helpful discussions.
AM is funded in part by the USACE Engineer Research and Development Center Cooperative Agreement W9132T-22-2-0014.
}{}

{
    \small
    \bibliographystyle{ieeenat_fullname}
    \bibliography{main}
}

\iftoggle{cvprfinal}{
\appendix
\clearpage
\setcounter{page}{1}
\maketitlesupplementary

\section{Backpropagation}
We use an adjoint rendering approach, starting from the last state of each ray, which we store, and reconstruct each ray state backwards using $p^{-1}(x, i)$. We can then use this reconstructed state to backpropagate the error to each state, then propagate this error to each primitive. The gradient with respect to mean, scale, and orientation, all come from the derivative of the ray-ellipsoid intersection function.  
To retrieve the list of primitives intersected, we store the list of intersections on the forward pass. Since rays tend to terminate before 300 total intersections, this turns out to be relatively cheap, at 1.2 KB per a ray. Although we experimented with using ray tracing to retrieve the list of surfaces in reverse order, we found the instability too high.

\begin{figure*}[ht]
\centering
\bgroup
\begin{tabular}{@{}c@{}c@{}c@{}c@{}c@{}}
\graphimtwelve{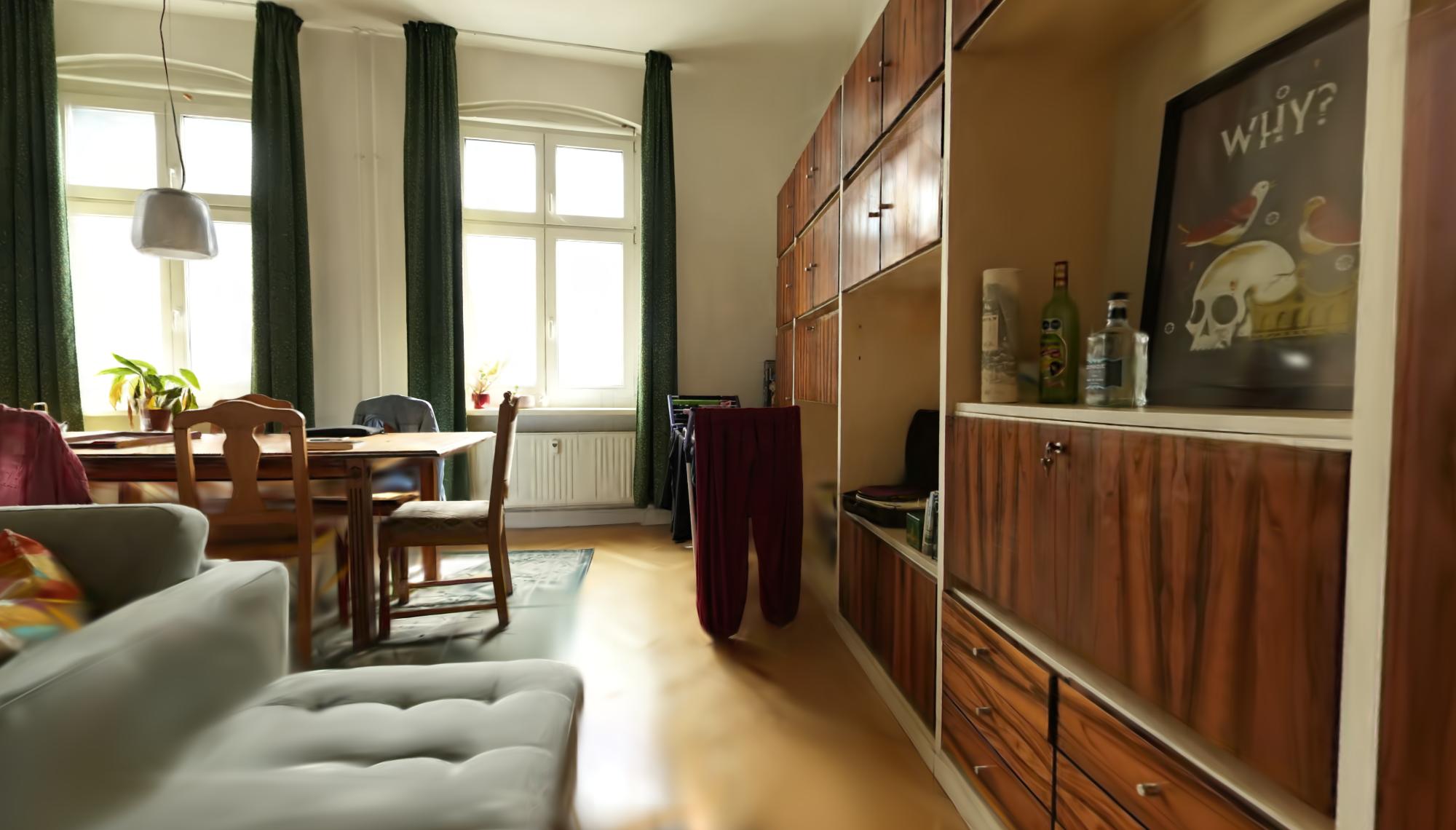} &
\graphimtwelve{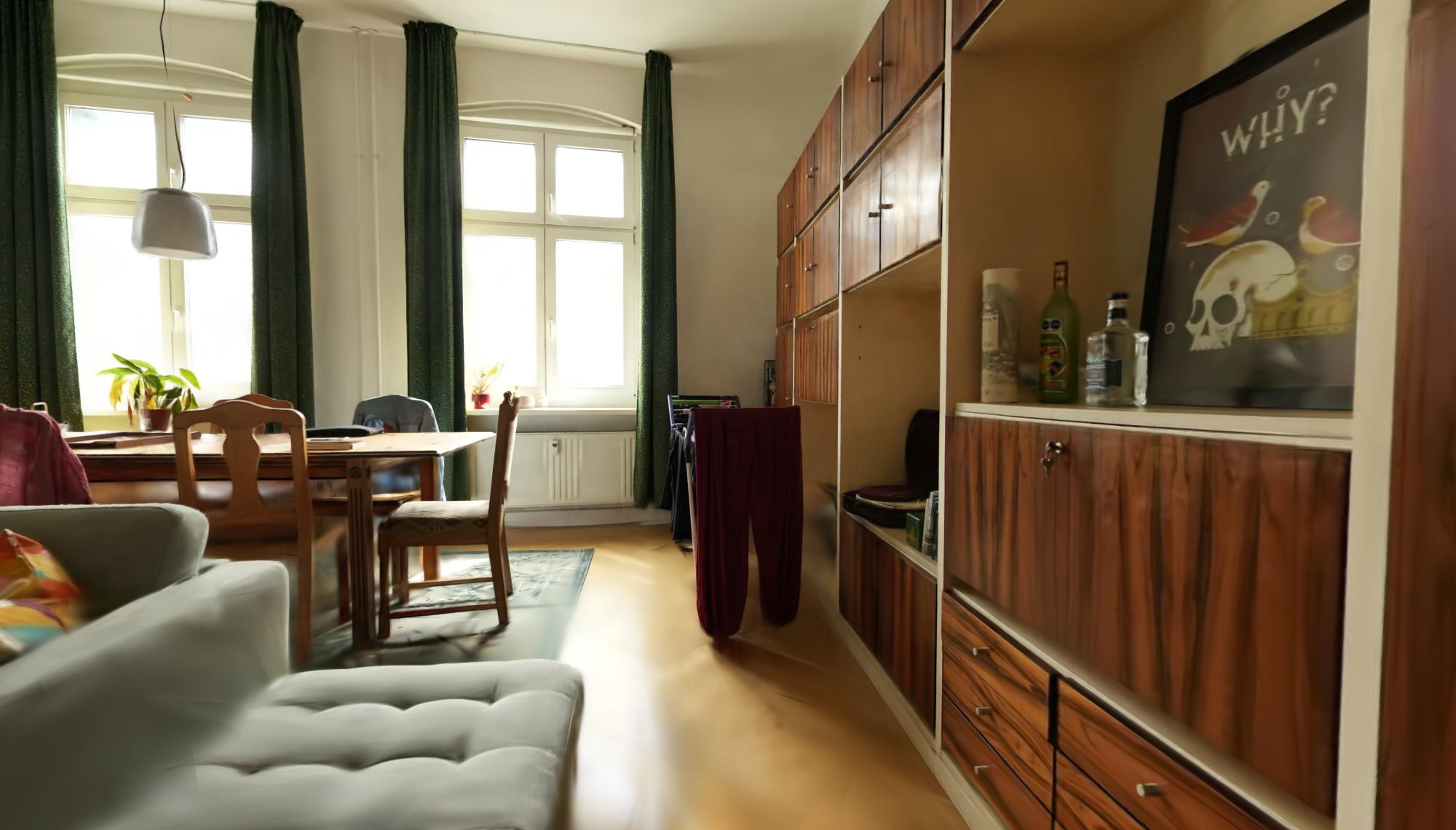} &
\graphimtwelve{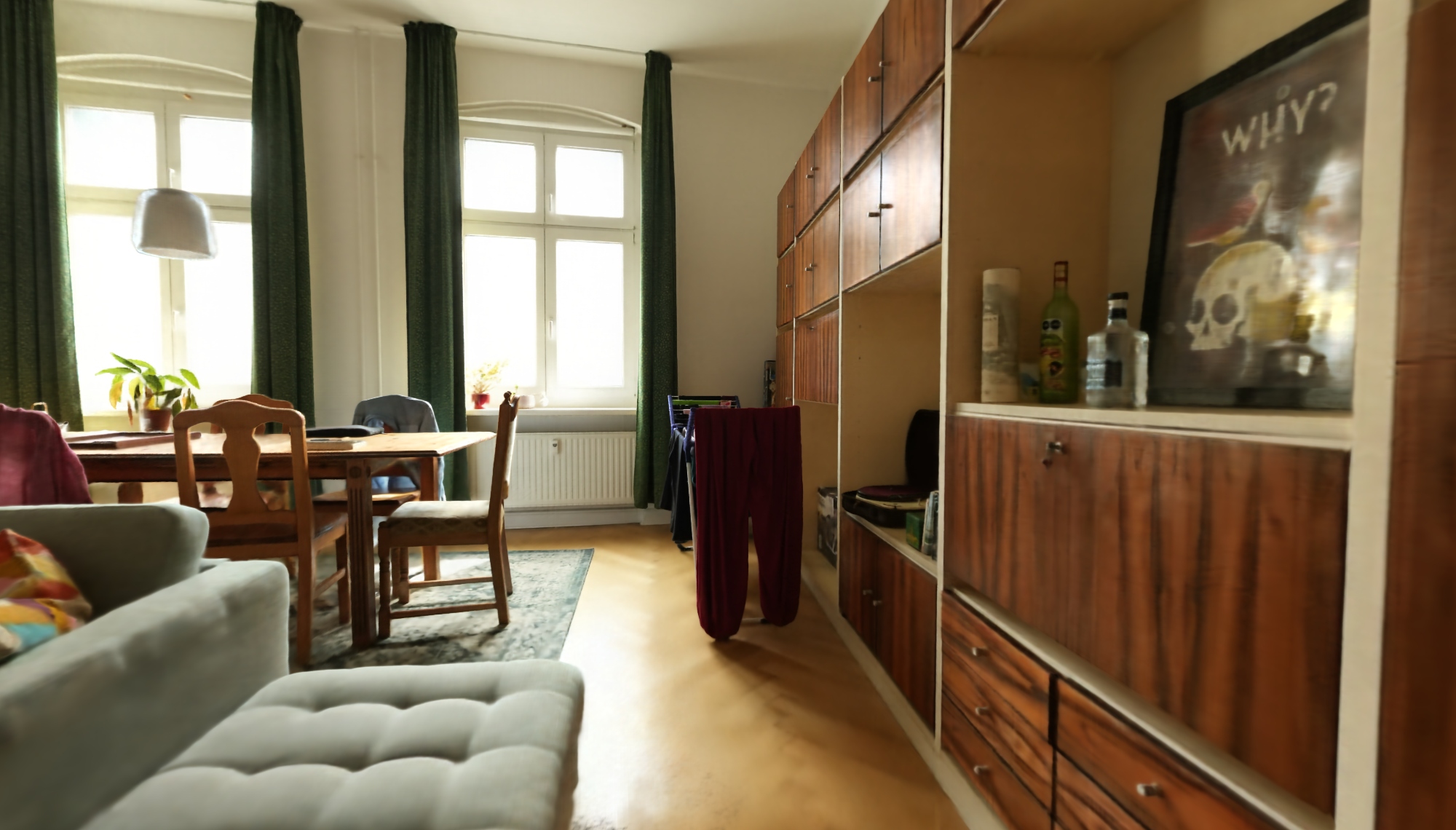} &
\graphimtwelve{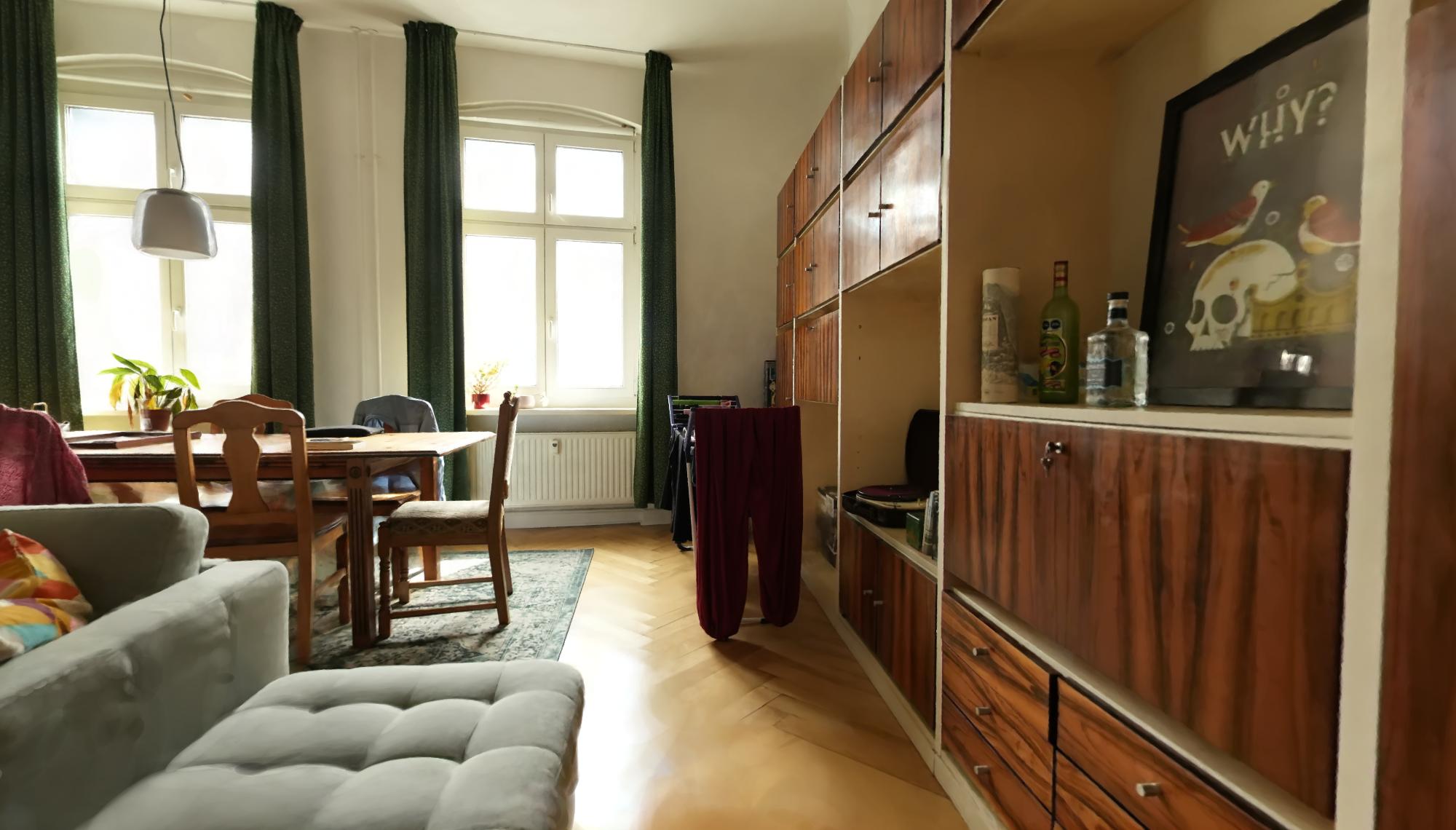} &
\graphimtwelve{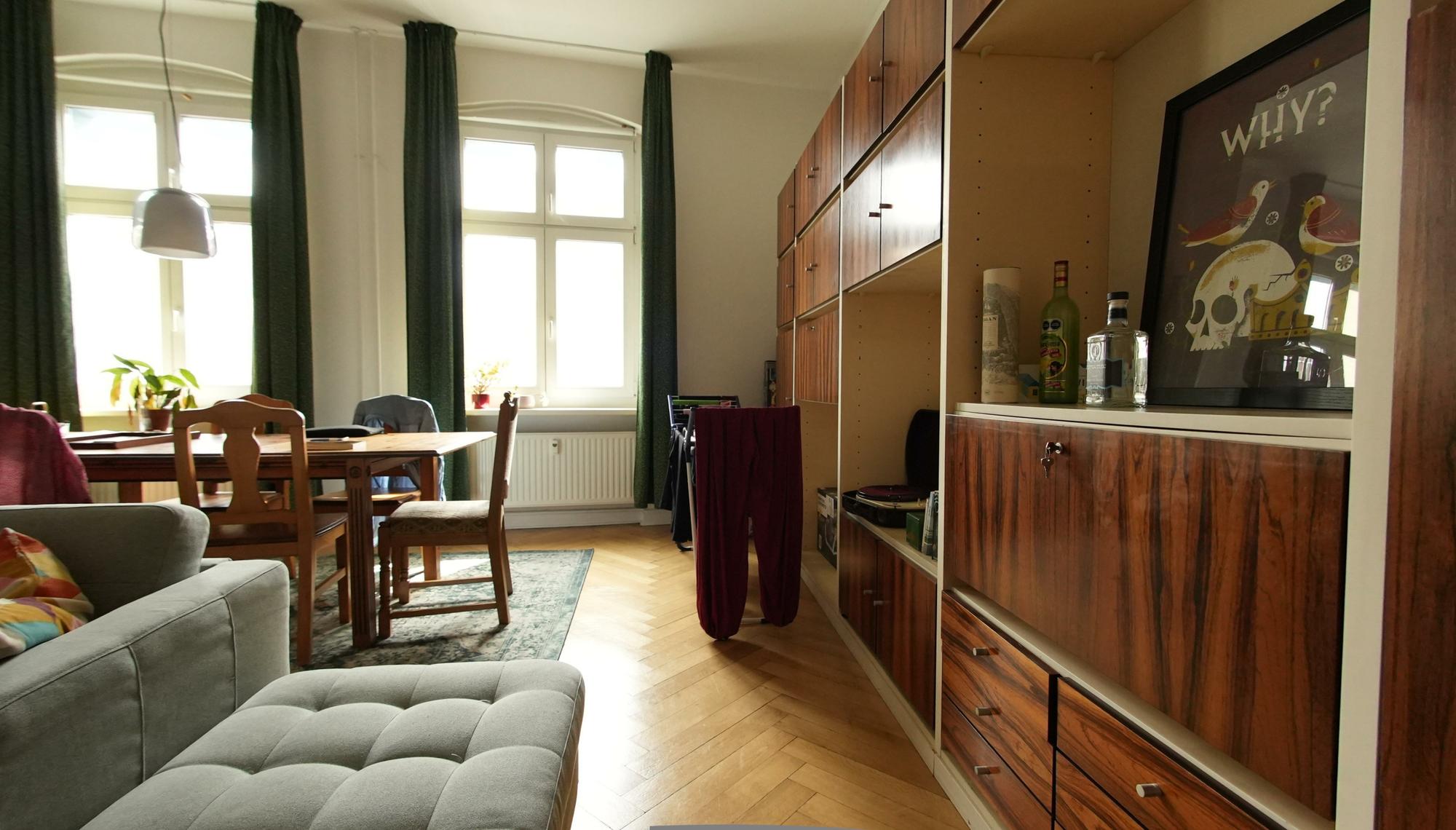} \\
\graphimthirteen{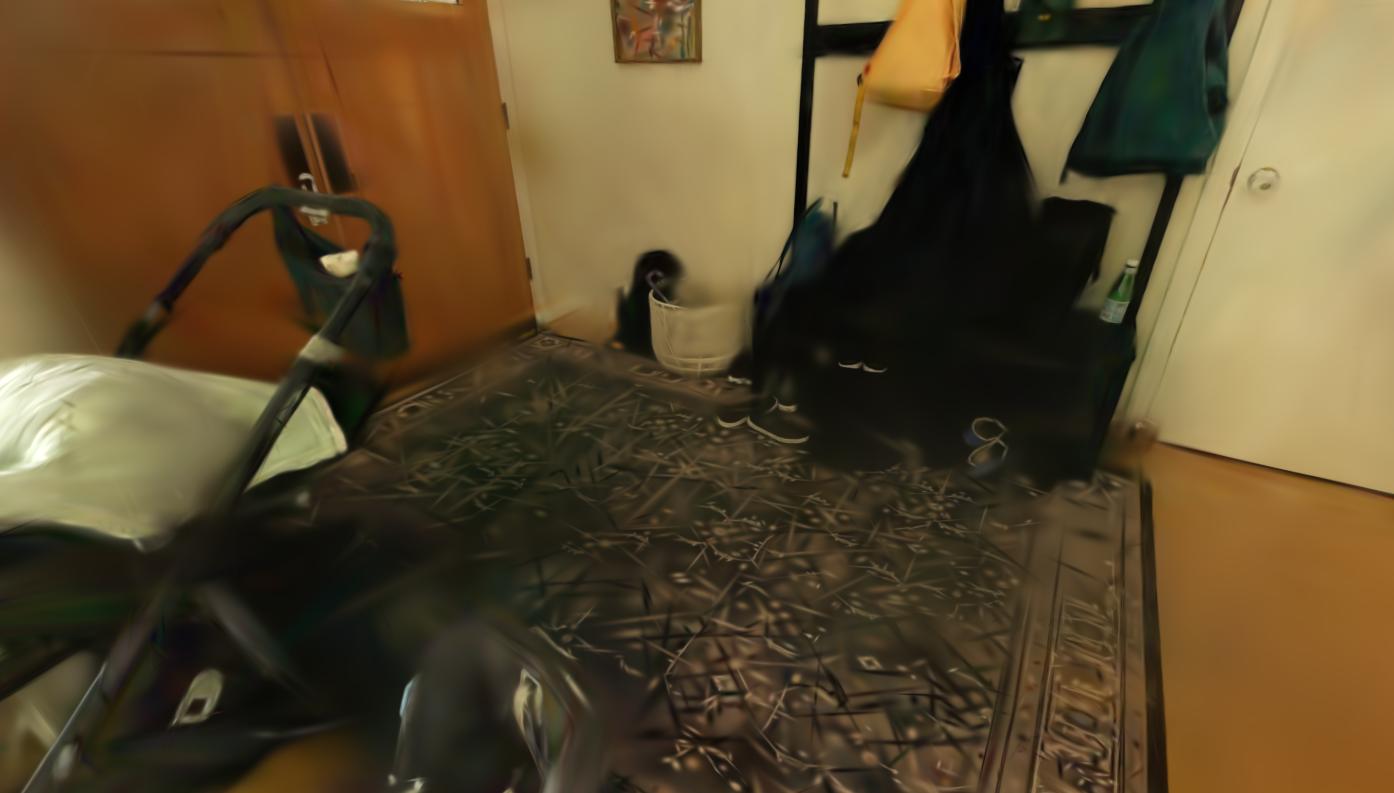} &
\graphimthirteen{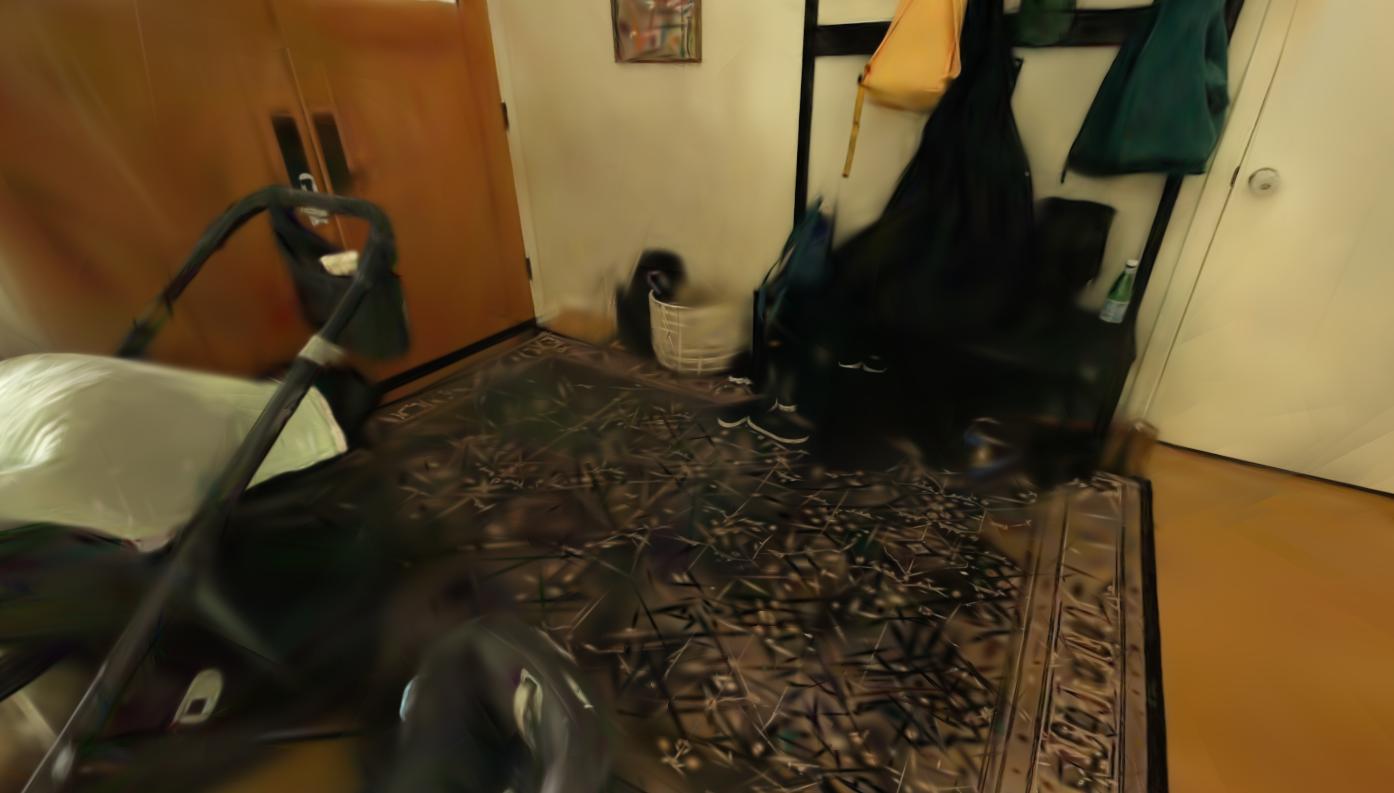} &
\graphimthirteen{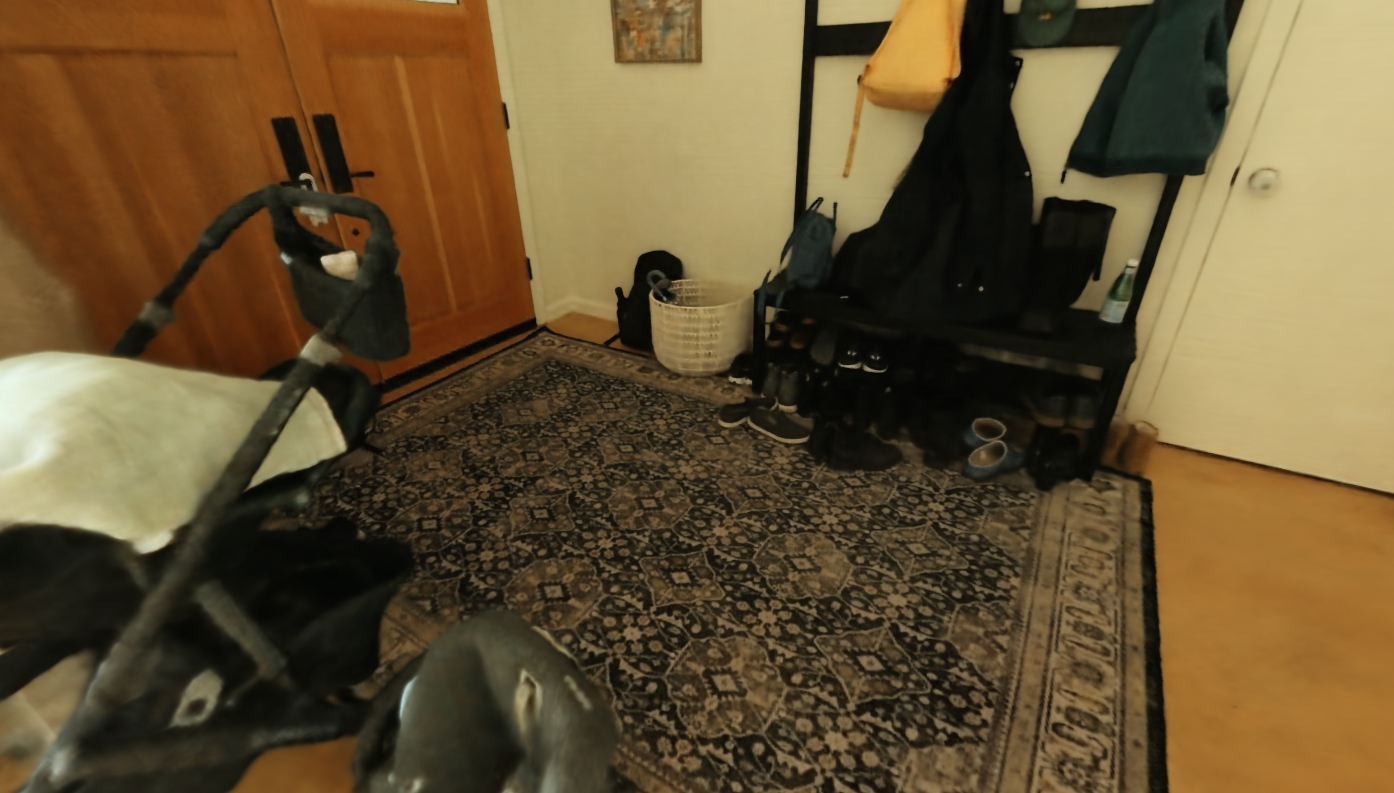} &
\graphimthirteen{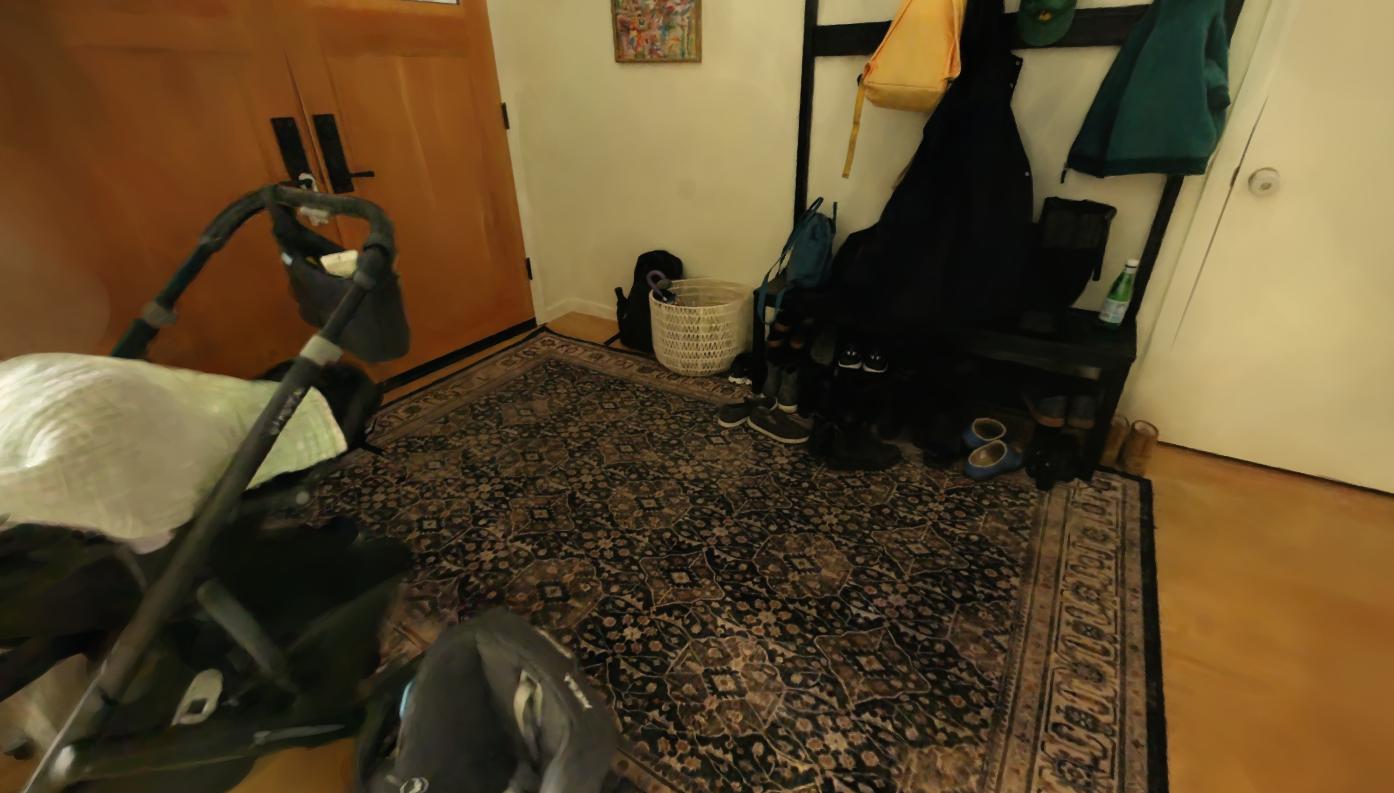} &
\graphimthirteen{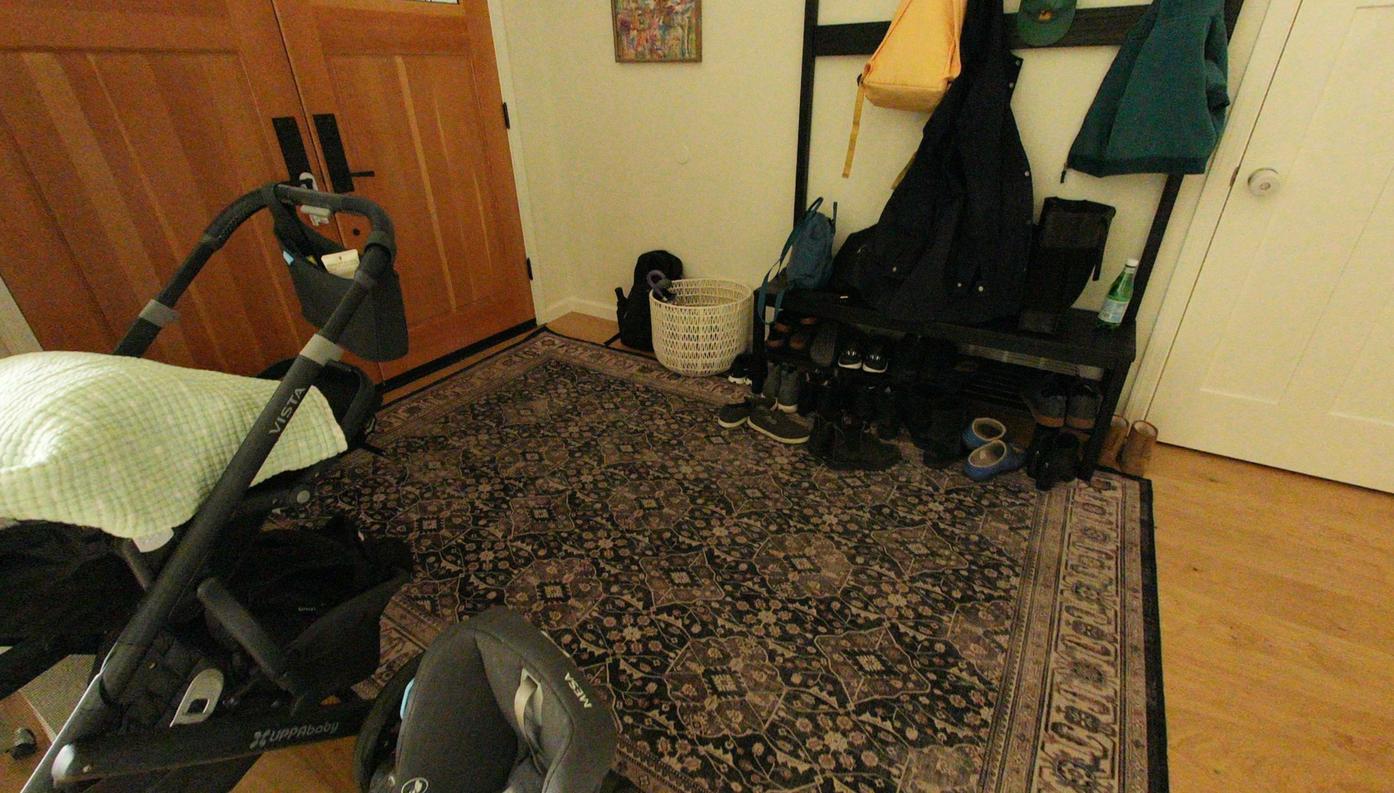} \\
\graphimseven{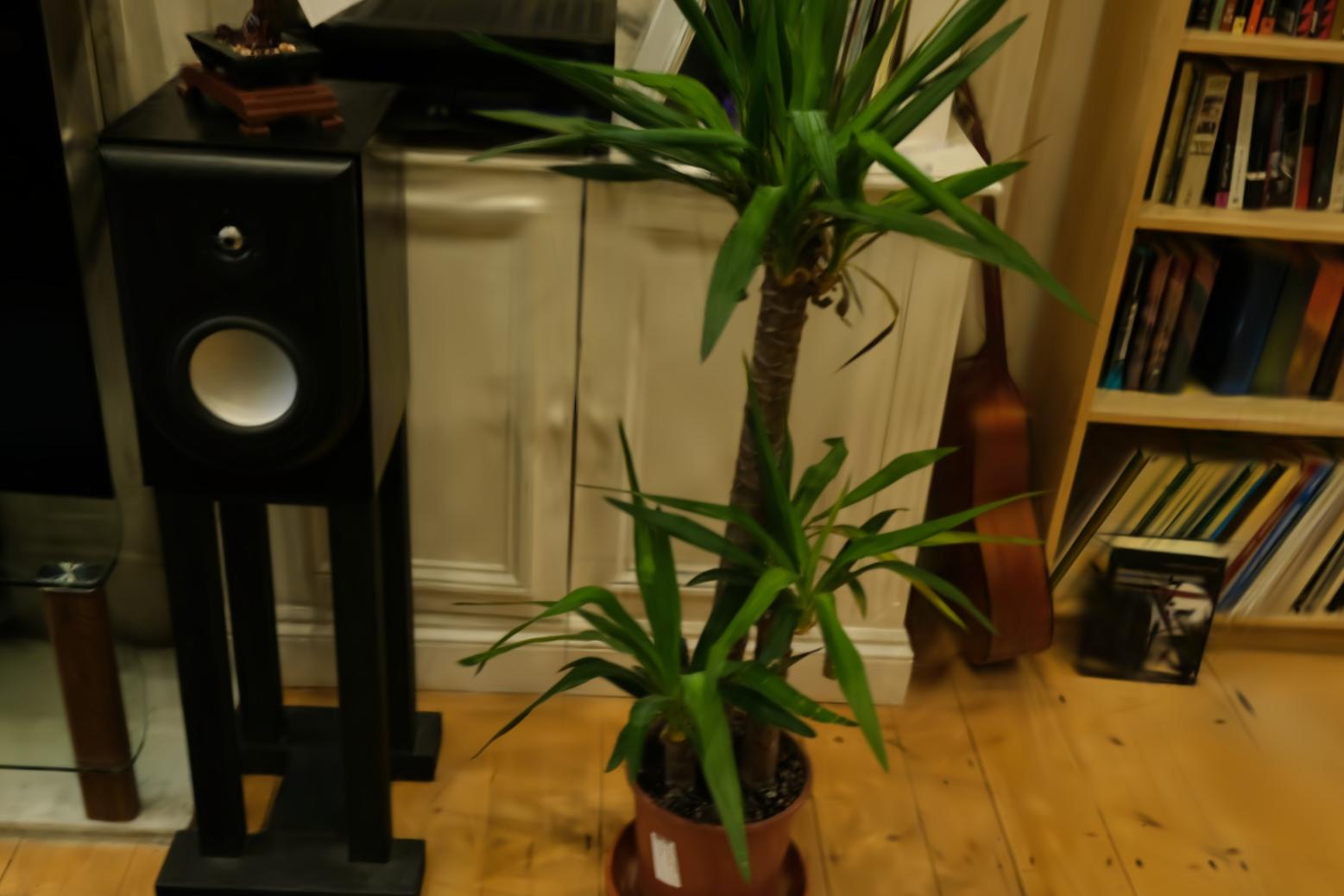} &
\graphimseven{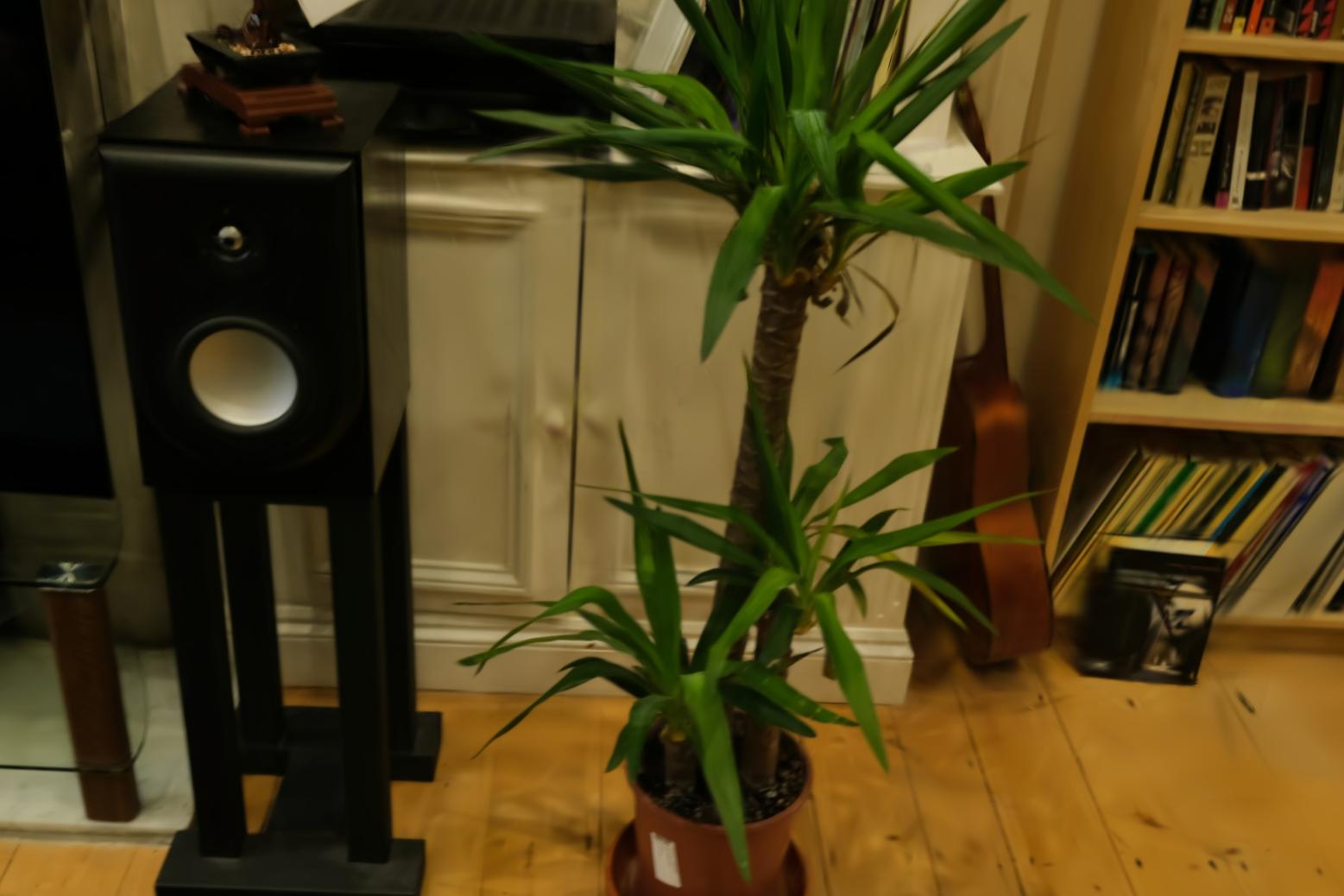} &
\graphimseven{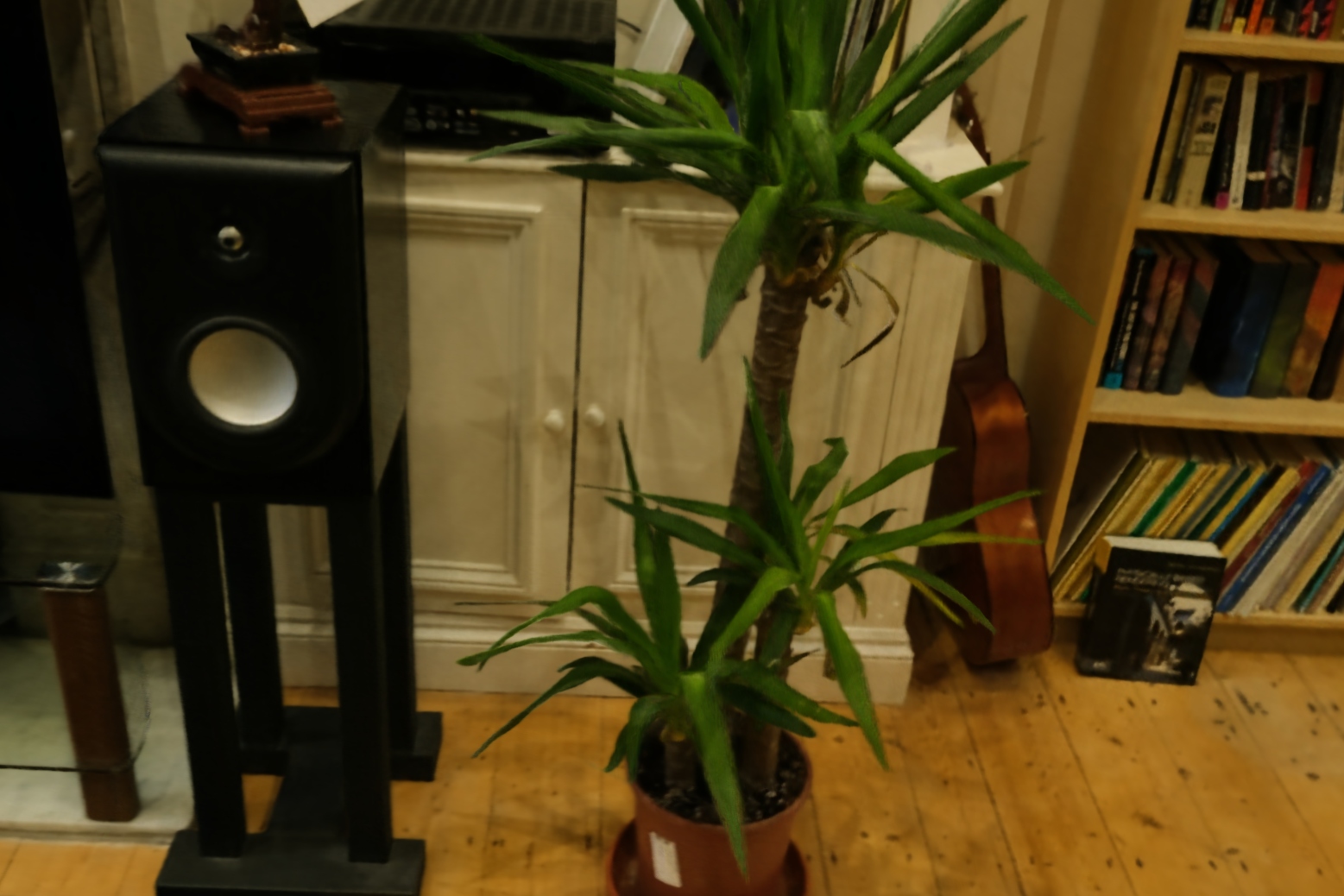} &
\graphimseven{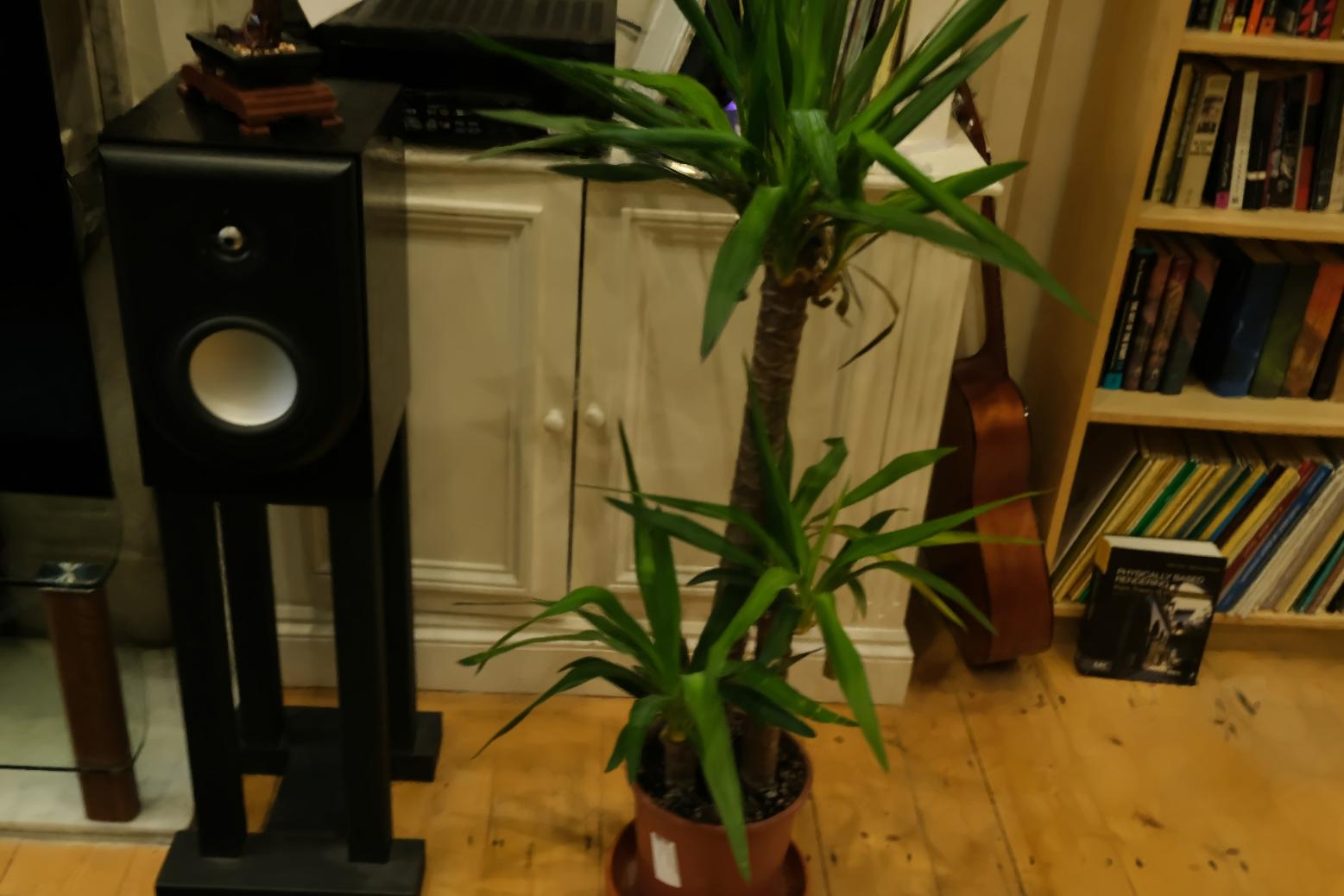} &
\graphimseven{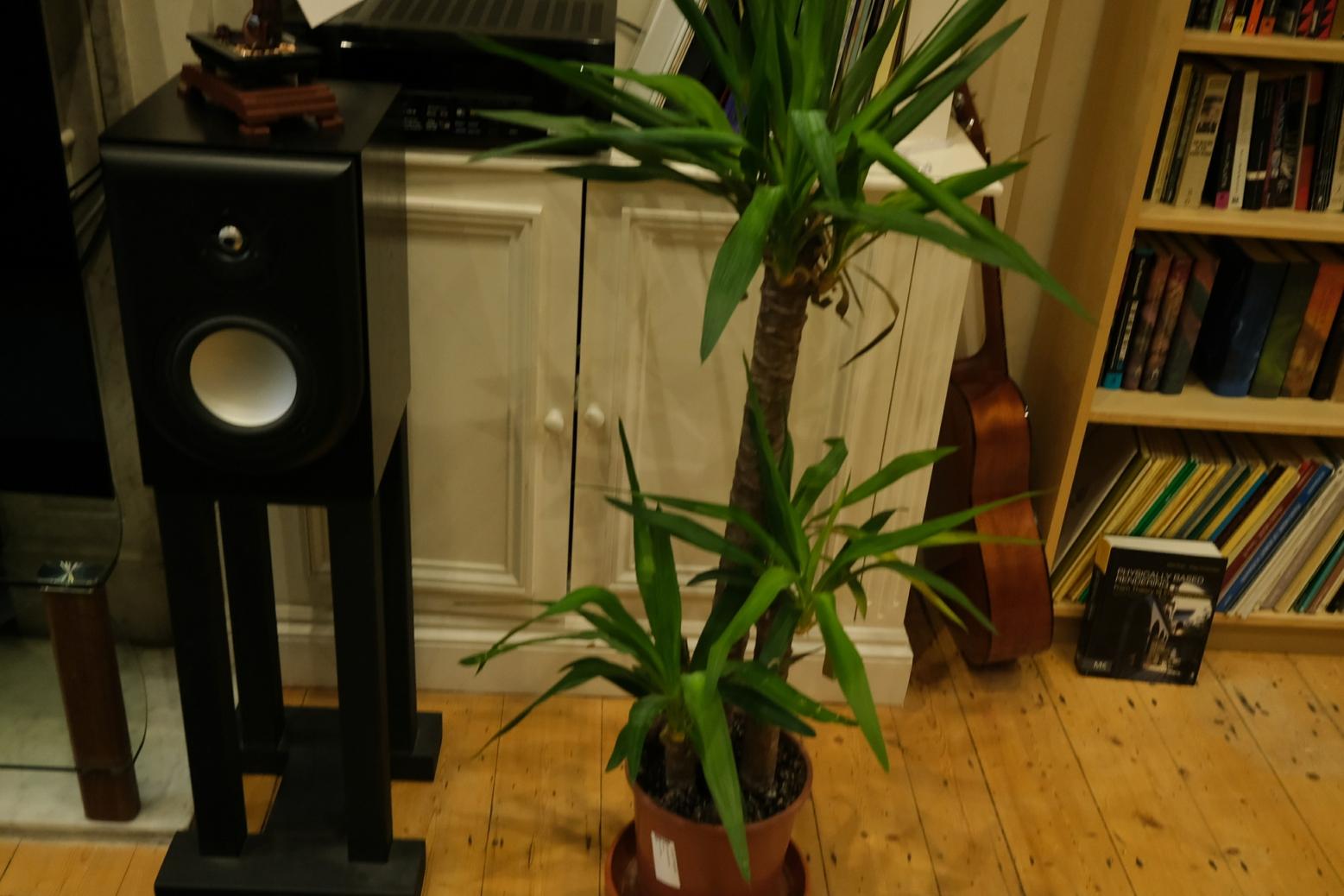} \\

\graphimfourteen{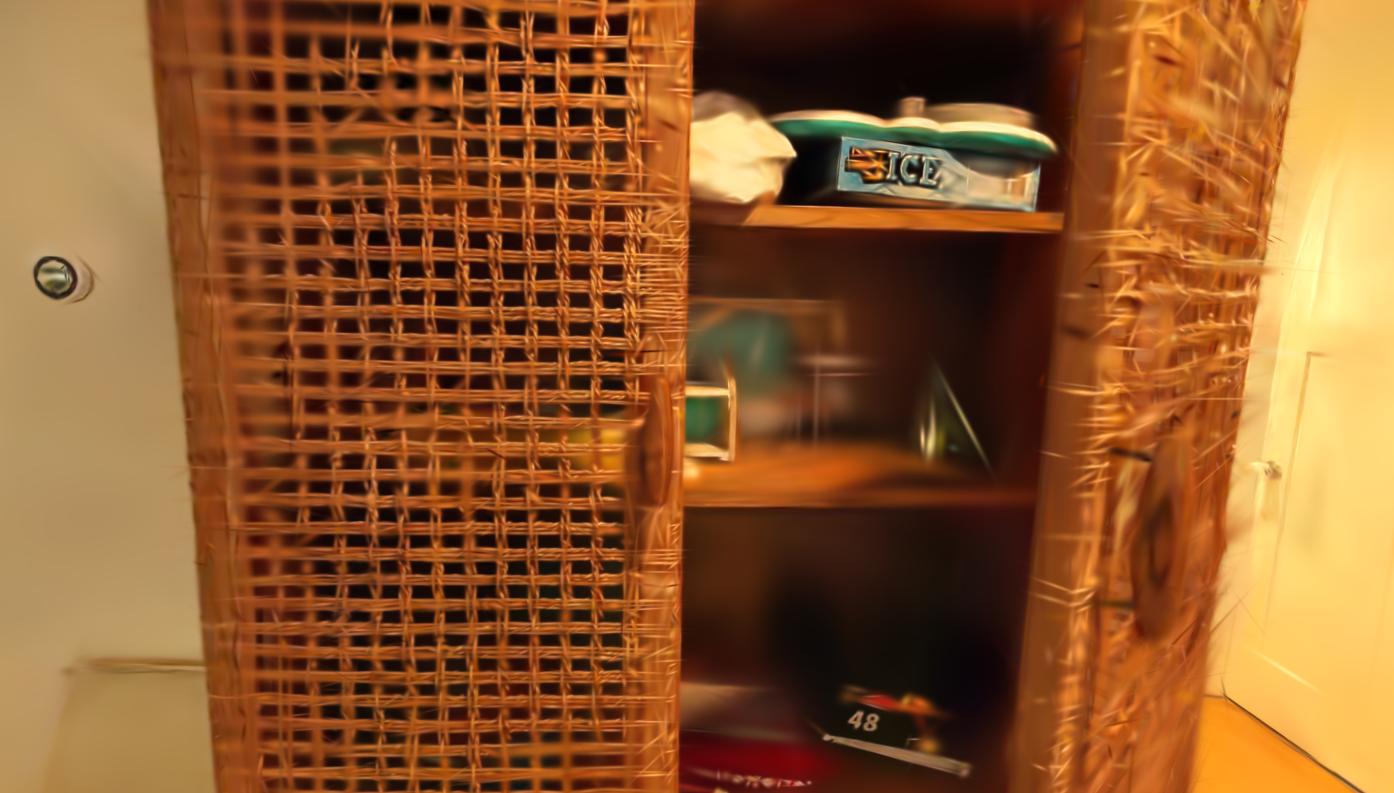} &
\graphimfourteen{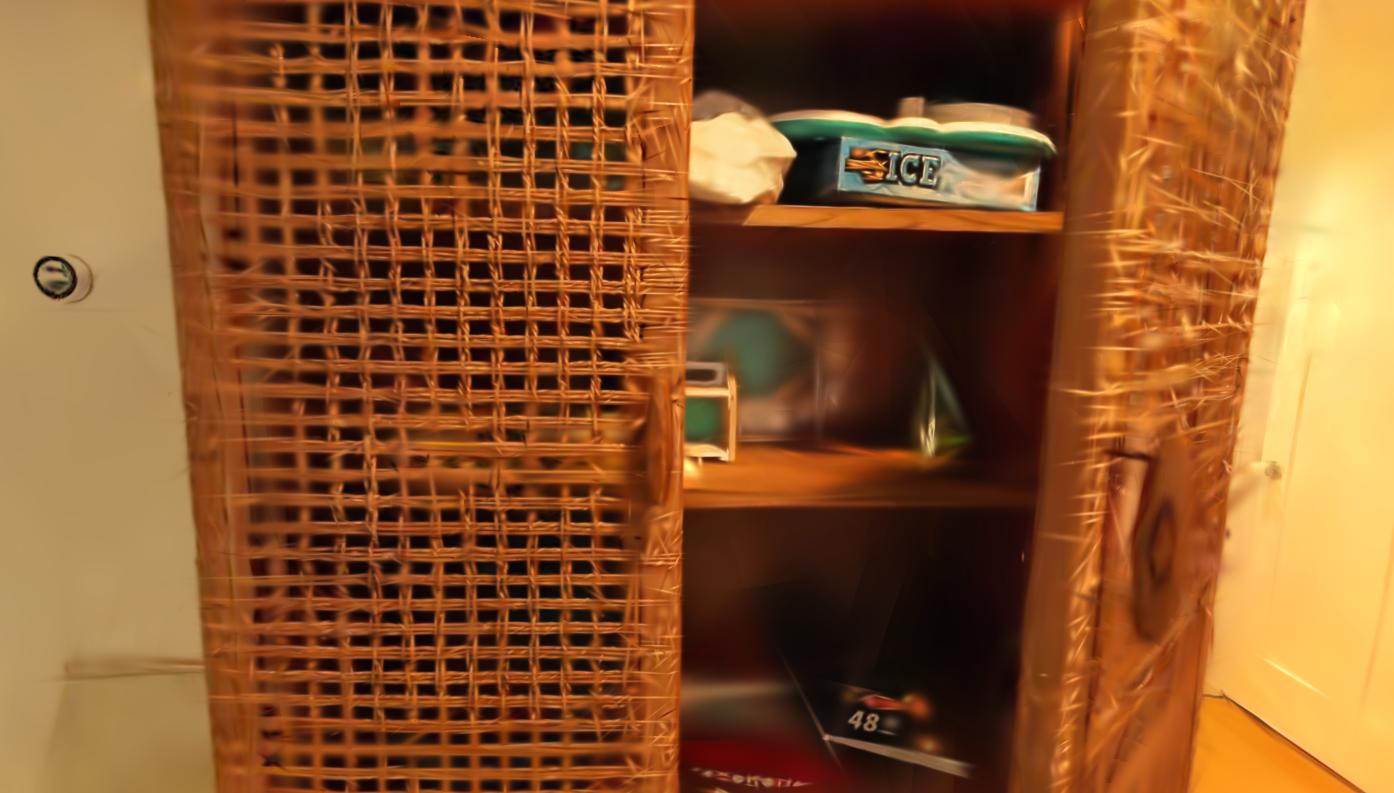} &
\graphimfourteen{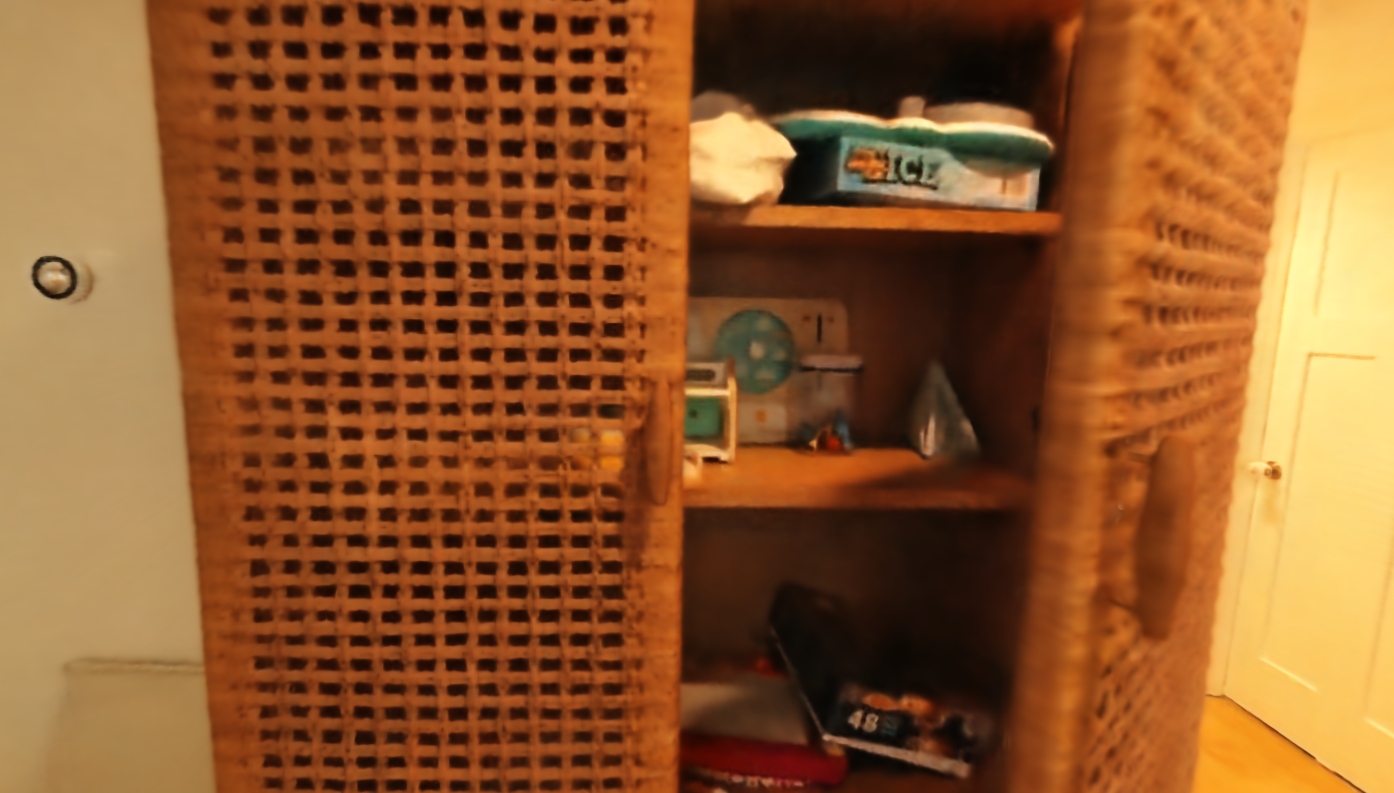} &
\graphimfourteen{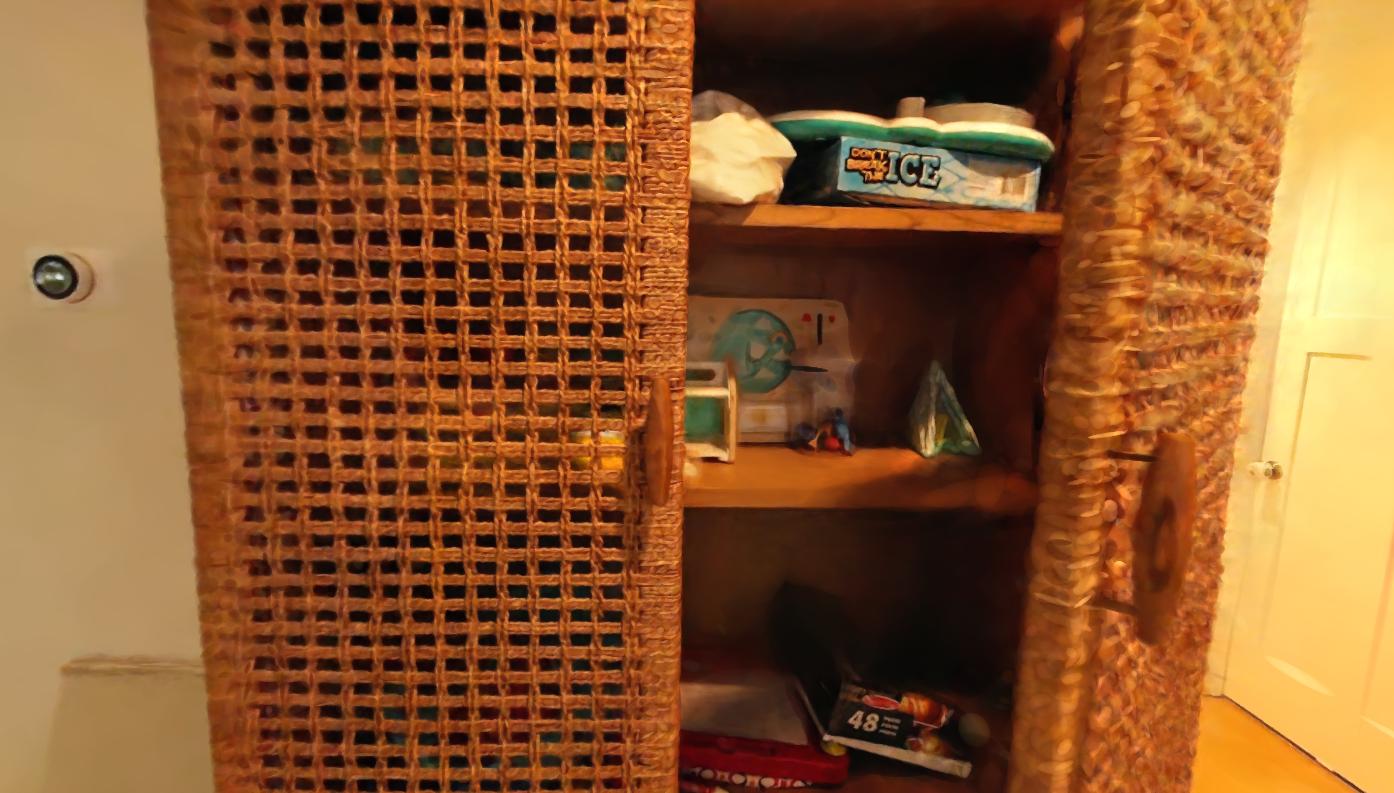} &
\graphimfourteen{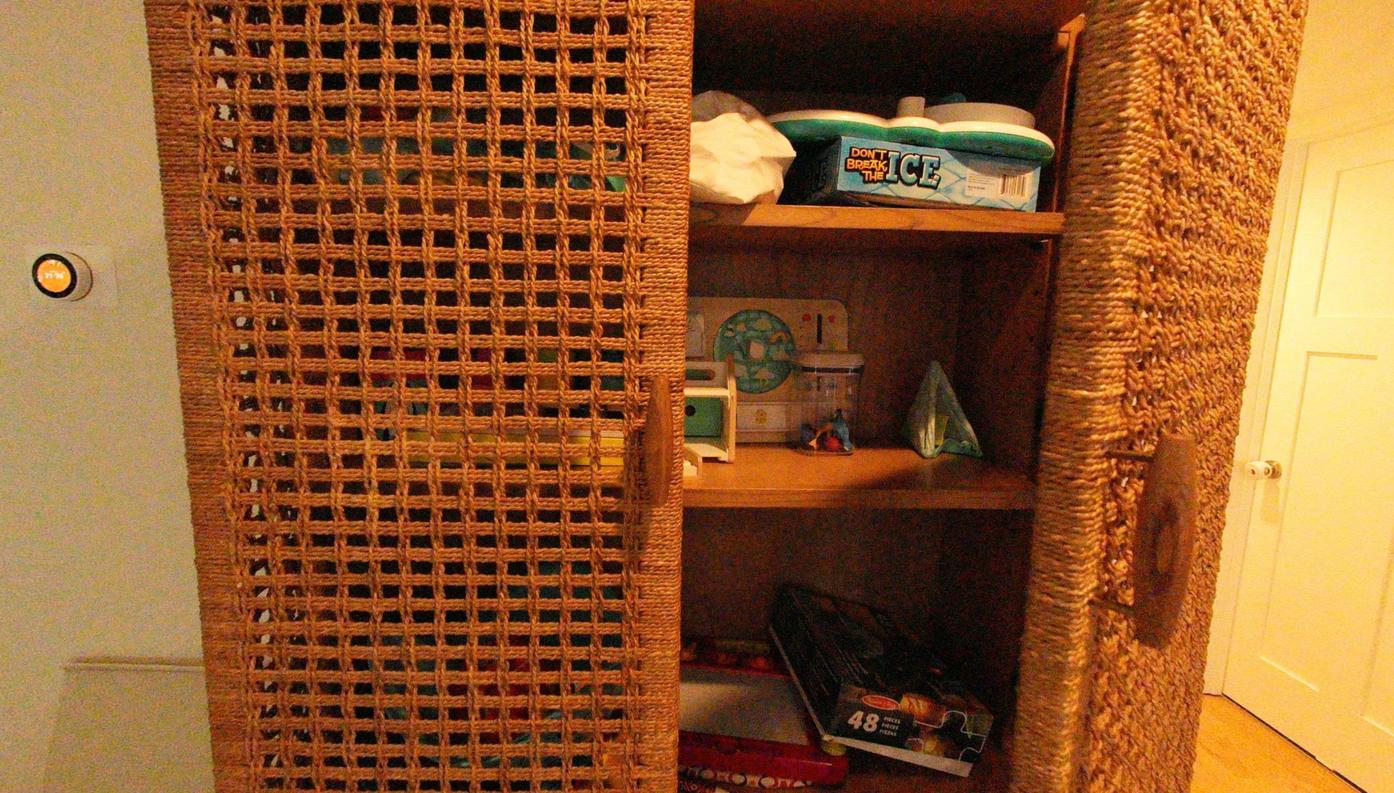} \\


3DGS & StopThePop & SMERF & Ours & GT
\end{tabular}
\egroup
\caption{Additional visual comparison of our method on the Mip-NeRF 360 dataset~\cite{barron2023zip}.}
\label{fig:comparison_supp}
\end{figure*}

\begin{table*}[ht]
    \centering
    \begin{tabular}{l|rrrrr}
\rowcolor[HTML]{F1F3F4} 
\textbf{PSNR $\uparrow$}    & \multicolumn{1}{l}{\cellcolor[HTML]{F1F3F4}berlin} & \multicolumn{1}{l}{\cellcolor[HTML]{F1F3F4}nyc} & \multicolumn{1}{l}{\cellcolor[HTML]{F1F3F4}alameda} & \multicolumn{1}{l}{\cellcolor[HTML]{F1F3F4}london} & \multicolumn{1}{l}{\cellcolor[HTML]{F1F3F4}Average} \\ \hline
3DGS                        & \cellcolor[HTML]{FFFFB4}26.83                      & 26.90                                           & \cellcolor[HTML]{FFFFB4}24.14                       & 25.48                                              & 25.84                                               \\
Mip Splatting               & \cellcolor[HTML]{FFB3B3}27.30                      & \cellcolor[HTML]{FFD9B3}27.52                   & \cellcolor[HTML]{FFB3B3}24.76                       & \cellcolor[HTML]{FFD9B3}26.28                      & \cellcolor[HTML]{FFD9B3}26.46                       \\
StopThePop                  & 26.81                                              & \cellcolor[HTML]{FFFFB4}27.14                   & 24.12                                               & \cellcolor[HTML]{FFFFB4}25.61                      & \cellcolor[HTML]{FFFFB4}25.92                       \\
SMERF                       & 28.52                                              & \cellcolor[HTML]{FFFFB4}28.21                   & 25.35                                               & \cellcolor[HTML]{FFFFB4}27.05                      & \cellcolor[HTML]{FFFFB4}27.28                       \\
Ours                        & 27.24                                              & 27.93                                           & 24.72                                               & 26.49                                              & \cellcolor[HTML]{FFB3B3}26.60                       \\ \hline
Zip-NeRF                    & 28.59                                              & 28.42                                           & 25.41                                               & 27.06                                              & 27.37                                               \\ \hline
\rowcolor[HTML]{F1F3F4} 
\textbf{SSIM $\uparrow$}    & \multicolumn{1}{l}{\cellcolor[HTML]{F1F3F4}berlin} & \multicolumn{1}{l}{\cellcolor[HTML]{F1F3F4}nyc} & \multicolumn{1}{l}{\cellcolor[HTML]{F1F3F4}alameda} & \multicolumn{1}{l}{\cellcolor[HTML]{F1F3F4}london} & \multicolumn{1}{l}{\cellcolor[HTML]{F1F3F4}Average} \\ \hline
3DGS                        & .899                                               & .861                                            & .776                                                & .830                                               & .842                                                \\
Mip Splatting               & \cellcolor[HTML]{FFD9B3}.892                       & \cellcolor[HTML]{FFD9B3}.853                    & \cellcolor[HTML]{FFD9B3}.768                        & \cellcolor[HTML]{FFD9B3}.822                       & \cellcolor[HTML]{FFD9B3}.834                        \\
StopThePop                  & \cellcolor[HTML]{FFFFB4}.885                       & \cellcolor[HTML]{FFFFB4}.844                    & \cellcolor[HTML]{FFFFB4}.748                        & \cellcolor[HTML]{FFFFB4}.801                       & \cellcolor[HTML]{FFFFB4}.819                        \\
SMERF                       & \cellcolor[HTML]{FFFFB4}.887                       & \cellcolor[HTML]{FFFFB4}.844                    & \cellcolor[HTML]{FFFFB4}.758                        & \cellcolor[HTML]{FFFFB4}.829                       & \cellcolor[HTML]{FFFFB4}.830                        \\
Ours                        & \cellcolor[HTML]{FFB3B3}.900                       & \cellcolor[HTML]{FFB3B3}.863                    & \cellcolor[HTML]{FFB3B3}.779                        & \cellcolor[HTML]{FFB3B3}.837                       & \cellcolor[HTML]{FFB3B3}.845                        \\ \hline
Zip-NeRF                    & .891                                               & .850                                            & .767                                                & .835                                               & .836                                                \\ \hline
\rowcolor[HTML]{F1F3F4} 
\textbf{LPIPS $\downarrow$} & \multicolumn{1}{l}{\cellcolor[HTML]{F1F3F4}berlin} & \multicolumn{1}{l}{\cellcolor[HTML]{F1F3F4}nyc} & \multicolumn{1}{l}{\cellcolor[HTML]{F1F3F4}alameda} & \multicolumn{1}{l}{\cellcolor[HTML]{F1F3F4}london} & \multicolumn{1}{l}{\cellcolor[HTML]{F1F3F4}Average} \\ \hline
3DGS                        & \cellcolor[HTML]{FFFFB4}.406                       & \cellcolor[HTML]{FFFFB4}.380                    & \cellcolor[HTML]{FFFFB4}.441                        & \cellcolor[HTML]{FFFFB4}.446                       & \cellcolor[HTML]{FFFFB4}.418                        \\
Mip Splatting               & .392                                               & .356                                            & .410                                                & .411                                               & .392                                                \\
StopThePop                  & \cellcolor[HTML]{FFD9B3}.402                       & \cellcolor[HTML]{FFD9B3}.373                    & \cellcolor[HTML]{FFD9B3}.433                        & \cellcolor[HTML]{FFD9B3}.438                       & \cellcolor[HTML]{FFD9B3}.411                        \\
SMERF                       & \cellcolor[HTML]{FFD9B3}.391                       & \cellcolor[HTML]{FFD9B3}.361                    & \cellcolor[HTML]{FFD9B3}.416                        & \cellcolor[HTML]{FFD9B3}.390                       & \cellcolor[HTML]{FFD9B3}.389                        \\
Ours                        & \cellcolor[HTML]{FFB3B3}.371                       & \cellcolor[HTML]{FFB3B3}.337                    & \cellcolor[HTML]{FFB3B3}.389                        & \cellcolor[HTML]{FFB3B3}.374                       & \cellcolor[HTML]{FFB3B3}.368                        \\ \hline
Zip-NeRF                    & .378                                               & .331                                            & .387                                                & .360                                               & .364                                               
\end{tabular}
    \caption{Full results for Zip-NeRF dataset}
    \centering
    \begin{tabular}{l|rrrrr|rrrr}
\rowcolor[HTML]{F1F3F4} 
\textbf{PSNR $\uparrow$}    & \multicolumn{1}{l}{\cellcolor[HTML]{F1F3F4}bicycle} & \multicolumn{1}{l}{\cellcolor[HTML]{F1F3F4}flowers} & \multicolumn{1}{l}{\cellcolor[HTML]{F1F3F4}garden} & \multicolumn{1}{l}{\cellcolor[HTML]{F1F3F4}stump} & \multicolumn{1}{l|}{\cellcolor[HTML]{F1F3F4}treehill} & \multicolumn{1}{l}{\cellcolor[HTML]{F1F3F4}room} & \multicolumn{1}{l}{\cellcolor[HTML]{F1F3F4}counter} & \multicolumn{1}{l}{\cellcolor[HTML]{F1F3F4}kitchen} & \multicolumn{1}{l}{\cellcolor[HTML]{F1F3F4}bonsai} \\ \hline
3DGS                        & \cellcolor[HTML]{FFFFB4}25.24                       & 21.55                                               & \cellcolor[HTML]{FFFFB4}27.38                      & 26.56                                             & 22.43                                                 & \cellcolor[HTML]{FFD9B3}31.53                    & \cellcolor[HTML]{FFFFB4}29.00                       & \cellcolor[HTML]{FFD9B3}31.45                       & \cellcolor[HTML]{FFFFB4}32.21                      \\
StopThePop                  & 25.23                                               & \cellcolor[HTML]{FFFFB4}21.62                       & 27.33                                              & \cellcolor[HTML]{FFD9B3}26.65                     & \cellcolor[HTML]{FFFFB4}22.44                         & 30.91                                            & 28.79                                               & \cellcolor[HTML]{FFFFB4}31.13                       & 31.85                                              \\
3DGRT                       & 25.13                                               & 21.58                                               & 26.99                                              & \cellcolor[HTML]{FFFFB4}26.57                     & 22.40                                                 & 30.92                                            & 28.78                                               & 30.60                                               & 31.85                                              \\
SMERF                       & \cellcolor[HTML]{FFB3B3}25.58                       & \cellcolor[HTML]{FFB3B3}22.24                       & \cellcolor[HTML]{FFB3B3}27.66                      & \cellcolor[HTML]{FFB3B3}27.19                     & \cellcolor[HTML]{FFB3B3}23.93                         & \cellcolor[HTML]{FFFFB4}31.38                    & \cellcolor[HTML]{FFB3B3}29.02                       & \cellcolor[HTML]{FFB3B3}31.68                       & \cellcolor[HTML]{FFB3B3}33.19                      \\
Our model                   & \cellcolor[HTML]{FFD9B3}25.34                       & \cellcolor[HTML]{FFD9B3}21.70                       & \cellcolor[HTML]{FFD9B3}27.46                      & 26.41                                             & \cellcolor[HTML]{FFD9B3}22.74                         & \cellcolor[HTML]{FFB3B3}31.39                    & \cellcolor[HTML]{FFD9B3}28.91                       & 31.36                                               & \cellcolor[HTML]{FFD9B3}32.24                      \\ \hline
ZipNeRF                     & 25.80                                               & 22.40                                               & 28.20                                              & 27.55                                             & 23.89                                                 & 32.65                                            & 29.38                                               & 32.50                                               & 34.46                                              \\ \hline
\rowcolor[HTML]{F1F3F4} 
\textbf{SSIM $\uparrow$}    & \multicolumn{1}{l}{\cellcolor[HTML]{F1F3F4}bicycle} & \multicolumn{1}{l}{\cellcolor[HTML]{F1F3F4}flowers} & \multicolumn{1}{l}{\cellcolor[HTML]{F1F3F4}garden} & \multicolumn{1}{l}{\cellcolor[HTML]{F1F3F4}stump} & \multicolumn{1}{l|}{\cellcolor[HTML]{F1F3F4}treehill} & \multicolumn{1}{l}{\cellcolor[HTML]{F1F3F4}room} & \multicolumn{1}{l}{\cellcolor[HTML]{F1F3F4}counter} & \multicolumn{1}{l}{\cellcolor[HTML]{F1F3F4}kitchen} & \multicolumn{1}{l}{\cellcolor[HTML]{F1F3F4}bonsai} \\ \hline
3DGS                        & .766                                                & .606                                                & \cellcolor[HTML]{FFD9B3}.866                       & .771                                              & .633                                                  & \cellcolor[HTML]{FFD9B3}.919                     & \cellcolor[HTML]{FFD9B3}.909                        & \cellcolor[HTML]{FFB3B3}.928                        & \cellcolor[HTML]{FFB3B3}.942                       \\
StopThePop                  & \cellcolor[HTML]{FFFFB4}.768                        & .607                                                & \cellcolor[HTML]{FFD9B3}.866                       & .775                                              & .635                                                  & \cellcolor[HTML]{FFD9B3}.919                     & .907                                                & \cellcolor[HTML]{FFD9B3}.927                        & \cellcolor[HTML]{FFFFB4}.941                       \\
3DGRT                       & \cellcolor[HTML]{FFD9B3}.770                        & \cellcolor[HTML]{FFFFB4}.624                        & \cellcolor[HTML]{FFFFB4}.858                       & \cellcolor[HTML]{FFFFB4}.779                      & \cellcolor[HTML]{FFFFB4}.636                          & .917                                             & \cellcolor[HTML]{FFFFB4}.908                        & .924                                                & \cellcolor[HTML]{FFB3B3}.942                       \\
SMERF                       & .760                                                & \cellcolor[HTML]{FFD9B3}.626                        & .844                                               & \cellcolor[HTML]{FFB3B3}.784                      & \cellcolor[HTML]{FFB3B3}.682                          & \cellcolor[HTML]{FFFFB4}.918                     & .892                                                & .916                                                & \cellcolor[HTML]{FFFFB4}.941                       \\
Our model                   & \cellcolor[HTML]{FFB3B3}.776                        & \cellcolor[HTML]{FFB3B3}.639                        & \cellcolor[HTML]{FFB3B3}.869                       & \cellcolor[HTML]{FFD9B3}.781                      & \cellcolor[HTML]{FFD9B3}.656                          & \cellcolor[HTML]{FFB3B3}.922                     & \cellcolor[HTML]{FFB3B3}.910                        & \cellcolor[HTML]{FFFFB4}.926                        & \cellcolor[HTML]{FFD9B3}.943                       \\ \hline
ZipNeRF                     & .769                                                & .642                                                & .860                                               & .800                                              & .681                                                  & .925                                             & .902                                                & .928                                                & .949                                               \\ \hline
\rowcolor[HTML]{F1F3F4} 
\textbf{LPIPS $\downarrow$} & \multicolumn{1}{l}{\cellcolor[HTML]{F1F3F4}bicycle} & \multicolumn{1}{l}{\cellcolor[HTML]{F1F3F4}flowers} & \multicolumn{1}{l}{\cellcolor[HTML]{F1F3F4}garden} & \multicolumn{1}{l}{\cellcolor[HTML]{F1F3F4}stump} & \multicolumn{1}{l|}{\cellcolor[HTML]{F1F3F4}treehill} & \multicolumn{1}{l}{\cellcolor[HTML]{F1F3F4}room} & \multicolumn{1}{l}{\cellcolor[HTML]{F1F3F4}counter} & \multicolumn{1}{l}{\cellcolor[HTML]{F1F3F4}kitchen} & \multicolumn{1}{l}{\cellcolor[HTML]{F1F3F4}bonsai} \\ \hline
3DGS                        & .240                                                & .367                                                & \cellcolor[HTML]{FFD9B3}.123                       & .251                                              & .376                                                  & .287                                             & .258                                                & \cellcolor[HTML]{FFD9B3}.155                        & .252                                               \\
StopThePop                  & \cellcolor[HTML]{FFFFB4}.233                        & .362                                                & \cellcolor[HTML]{FFB3B3}.120                       & \cellcolor[HTML]{FFFFB4}.244                      & .366                                                  & .281                                             & \cellcolor[HTML]{FFFFB4}.253                        & \cellcolor[HTML]{FFB3B3}.154                        & .249                                               \\
3DGRT                       & \cellcolor[HTML]{FFD9B3}.226                        & \cellcolor[HTML]{FFFFB4}.335                        & \cellcolor[HTML]{FFFFB4}.134                       & \cellcolor[HTML]{FFD9B3}.243                      & \cellcolor[HTML]{FFFFB4}.364                          & \cellcolor[HTML]{FFFFB4}.280                     & \cellcolor[HTML]{FFD9B3}.248                        & \cellcolor[HTML]{FFFFB4}.156                        & \cellcolor[HTML]{FFFFB4}.242                       \\
SMERF                       & .239                                                & \cellcolor[HTML]{FFD9B3}.317                        & .147                                               & \cellcolor[HTML]{FFD9B3}.243                      & \cellcolor[HTML]{FFB3B3}.302                          & \cellcolor[HTML]{FFB3B3}.259                     & .256                                                & \cellcolor[HTML]{FFD9B3}.155                        & \cellcolor[HTML]{FFB3B3}.222                       \\
Our model                   & \cellcolor[HTML]{FFB3B3}.220                        & \cellcolor[HTML]{FFB3B3}.307                        & \cellcolor[HTML]{FFB3B3}.120                       & \cellcolor[HTML]{FFB3B3}.230                      & \cellcolor[HTML]{FFD9B3}.318                          & \cellcolor[HTML]{FFD9B3}.275                     & \cellcolor[HTML]{FFB3B3}.240                        & \cellcolor[HTML]{FFD9B3}.155                        & \cellcolor[HTML]{FFD9B3}.236                       \\ \hline
ZipNeRF                     & .228                                                & .309                                                & .127                                               & .236                                              & .281                                                  & .238                                             & .223                                                & .134                                                & .196                                              
\end{tabular}
    \caption{Full results for Mip-NeRF 360 dataset}
    \label{tab:full_mipnerf}
\end{table*}

\section{Hyperparameters, Etc} \label{sec:hparams}
We change the opacity learning rate to $0.0125$, the initial position learning rate to $4\times10^{-5}$ and the final position learning rate to $4\times10^{-7}$. We change the parameter known as ``percent dense'' in 3DGS to $0.001785$, which controls the size threshold above which primitives are split instead of cloned. We perform this every 200 iterations, instead of 100, and set the splitting gradient threshold to $2.5\times10^{-7}$ and the clone gradient threshold to $0.1$. We also stop splitting and cloning at 7 million primitives, or at 16000 iterations, which ever comes first, and start at 1500 iterations.

For color, we apply a softplus activation ($\beta=10$) to the output of the spherical harmonics (instead of 3DGS's relu activation), which we find avoids certain local minima where primitives get locked into a color. We increase the spherical harmonic degree every $2,\!000$ iterations, instead of $1,\!000$. We set the max primitive size to $25$ units, which improves performance.

\subsection{Inverse Contraction Initialization}
To help initialize the primitives in a scene, we supplement the SfM initialization with $10,\!000$ additional primitives. We generate these primitives by sampling their means uniformly from a radius-2 sphere. The radius is set to a constant value based the max radius at which the spheres could be packed into the radius-2 sphere, and colors are set to a constant value of $0.5$.
These primitives are then transformed by ``uncontracting'' the resulting means and covariances using the inverse of the contraction used in mip-NeRF 360~\cite{barron2022mip}. We found that highly anisotropic primitives at initialization can cause issues, so we scale the primitives to be isotropic.

To review, the mip-NeRF 360 contraction function $\mathcal{C}$ that maps from a 3D coordinate in Euclidean space $\mathbf{x}$ to a 3D coordinate in contracted space $\mathbf{z}$ is:
\begin{equation}
  \mathcal{C}(\mathbf{x}) = \mathbf{x} \cdot \frac{2 \sqrt{\max(1, || \mathbf{x} ||^2)} - 1}{\max(1, || \mathbf{x} ||^2)}
\end{equation}
The inverse of $\mathcal{C}(\mathbf{x})$ can be defined straightforwardly:
\begin{equation}
    \mathcal{C}^{-1}(\mathbf{z}) = \frac{\mathbf{z}}{\sqrt{\max(1, ||\mathbf{z}||^2)} (2 - \min(2, \sqrt{\max(1, ||\mathbf{z}||^2)}))}
\end{equation}
To apply this inverse contraction to a Gaussian instead of a point, we use the same Kalman-esque approach as was used in mip-NeRF 360: we linearize the contraction around $\mathbf{z}$ into a Jacobian-vector product, which we apply twice to the input covariance matrix.

}{}
\appendix

\end{document}